\documentclass[runningheads]{llncs}
\usepackage{graphicx}
\usepackage{tikz}
\usepackage{comment}
\usepackage{amsmath,amssymb,amsfonts} 
\usepackage{color}
\usepackage[inkscapearea=page]{svg}
\usepackage{adjustbox}

\usepackage[accsupp]{axessibility}  

\usepackage[width=122mm,left=12mm,paperwidth=146mm,height=193mm,top=12mm,paperheight=217mm]{geometry}

\usepackage{bm}
\usepackage{url}
\usepackage{subcaption}
\definecolor{myblue}{rgb}{0.2, 0.396, 0.643}
\definecolor{mygreen}{rgb}{0.506, 0.831, 0.102}
\definecolor{myred}{rgb}{1.0, 0.0, 0.0}
\usepackage{multirow}
\usepackage{hyperref}

\DeclareMathOperator*{\argmax}{arg\,max}
\DeclareMathOperator*{\argmin}{arg\,min}

\usepackage{dsfont}

\begin{document}

\pagestyle{headings}
\mainmatter

\title{Object Detection as Probabilistic Set Prediction} 
\titlerunning{Object Detection as Probabilistic Set Prediction}
\author{Georg Hess\inst{1,2}\and
Christoffer Petersson\inst{1,2} \and
Lennart Svensson\inst{1}}
\authorrunning{G. Hess et al.}
\institute{Chalmers University of Technology,
Gothenburg, Sweden, \email{georghe@chalmers.se} \and
Zenseact, Gothenburg, Sweden}

\maketitle

\begin{abstract}
Accurate uncertainty estimates are essential for deploying deep object detectors in safety-critical systems. The development and evaluation of probabilistic object detectors have been hindered by shortcomings in existing performance measures, which tend to involve arbitrary thresholds or limit the detector's choice of distributions. In this work, we propose to view object detection as a set prediction task where detectors predict the distribution over the set of objects. Using the negative log-likelihood for random finite sets, we present a proper scoring rule for evaluating and training probabilistic object detectors. The proposed method can be applied to existing probabilistic detectors, is free from thresholds, and enables fair comparison between architectures. Three different types of detectors are evaluated on the COCO dataset. Our results indicate that the training of existing detectors is optimized toward non-probabilistic metrics. We hope to encourage the development of new object detectors that can accurately estimate their own uncertainty. Code available at \url{https://github.com/georghess/pmb-nll}.

\end{abstract}

\section{Introduction}
Accurately locating and classifying a set of objects has a range of applications, such as autonomous driving, transportation, surveillance, scene analysis, and image captioning. Common approaches for solving this rely on a deep object detector which provides a set of detections containing bounding box parameters, semantic class and classification confidence. However, as pointed out in previous works \cite{choi2019gaussian,feng2020review,hall2020probabilistic,kendall2017uncertainties,valdenegro2021find} most state-of-the-art networks lack the ability to assess their own regression confidence and fail to provide a complete uncertainty description. As an effect, this can limit the performance in downstream tasks such as multi-object tracking, sensor fusion, or decision making, ultimately hindering humans to establish trust in the deep learning agent.

There are many strategies to evaluate predictive uncertainties in the deep learning regime. Broadly speaking, a distribution should perform well on two criteria: calibration and sharpness. For a distribution to be well calibrated, it should not be over- or under-confident, but reflect the true confidence in its predictions. Sharpness instead promotes concentrated and, consequently, informative distributions \cite{gneiting2007probabilistic}. Both these properties can be measured simultaneously by using a proper scoring rule such as negative log-likelihood \cite{gneiting2007strictly}. Proper scoring rules assess the quality of predictive uncertainties and are minimized only when the prediction is equivalent to the distribution that generated the ground truth observations \cite{gneiting2007strictly}. Besides measuring calibration and sharpness, proper scoring rules enable a theoretically sound ranking of different predictive distributions. 

\begin{figure}[t]
    \centering
    \begin{subfigure}[t]{0.49\textwidth}
        \centering
        \includegraphics[width=0.7\linewidth,trim={5cm 2cm 4cm 0cm},clip,page=9]{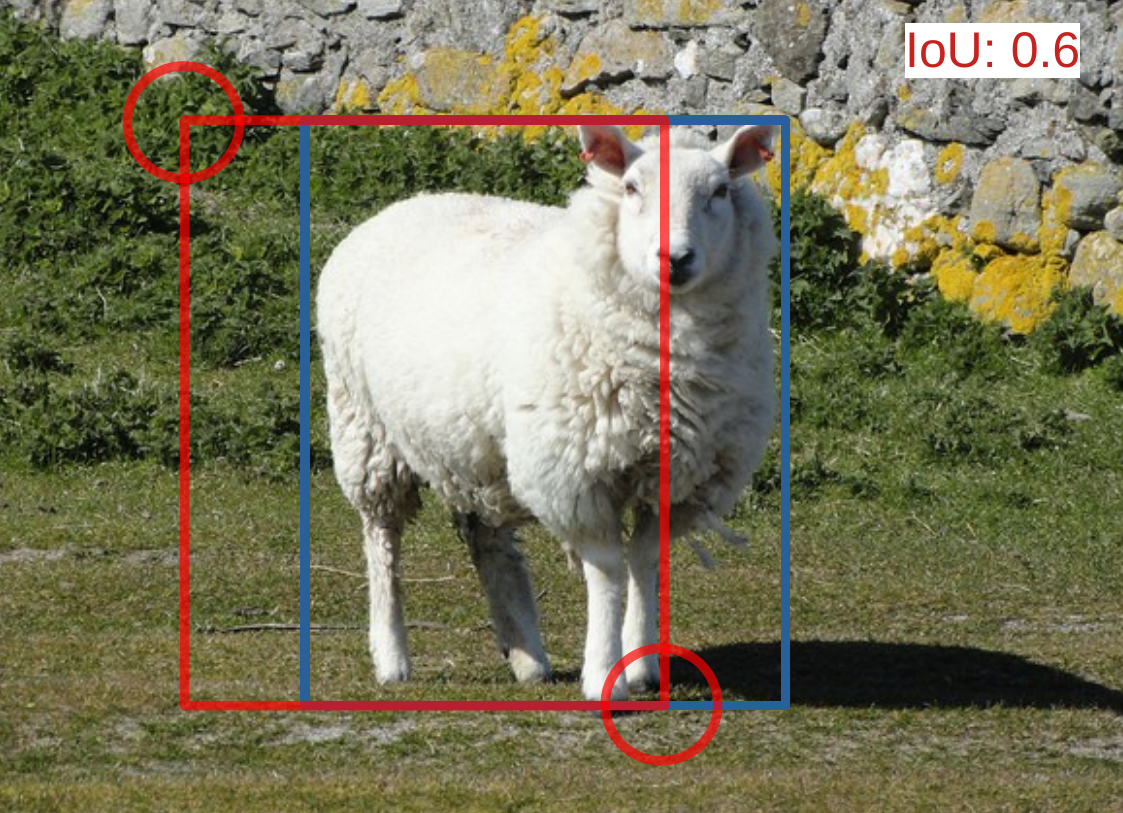}
        \caption{Assigning predictions to ground truth is non-trivial in probabilistic OD. The prediction can either be considered a true positive with bad uncertainty estimates or a false positive.}
        \label{fig:iou-is-bad-1}
    \end{subfigure}
    \hfill
    \begin{subfigure}[t]{0.49\textwidth}
        \centering
        \includegraphics[width=0.7\linewidth,trim={5cm 2cm 4cm 0cm},clip,page=10]{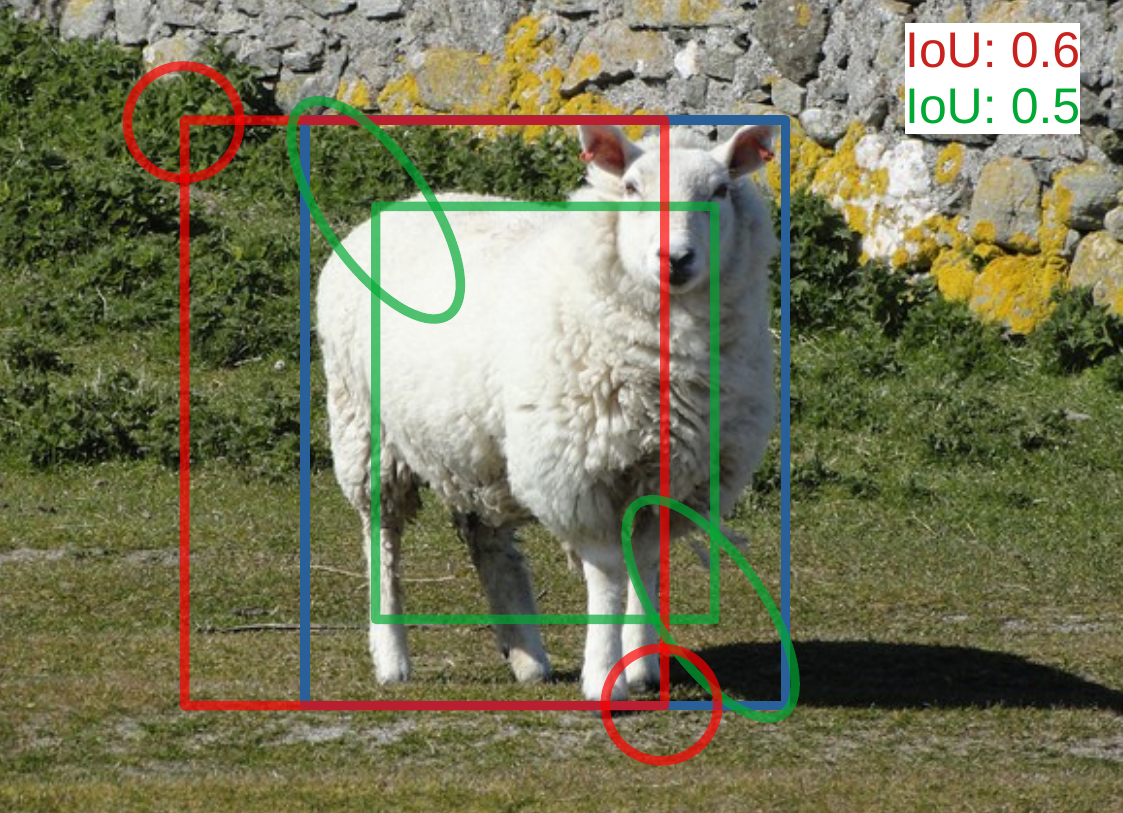}
        \caption{mAP and \cite{harakeh2021estimating} prefer the red prediction with larger IoU. Our method considers uncertainties and multiple assignments, and finds the green prediction a more probable match to the blue ground truth.}
        \label{fig:iou-is-bad-2}
    \end{subfigure}
    \caption{Predictions (red and green) and ground truth (blue), highlighting the object detection assignment problem. Ellipses represent spatial uncertainty.}
    \label{fig:iou-is-bad}
\end{figure}

Evaluating the quality of predictive uncertainties in object detection (OD) is a non-trivial task. First, any measure has to jointly consider the performance in terms of ability to detect, correctly classify and accurately locate objects. Second, as we do not know the correspondence between predictions and ground truth objects, any analysis is colored by the selected assignment rules. As an example, the prediction in Fig. \ref{fig:iou-is-bad-1} can be considered either a correct detection with bad uncertainty predictions or a false positive. Having multiple predictions makes the assignment even harder, as shown in Fig. \ref{fig:iou-is-bad-2}. The most common measure in OD, mAP, uses handcrafted assignment rules based on IoU and class confidence and fails to consider predicted uncertainties. The probability-based detection quality (PDQ) \cite{hall2020probabilistic} tries to address these issues, but is limited to Gaussian distributions for regression. More recently, the lack of proper scoring rules for evaluating probabilistic object detection was pointed out by \cite{harakeh2021estimating}, also proving that PDQ is not a proper scoring rule. However, while they use proper scoring rules for the different subtasks, such as the energy score for regression and the Brier score for classification, predictions are assigned to targets using ad hoc IoU-based rules which ignore regression uncertainties. As highlighted earlier, these types of assignment rules have a large influence on the reported performance, do not yield proper scoring rules, and make it harder to draw conclusions about model performance. 

In this paper, we propose to use random finite sets (RFS) to model the probabilistic object detection task. Object detection is often seen as a set prediction task, and we extend this perspective to probabilistic object detection (PrOD). We describe the set of objects in a given image by a single random variable, and the task of our object detection networks is to describe the distribution of that variable. This simple change of perspective enables us to use the negative log-likelihood to evaluate the uncertainty estimates of our detections, which gives rise to the first proper scoring rule for object detection. Our framework explicitly models the assignment problem, is general enough to be applied to any type of distribution, enables easy ranking between different algorithms, and can be decomposed to highlight different types of errors (detection, regression and classification). Our key contributions are the following.
\begin{itemize}
    \item We propose to view the set of objects in an image as a single stochastic variable. By applying the negative log-likelihood (NLL) to a distribution over sets, we present the first proper scoring rule for object detection.
    \item We show how to apply the random finite set framework to object detection by interpreting the detector output as parameters of multi-Bernoulli (MB) and Poisson multi-Bernoulli (PMB) densities. Further, we present how to efficiently calculate and interpret the NLL of the MB and PMB densities.
    \item  Using our proposed scoring rule, we evaluate one-stage, two-stage, and set-based detectors on the popular MS COCO dataset, and showcase their strengths and shortcomings using the decomposability of PMB-NLL. 
    \item Further, we leverage the fact that the proposed method is differentiable and fine-tune the detectors to optimize PMB-NLL directly. Our results show that this helps detectors to reduce the number of false and duplicate detections.
\end{itemize}


\section{Related Work}
Quantifying uncertainties with deep neural networks has been a long-standing challenge. We aim to provide a brief overview here, as to motivate the importance of our work. Interested readers are referred to \cite{abdar2021review,feng2020review,gawlikowski2021survey} for details.

\subsubsection{Types of Uncertainties.} In computer vision, uncertainties are generally divided into two categories: aleatoric and epistemic \cite{kendall2017uncertainties}. The first category refers to noise inherent to the data, which can originate from sensor noise, class ambiguities, label noise and such, and cannot be reduced with more data. Epistemic uncertainties are due to uncertainties in model parameters, and can, in principle, be eliminated given enough data. In this work, we do not aim to disentangle the two types, but consider overall predictive uncertainty \cite{skafte2019reliable}.

\subsubsection{Uncertainty Estimation.} Most approaches for quantifying uncertainties in object detection either apply Monte Carlo dropout \cite{harakeh2020bayesod,miller2018dropout,wirges2019capturing}, deep ensembles \cite{feng2019deep,lyu2020probabilistic} or direct modeling \cite{he2020deep,lee2020localization,zhou2021probabilistic}. Unfortunately, uncertainty estimates are often overlooked when evaluating probabilistic detectors, while methods that do evaluate their uncertainties use a range of different performance measures, making comparison challenging. The lack in standard performance measures has also been pointed out as a main obstacle for uncertainty estimation \cite{abdar2021review,feng2020review,valdenegro2021find}.

\subsubsection{Evaluating Uncertainty.} As the commonly used performance measure mAP fails to consider spatial uncertainties and is insensitive to badly calibrated classification, several methods trying to address these issues have been suggested. The Probability-based Detection Quality (PDQ) \cite{hall2020probabilistic} evaluates both spatial and semantic uncertainties, but is limited to Gaussian spatial uncertainties, requires practitioners to select confidence thresholds, and has been shown by \cite{harakeh2021estimating} to not be a proper scoring rule, thereby introducing biases into its ranking of detectors. The authors of \cite{harakeh2021estimating} promote the use of proper scoring rules for object detection. However, their approach disregards the spatial uncertainty information when assigning predictions to targets, requires confidence thresholds, and does not provide clear recommendations on model ranking.

\subsubsection{Set Prediction.} While object detection inherently can be seen as a set prediction task, this has been made more explicit by a range of set-based detectors \cite{detr,sun2021rethinking,zhang2019deep,deformable-detr}. These detectors highlight the assignment problem, i.e., how to assign predictions to ground truth elements when calculating losses or metrics. In this work, we extend this perspective to probabilistic object detection by modeling the problem using distributions over random finite sets. This paradigm is applicable to any type of detector, set-based or not, and naturally models and solves the assignment problem.

\subsubsection{Random Finite Sets.} Random finite sets have been used extensively in the model-based multi-object tracking community \cite{garcia2018poisson,mahler2014advances,vo2013labeled,williams2015marginal}. The RFS framework has proven useful for modeling potentially detected and undetected objects as it captures uncertainties in the cardinality of present objects and their individual properties. However, these algorithms are often evaluated without taking their uncertainties into account. Recently, the authors of \cite{pinto2021uncertainty} suggested the use of negative log-likelihood for probabilistic evaluation of model-based multi-object trackers and presented an efficient approximation of the NLL. Our work shows how to interpret parameters of existing deep object detectors as RFSs and uses \cite{pinto2021uncertainty} to calculate our proper scoring rule. Unlike the custom designed and low-dimensional regression problems explored in \cite{pinto2021uncertainty}, we apply this method to a large scale dataset, jointly evaluating detection, classification, and regression.

\section{Probabilistic Modeling for Object Detection}
Object detection is a set prediction task, where, given an image $\bm{X}$, the aim is to predict the set of corresponding objects $\mathbb{Y}$ present in said image. Here, the number of objects $n$ in the set $\mathbb{Y}=\{y_1,y_2,\dots,y_n\}$ is unknown beforehand. Further, for each object $y_i=(c_i, b_i)$, we do not know which class $c_i\in\{1,\dots,C\}$ it belongs to, nor where its bounding box $b_i\in \mathbb{R}^4$ is located in the image. In supervised learning, we aim to learn a model that, given the image $\bm{X}$, predicts a set of $\hat{n}$ objects $\hat{\mathbb{Y}}=\{\hat{y}_1, \hat{y}_2, \cdots, \hat{y}_{\hat{n}}\}$ which is close to the ground truth label $\mathbb{Y}$ in some sense. For probabilistic object detection, we further want an uncertainty description for the number of objects and their individual properties.

In this work, we evaluate probabilistic object detectors by seeing the set of objects $\mathbb{Y}$ as a single random variable. The task for our networks is to predict the distribution of this set $f(\mathbb{Y}|\bm{X})$. This is a natural and general probabilistic extension to the set prediction perspective, as a distribution over sets can capture the varying cardinality and uncertainty in properties for individual objects. Using this novel perspective, all predictions for a single image are evaluated together by applying the negative log-likelihood  
\begin{equation}
    \text{NLL}((\mathbb{Y},\bm X), f) = -\log(f(\mathbb{Y}|\bm{X})).
\end{equation}
This can be compared to existing methods where classification and regression are treated separately and evaluated conditioned on an ad hoc assignment rule \cite{harakeh2021estimating}, or network performance is measured using non-proper scoring rules \cite{hall2020probabilistic}. 

To use the negative log-likelihood in practice, we need our deep object detectors to predict distributions $f(\mathbb{Y}|\bm{X})$. We propose to use random finite sets (RFSs) and the Poisson multi-Bernoulli (PMB) distribution and demonstrate how the PMB parameters are naturally obtained from the output of standard probabilistic deep object detectors. Further, using the results of \cite{pinto2021uncertainty}, we show how to efficiently calculate and decompose the negative log-likelihood of $f(\mathbb{Y}|\bm{X})$ into detection, classification and regression errors.

\textit{Notation:} Scalars and vectors are denoted by lowercase or uppercase letters with no special typesetting $x$, matrices by uppercase boldface letters $\bm X$, and sets by uppercase blackboard-bold letters $\mathbb{X}$. We define $\mathbb N_a = \{i \in \mathbb N|i\leq a\}, a\in\mathbb N$. 

\subsection{Modeling Detections with Random Finite Sets}

We need a way to describe the distribution over $\mathbb{Y}$ using deep neural networks. Interestingly, existing probabilistic detectors already contain the parameters needed. To this end, we propose to model $\mathbb{Y}$ with random finite sets. Random finite sets are described using a multi-object density $f(\mathbb{Y})$, which means that sampling from $f(\mathbb{Y})$ yields finite sets of objects with varying cardinality, where objects consist of a class and a bounding box. We should note that RFSs are not the only way to describe a distribution over $\mathbb{Y}$. However, we will show that our method has multiple properties suitable for object detection and advantages such as being compatible with existing architectures.

\subsubsection{Bernoulli RFS.} One of the simplest RFSs is the Bernoulli RFS, commonly used for modeling single potential objects in the multi-target tracking community \cite{garcia2018poisson,GLMB}. Here, we use it to model each individual detected object, and its density is
\begin{equation}
    f_\text{B}(\mathbb{Y}) = \begin{cases}
    1-r & \text{if } \mathbb{Y}=\emptyset,
    \\
    rp(y) & \text{if } \mathbb{Y} = \{y\},
    \\
    0 & \text{if } |\mathbb{Y}| > 1,
    \end{cases}
\end{equation}
where $p(y)$ is the single-object density. For instance, assuming the class and bounding box to be independent, $p(y)=p_\text{cls}(c)p_\text{reg}(b)$ contains the class distribution $p_\text{cls}(c)$ and some density describing the object's spatial distribution $p_\text{reg}(b)$. Further, $r\in[0,1]$ is the probability of existence, which is the probability that the Bernoulli RFS yields an object when sampling from it. Note that a Bernoulli RFS can account for at most one object since the likelihood is zero for any set with cardinality greater than one.

\begin{figure}[t]
    \centering
    \includegraphics[width=\textwidth,trim={0cm 6.2cm 0cm 5.8cm},clip,page=11]{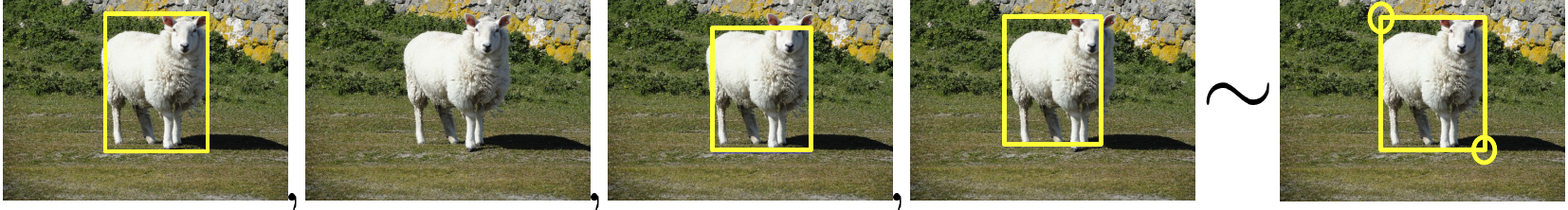}
    \caption{Four sampled sets (left) from a Bernoulli RFS (right) with existence probability $r=0.75$. The RFS can model the absence of objects, as well as semantic and spatial uncertainties. The image is only included for context.}
    \label{fig:bernoulli-example}
\end{figure}

The parameters of $p(y)$ are already present in probabilistic detectors. Depending on the architecture, we can interpret $r$ as objectness and predicted it directly, or, find it as the sum of probabilities assigned to foreground classes and let $p_\text{cls}(c)$ be the class distribution conditioned on existence. Note that $f_\text{B}(\emptyset)=1-r$ is the probability that the object is not present, which we may think of as the event where the prediction is background. An example of a Bernoulli RFS prediction and corresponding samples is shown in Fig. \ref{fig:bernoulli-example}.

\subsubsection{Multi-Bernoulli RFS.} Generally, the number of objects in an image can vary greatly. Modeling many potential objects can be achieved by taking the union of multiple Bernoulli RFSs \cite{garcia2018poisson}, resulting in a multi-Bernoulli (MB) RFS. In other words, individual predictions made by a detector are interpreted as parameters of individual Bernoulli RFSs, and by taking their union we combine them into a single random variable. Unlike a Bernoulli RFS, an MB RFS can be used to model the set of potentially detected objects for an entire image.

Formally, let $\mathbb{X}_1,\dots,\mathbb{X}_m$ be $m$ independent Bernoulli RFSs with the densities \\$f_{\text{B}_1}(\mathbb{X}_1), \dots, f_{\text{B}_m}(\mathbb{X}_m)$, existence probabilities $r_1, \dots, r_m$, and single-object densities $p_1(x), \dots, p_m(x)$. Then $\mathbb{X}=\cup_{i=1}^m\mathbb{X}_i$ is an MB RFS with multi-object density
\begin{equation}
\label{eq:mb-density}
    f_{\text{MB}}(\mathbb{X}) = \sum_{\uplus_{i=1}^m\mathbb{X}_i=\mathbb{X}} \prod_{j=1}^m f_{\text{B}_j}(\mathbb{X}_j),
\end{equation}
where $\sum_{\uplus_{i=1}^m\mathbb{X}_i=\mathbb{X}}$ denotes the sum over all disjoint sets whose union is $\mathbb{X}$. In other words, when evaluating the multi-object density $f_{\text{MB}}(\mathbb{Y})$ of a set $\mathbb{Y}$ we sum the multi-object densities of all possible assignments between elements in $\mathbb{Y}$ and Bernoulli components in $f_\text{MB}$.

\begin{figure}[t]
    \centering
    \begin{subfigure}[t]{0.49\textwidth}
        \centering
        \includegraphics[width=0.8\textwidth,trim={4cm 1cm 4cm 1cm},clip,page=4]{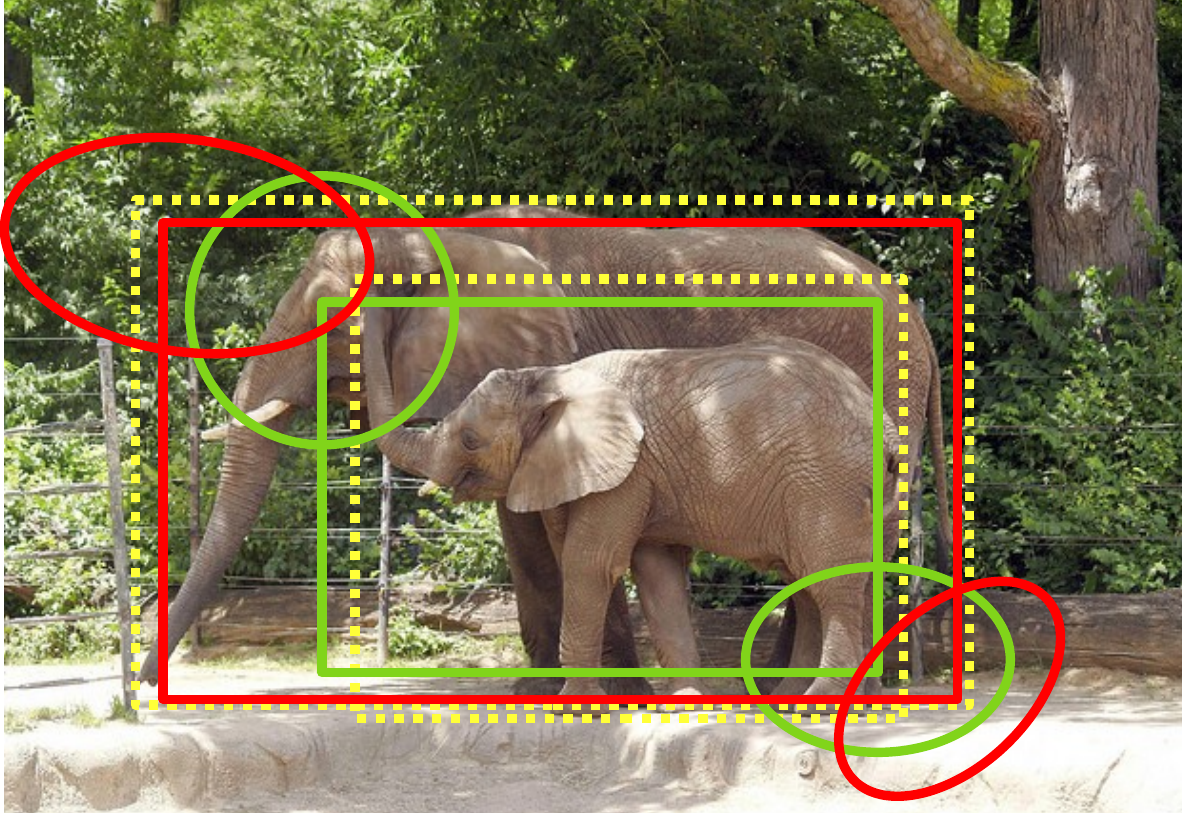}
        \caption{Set $\mathbb Y$ (yellow) with two ground truth objects and a multi-Bernoulli with two Bernoullis \textcolor{myred}{$\mathbb X_1$} and \textcolor{mygreen}{$\mathbb X_2$} with densities \textcolor{myred}{$f_{\text{B}_1}(\cdot)$} and \textcolor{mygreen}{$f_{\text{B}_2}(\cdot)$}. Ellipses represent uncertainties in bounding box location and shape. The large spatial uncertainties in \textcolor{myred}{$f_{\text{B}_1}(\cdot)$} make it a decent description of both true objects.}
        \label{fig:MB_ex_gt_and_predictions}
    \end{subfigure}
    \hfill
    \begin{subfigure}[t]{0.49\textwidth}
        \centering
        \includegraphics[width=0.8\textwidth,trim={7cm 3cm 7cm 3cm},clip,page=3]{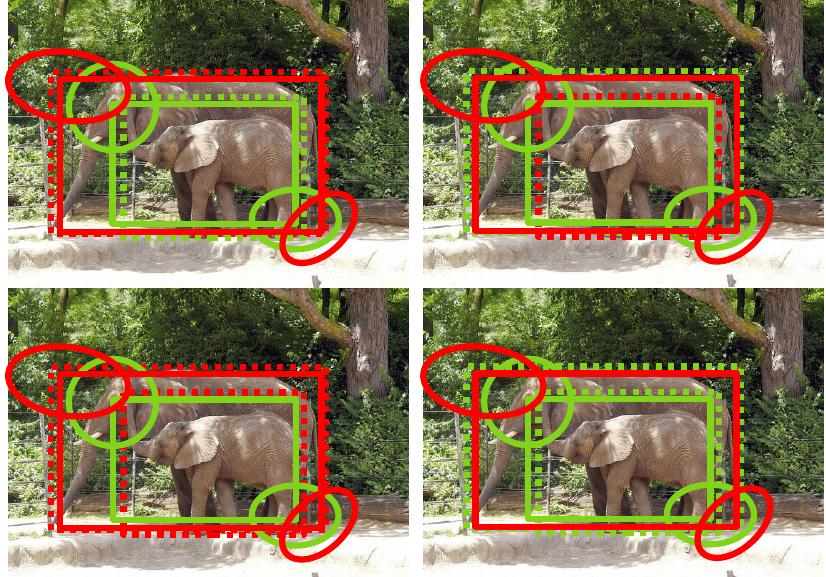}
        \caption{Visualization of the four potential assignments, ordered in decreasing likelihood. In the bottom row, both ground truth objects have been assigned to \textcolor{myred}{$\mathbb X_1$} and \textcolor{mygreen}{$\mathbb X_2$} respectively. As \textcolor{myred}{$f_{\text{B}_1}(\mathbb Y)$}$=$\textcolor{mygreen}{$f_{\text{B}_2}(\mathbb Y)$}$=0$ for sets with cardinality larger than one, both these assignments have a likelihood of zero.}
        \label{fig:MB_ex_matched}
    \end{subfigure}
    \caption{Visualization of likelihood evaluation for a multi-Bernoulli RFS.}
    \label{fig:mb_example}
\end{figure}

We illustrate this concept with an example. Consider an image containing two objects $\mathbb{Y}=\{y_1,y_2\}, y_i=(c_i,b_i)$ and two predictions, as shown in Fig. \ref{fig:mb_example}. Each prediction consists of a class distribution and a spatial pdf. We let these parameterize the densities $f_{\text{B}_1}(\cdot)$, $f_{\text{B}_2}(\cdot)$ of two separate Bernoulli RFS, whereas the multi-object density $f_{\text{MB}}(\mathbb Y)$ of their union is the MB RFS used to describe all objects in the image. When evaluating the likelihood $f_\text{MB}(\mathbb Y)$ using \eqref{eq:mb-density}, we sum the four ways to assign ground truth objects to the Bernoulli RFSs
\begin{equation}
\begin{split}
    f_\text{MB}(\mathbb Y) = &f_{\text{B}_1}(\{y_1\})f_{\text{B}_2}(\{y_2\})+
    f_{\text{B}_1}(\{y_2\})f_{\text{B}_2}(\{y_1\})+ 
    \\
    &f_{\text{B}_1}(\{y_1,y_2\})f_{\text{B}_2}(\emptyset)+
    f_{\text{B}_1}(\emptyset)f_{\text{B}_2}(\{y_1,y_2\}),
\end{split}
\end{equation}
where each individual assignment is visualized in Fig. \ref{fig:MB_ex_matched}. As $f_{\text{B}_1}(\cdot)$ and $f_{\text{B}_2}(\cdot)$ both evaluate to zero for sets with more than one element, the last two assignments have a likelihood of zero, and we are left with two terms
\begin{equation}
\begin{split}
    f_\text{MB}(\mathbb Y) = & r_1 p_{1,\text{cls}}(c_1)p_{1,\text{reg}}(b_1) \cdot 
    r_2 p_{2,\text{cls}}(c_2)p_{2,\text{reg}}(b_2) +
    \\
    & r_1 p_{1,\text{cls}}(c_2)p_{1,\text{reg}}(b_2) \cdot
    r_2 p_{2,\text{cls}}(c_1)p_{2,\text{reg}}(b_1).
\end{split}
\end{equation}

In contrast to existing methods with handcrafted assignment rules \cite{hall2020probabilistic,harakeh2021estimating,lin2014microsoft}, the assignment problem is modeled explicitly and rigorously by the RFS framework. The intuition behind considering all possible assignments is that we cannot know the correspondence between ground truths and predictions. In cases with overlapping boxes, predictions may have large IoU with multiple objects, making the assignment highly ambiguous. Further, for PrOD, large uncertainties can make it even harder to pair predictions to true objects.

\subsection{Proper Scoring Rule for Object Detection}
A scoring rule measures the quality of predictive uncertainty in terms of sharpness and calibration \cite{gneiting2007strictly}. It does so by assigning a numerical value $S(p_\theta, (\mathbf{x},\mathbf{y}))$ to a predicted distribution $p_\theta(\mathbf{y}|\mathbf{x})$, given that some event $(\mathbf{x},\mathbf{y}) \sim p^*(\mathbf{y}|\mathbf{x})p(\mathbf{x})$ materialized, where a lower number indicates better quality. A scoring rule is further known to be strictly proper if it is minimized only when $p_\theta$ is equal to the distribution $p^*$ that generated the observed event. For OD this translates to the predictive distribution being the same as the distribution from which the annotations have been generated. The noise present in a perfect prediction $p_\theta$ should in other words be equal to the noise in the annotations. These properties make proper scoring rules suitable for evaluating and ranking different predictions. 

Negative log-likelihood (NLL) is a local proper scoring rule used to evaluate the quality of predictive distributions for both regression and classification. Local refers to the fact that the predicted distribution is only evaluated at the event that materialized \cite{dawid2014theory}. If we let $\mathbb{Y}$ denote the set of ground truth objects present in the current image and let $f_\text{MB}(\mathbb{X})$ be the multi-object density of an MB RFS produced by some model, then
\begin{equation}
\label{eq:nll_mb}
    \text{NLL}(\mathbb{Y}, f_\text{MB}) = -\log f_\text{MB}(\mathbb{Y}) = -\log \left( \sum_{\uplus_{i=1}^m\mathbb{Y}_i=\mathbb{Y}} \prod_{j=1}^m f_{\text{B}_j}(\mathbb{Y}_j) \right).
\end{equation}

As discussed in the previous section, to evaluate the likelihood of an MB RFS density $f_\text{MB}(\mathbb{Y})$, we consider all possible assignments. As the number of predictions $m$, or the cardinality of the ground truth set $|\mathbb{Y}|$ grows, the number of assignments grows super-exponentially, making the NLL computation intractable. However, recently it was shown how to efficiently approximate the NLL of certain RFS densities, including the MB density \cite{pinto2021uncertainty}. Assuming that the ground truth objects, as well as the individual Bernoulli components, are somewhat separated, only a few assignments have a substantial contribution to the overall likelihood. Referring back to the example from Fig. \ref{fig:mb_example}, we can see that mainly the first assignment contributed to the sum of likelihoods. Thus, we approximate the NLL by only considering the most likely assignments. We find these assignments efficiently by solving an optimal assignment problem
\begin{subequations}
\label{eq:assignment_problem}
\begin{align}
    \min_{\bm{A}} \quad & \sum_k\sum_{l} C_{k, l}A_{k,l}
    \\
    \label{eq:assignment_problem_constraints}
    \text{s.t.} \quad & \sum_{k=1}^{m+|\mathbb Y|} A_{k,l}=1,\, \sum_{l=1}^{|\mathbb Y|} A_{k,l}\leq 1,
    \\
    \label{eq:cost_matrix_mb}
    & C_{k, l} = 
    -\log\left(\frac{p_k(y_l)}{1-r_k}r_k\right),
\end{align}
\end{subequations}
where $\bm{C}$ is a cost matrix and its derivation can be found in Appendix A. In \eqref{eq:cost_matrix_mb}, both the cost of assigning the object $l$ to prediction $k$, $p_k(y_l)r_k$, and the alternative of not assigning the prediction to anything, $1-r_k$, are considered jointly. The assignment matrix $\bm{A}$ describes the pairing between predictions and ground-truth objects, where ground truth object $y_l$ is assigned to the $k$-th component of the MB i.f.f. $[\bm{A}]_{k,l}=1$. Murty's algorithm \cite{motro2019scaling,murty1968algorithm} efficiently computes the $Q$ lowest cost associations $\bm{A}^{*}_1, \cdots, \bm{A}^{*}_Q$ to this assignment problem. We obtain
\begin{align}
    \label{eq:approximated_nll_mb}
    \mathrm{NLL}(\mathbb Y,f_{\text{MB}}) \approx 
    -\log \left( \sum_{q=1}^Q
    \prod_{k=1}^{m}f_{\text{B}_k}(\mathbb{Y}_k(\bm{A}_q^{*})) \right),
\end{align}
where $\mathbb Y_k(\bm{A}_q^{*})=\{y_j\in\mathbb Y | [\bm{A}_q^{*}]_{k,j}=1\}$, i.e., $\mathbb{Y}_k$ contains the ground truth $y_j$ if $y_j$ was assigned to Bernoulli component $k$, otherwise it is the empty set. Comparing this expression to \eqref{eq:nll_mb}, only $Q$ terms have to be calculated. During our experiments, we use $Q=25$ as we find that the approximation does not change considerably when using additional assignments. 

\subsection{Modeling All Objects}
Using only a MB RFS to describe the objects in an image can be problematic as it assumes that the number of predictions is greater than or equal to the number of objects present. For an algorithm providing too few detections, multiple objects are assigned to the same Bernoulli in \eqref{eq:mb-density}, resulting in the MB likelihood being zero and an infinite NLL. Fortunately, there are RFSs that can model an arbitrary number of objects. Within model-based multi-object tracking, the Poisson Point Process (PPP) is used to model undetected objects \cite{garcia2018poisson}, and we show here how to use it for OD to ensure a finite NLL. The PPP is then combined with the detections, yielding the Poisson multi-Bernoulli (PMB) RFS. Importantly, we also establish a technique to obtain the PPP directly from the output of our deep object detectors.

\subsubsection{Poisson Point Process.} Intuitively speaking, the PPP is intended to capture objects that are not properly detected. By complementing the detections in the MB, we model both the detected and undetected objects in an image. In contrast to the MB, the cardinality of a PPP is Poisson distributed which gives it a non-zero probability for any set cardinality. Thus we avoid the issue of infinite NLL due to lack of detections. 
The multi-object density of a PPP is
\begin{equation}
    f_\text{PPP}(\mathbb{X}) = \exp \left( -\Bar{\lambda} \right)\prod_{x\in \mathbb{X}}\lambda(x),
\end{equation}
where $\lambda(\cdot)$ is the intensity function and $\Bar{\lambda}=\int \lambda(x')dx'$ is the expected cardinality of the set. The intensity function is expected to describe the properties of poorly detected objects, e.g., partially occluded objects, far-away objects, or even classes of objects that are inherently harder to detect. The intensity function is similar to a density function, but its integral does not have to sum to one. 

\subsubsection{Poisson multi-Bernoulli RFS.} With models for both detected and undetected objects, we have to combine them to a single model for all objects. To this end, we propose to use a Poisson multi-Bernoulli (PMB) RFS, which is the union of a PPP and an MB RFS. The PMB RFS also arises naturally as the posterior density of all objects after a single measurement update, when using standard models in model-based target tracking \cite{garcia2018poisson,williams2015marginal}. 

The multi-object density of a PMB is
\begin{equation}
    f_\text{PMB}(\mathbb{X}) = \sum_{\mathbb{X}^{\text{U}} \uplus \mathbb{X}^{\text{D}}=\mathbb X}f_\text{PPP}(\mathbb{X}^{\text{U}})f_\text{MB}(\mathbb{X}^{\text{D}}),
\end{equation}
were $\mathbb{X}^{\text{U}} \uplus \mathbb{X}^{\text{D}}$ refers to summing over all possible ways of partitioning $\mathbb{X}$ into two disjoint sets, one being the set of undetected objects $\mathbb{X}^\text{U}$ and the other one being the set of detected objects $\mathbb{X}^\text{D}$. When evaluating the likelihood of a set $\mathbb{Y}$ this translates to, for each object in $\mathbb{Y}$, considering it to be detected and assigning it to a Bernoulli following \eqref{eq:mb-density}, or it being undetected and assigning it to the PPP.

\subsubsection{Selecting the PPP Intensity.}
To use the PMB in object detection, we must describe the PPP intensity function $\lambda(\cdot)$. During this work, we explored various ways of learning $\lambda(\cdot)$ from data, e.g., estimating the parameters of a uniform intensity function or describing $\lambda(\cdot)$ as a constant mixture model. However, the method we found to work best for the detectors considered in our experiments, is to create $\lambda(\cdot)$ from low confidence predictions. In practice, we parameterize the intensity function as the unnormalized mixture of low confidence predictions where the mixture weights are the existence probabilities 
\begin{equation}
    \label{eq:intensity_func_mixture}
    \lambda(x) = \sum_i r_ip_i(x).
\end{equation}
Specifically, we remove all predictions from the MB RFS, whose existence probabilities are $r<0.1$, and instead use them to construct the intensity function using \eqref{eq:intensity_func_mixture}. The theoretical motivation for this change is that the Kullback-Leibler divergence between a Bernoulli RFS with existence probability $r<0.1$ and a PPP with intensity function $\lambda(x)=r p(x)$ is small \cite{williams2012hybrid}. The proposed PMB density should therefore be a good approximation to the MB density that we had before, but this minor adjustment is sufficient to avoid issues with infinite NLL.

\subsubsection{NLL Evaluation.}
For evaluating the NLL of a PMB RFS, we use the same approach as for the MB and consider only the $Q$ most likely assignments. The cost matrix from \eqref{eq:cost_matrix_mb} used in the optimization is extended to
\begin{equation}
    \label{eq:cost_matrix_pmb}
    C_{k, l} = \begin{cases} 
    -\log\left(\frac{p_k(y_l)}{1-r_k}r_k\right), &\text{if }k\leq |\mathbb Y|
    \\
    -\log\lambda(y_l),& \text{if } k=l+|\mathbb Y|
    \\
    \infty,& \text{otherwise,}
    \end{cases}
\end{equation}
which translates to appending a diagonal matrix of size $|\mathbb{Y}|\times |\mathbb{Y}|$ to the original cost matrix. The NLL from \eqref{eq:approximated_nll_mb} is extended as
\begin{align}
    \label{eq:approximated_nll_pmb}
    \mathrm{NLL}(\mathbb Y,f_{\text{PMB}}) \approx  \int\lambda(y')\mathrm{d}y'
    -\log \Big( \sum_{q=1}^Q \prod_{y\in\mathbb{ Y}^{\text{U}}(\bm{A}_q^{*})}\lambda(y)\prod_{k=1}^{m}f_{\text{B}_k}(\mathbb{Y}_k(\bm{A}_q^{*})) \Big),
\end{align}
where we define $\mathbb Y^\text{U}(\bm{A}_q^{*})=\mathbb Y \setminus \cup_{i=1}^m \mathbb Y_i(\bm{A}_q^{*})$, i.e., $\mathbb{Y}^U$ contains all the ground truth elements matched to the PPP. 

\subsubsection{NLL Decomposition.} Often the most likely assignment yields a good approximation to the NLL. For $Q=1$, the NLL can be decomposed into four parts and expressed in terms of assignments
\begin{align}
    \label{eq:pmb_nll_decomposition}
    \text{NLL}(\mathbb Y,f_\text{PMB})\approx\min_{\gamma\in\Gamma}
    & \underbrace{-\sum_{(i,j)\in\gamma}\log \big(r_i p_{i,\text{cls}}(c_j)\big)}_\text{Classification} \underbrace{- \sum_{(i,j)\in\gamma}\log \big(p_{i,\text{reg}}(b_j)\big)}_{\text{Regression}} 
    \\
    &\underbrace{-\sum_{i\in\mathbb F(\gamma)}\log(1-r_i)}_\text{False detections}
    \underbrace{+\int \lambda(y')\textrm{d}y'-\sum_{j\in\mathbb M(\gamma)}\log\lambda(y_j)}_\text{Missed objects},\notag
\end{align}
where $\Gamma$ is the set of all possible assignment sets, $(i,j)\in \gamma$ means that prediction $i$ has been assigned to ground truth $j$, and $\mathbb F(\gamma)=\left\{ i\in\mathbb N_m | \nexists j:(i,j)\in\gamma \right\}$ is the set of indices of the Bernoullis not matched to any ground-truth, i.e. false positives. Note that we assume the classification and regression distributions are independent $p_i(x)=p_{i,\text{cls}}(\cdot) p_{i,\text{reg}}(\cdot)$ for this decomposition. Further, we define $\mathbb M(\gamma)=\left\{j\in\mathbb N_{|\mathbb Y|}|\nexists i: (i,j)\in\gamma\right\}$ as the set of indices of ground-truths not matched to any Bernoulli component, i.e., missed objects. This decomposition enables further insight into the types of errors made by an algorithm, e.g., instead of treating all false positives equally as in \cite{hall2020probabilistic}, we take their existence probability into account for deciding how much to penalize an algorithm.
 
\section{Experiments}
For our experiments, we evaluate three existing object detection models: DETR \cite{detr}, RetinaNet \cite{retinanet}, and Faster-RCNN \cite{ren2015faster}, all using ResNet50 backbones. These are chosen to represent a set-based, one-stage, and two-stage detector, which highlights that the RFS framework is applicable regardless of architecture. All these models are publicly available through the Detectron2 \cite{wu2019detectron2} object detection framework, with hyperparameters\footnote{Hyperparameters are used as is unless stated otherwise.} optimized to produce competitive detection results for the COCO dataset \cite{lin2014microsoft}. Further, the models have previously been retrofitted with variance networks to estimate their spatial uncertainty \cite{harakeh2021estimating}. Due to hardware limitations, models in \cite{harakeh2021estimating} used a smaller batch size and adjusted learning rates, resulting in decreased mAP compared to numbers reported by Detectron2. For fair comparison, we use the same hyperparameters as \cite{harakeh2021estimating}, but note that increasing the batch size can improve mAP for all models.

The models are also fine-tuned with MB-NLL \eqref{eq:approximated_nll_mb} as loss function. During training, the aim is to detect all objects, hence the PPP for undetected objects is ignored. We also found training to be more stable when the number of assignments $Q$ is set to one. Further, when calculating the assignment costs for matching, ignoring spatial uncertainties improved training stability. That is, in \eqref{eq:cost_matrix_mb}, we use the L2 distance instead of $\log( p_{k,\text{reg}})$. This can be thought of as learning the spatial uncertainty given the predicted mean of the bounding box. 

For evaluation, the $Q=25$ assignments with the highest likelihood are used to approximate the PMB-NLL, as larger values for $Q$ do not affect the approximation for any of the models considered. In contrast to training, the matching cost is used as described in \eqref{eq:cost_matrix_pmb}. Further, following the COCO standard, models are limited to 100 predictions and no confidence threshold is used. DETR is designed to provide exactly 100 predictions, while we apply NMS and keep the 100 top-scoring predictions for RetinaNet and Faster-RCNN. For all models, bounding boxes are parameterized by their top-left and bottom-right coordinates $[x_1,y_1,x_2,y_2]$. While the pre-trained models from \cite{harakeh2021estimating} used a Gaussian distribution for regression, we found that using a Laplace distribution results in considerably lower NLL for both training and evaluation, across all models.

\subsubsection{Evaluating Object Detection with Proper Scoring Rule.}
We report mAP, PDQ \cite{hall2020probabilistic} and PMB-NLL in Table \ref{tab:coco_results} and the decomposed results following \eqref{eq:pmb_nll_decomposition} are shown in Table \ref{tab:coco_decomposition}, with additional analysis in the supplementary material. For models with loss ES (energy score) and NLL, please refer to \cite{harakeh2021estimating} for their details. PDQ is reported both when thresholding prediction confidences at the detectors' optimal F1-score and without any thresholding. The decomposition in Table \ref{tab:coco_decomposition} is calculated for the assignment with the lowest NLL and shown averaged per image and per assigned objects. For instance, the DETR ES regression term 82.3/12.3 is read as, on average the regression distribution of \textit{all} matched predictions contributes with 82.3 to the overall NLL. For a \textit{single} matched prediction, it contributes with 12.3 to the total NLL on average. 

\begin{table}[t]
    \centering
    \caption{mAP, PMB-NLL and PDQ with/without threshold for three detectors on the COCO validation set. $^*$ detections excluded due to $\infty$ NLL.}
    \begin{tabular}{llllll}
        \hline\noalign{\smallskip}
        Detector & Loss &  PMB-NLL $\downarrow$ & mAP $\uparrow$ & PDQ@F1 $\uparrow$ & PDQ@0.0 $\uparrow$\\
        \hline\noalign{\smallskip}
        \multirow{3}{*}{DETR} & ES & \textbf{120.33} & \textbf{0.407 }& 0.262 & \textbf{0.033} \\  \cline{2-6}
        & NLL  & 152.13 & 0.376 & 0.113 & 0.014 \\ \cline{2-6}
        & MB-NLL & 124.20 & 0.389 & \textbf{0.271} & 0.023 \\ \hline
        \multirow{3}{*}{RetinaNet} & ES & 127.66 & \textbf{0.362} & 0.228 & \textbf{0.028} \\ \cline{2-6}
        & NLL & 126.86 & 0.351 & 0.185 & 0.021 \\ \cline{2-6}
        & MB-NLL & \textbf{121.02} & 0.361 & \textbf{0.251} & 0.023 \\ \hline
        \multirow{3}{*}{Faster-RCNN} & ES$^*$  & 140.53 & \textbf{0.373} & 0.281 & \textbf{0.087} \\ \cline{2-6}
        & NLL$^*$ & 139.08 & 0.371 & \textbf{0.282} & \textbf{0.087} \\ \cline{2-6}
        & MB-NLL & \textbf{117.77} & 0.326 & 0.199 & 0.024 \\ \hline
    \end{tabular}
    \label{tab:coco_results}
\end{table}

We can see from Table \ref{tab:coco_results} that optimizing the networks toward MB-NLL rather than the NLL formulation used in \cite{harakeh2021estimating} consistently gives lower PMB-NLL at evaluation. With the exception of Faster-RCNN, this lower PMB-NLL is achieved without sacrificing mAP performance. We can also note that mAP does not indicate quality of predictive uncertainty. For instance, Faster-RCNN trained with the energy score achieves competitive performance in terms of mAP, but its uncertainty estimates result in the second worse PMB-NLL among the models. Further, although PDQ is described as a threshold-free performance measure, it is sensitive to false positive detections regardless of their confidence, as predictions with low and high confidence receive the same penalty by PDQ. When including low confidence predictions (PDQ@0.0), the reported PDQ results become hard to distinguish between detectors. For PMB-NLL, FP penalties are instead proportional to the predicted existence probabilities.

\begin{table}[t]
    \centering
    \caption{Decomposed PMB-NLL on COCO validation set. Numbers are given as $[$mean per image$]$/$[$mean per prediction$]$. FP=NLL of unmatched predictions. PPP match+PPP rate=missed objects. $^*$detections excluded due to $\infty$ NLL.}
    \begin{tabular}{lllllll}
        \hline\noalign{\smallskip}
        Detector & Loss & Regression $\downarrow$ & Classification $\downarrow$ & FP $\downarrow$ & PPP match $\downarrow$ & PPP rate $\downarrow$\\
        \hline\noalign{\smallskip}
        \multirow{3}{*}{DETR} & ES & 82.3\hphantom{0}/\textbf{12.3} & 3.79/0.57 & 17.6/0.71 & 15.6/\textbf{23.3} & 1.46 \\ \cline{2-7}
        & NLL & 124.7/17.5 & 3.57/\textbf{0.50} & 18.4/\textbf{0.59} & 5.0\hphantom{0}/23.8 & 1.52 \\ \cline{2-7}
        & MB-NLL & 93.3\hphantom{0}/14.6 & 3.26/0.51 & 3.1\hphantom{0}/0.99 & 24.8/25.4 & 0.18 \\ \hline
        \multirow{3}{*}{RetinaNet} & ES & 103.1/14.4 & 7.95/1.11 & 10.1/\textbf{0.22} & 4.3\hphantom{0}/23.8 & 2.87 \\ \cline{2-7}
        & NLL &105.0/14.5 & 7.87/1.09 & 9.8\hphantom{0}/\textbf{0.22} & 2.3\hphantom{0}/\textbf{20.7} & 2.91 \\ \cline{2-7}
        & MB-NLL & 79.9\hphantom{0}/\textbf{13.9} & 4.00/\textbf{0.70} & 2.6\hphantom{0}/0.44 & 34.1/21.1 & 0.98 \\ \hline
        \multirow{3}{*}{F-RCNN} & ES$^*$ & 105.5/15.1 & 8.23/1.18 & 12.9/0.57 & 14.0/37.1 & 0.43 \\ \cline{2-7}
        & NLL$^*$ & 104.7/15.0 & 8.23/1.18 & 13.2/0.58 & 13.5/36.1 & 0.43 \\ \cline{2-7}
        & MB-NLL & 62.0\hphantom{0}/\textbf{12.3} & 4.94/\textbf{0.98} & 3.3\hphantom{0}/\textbf{0.36} & 46.8/\textbf{20.4} & 1.30 \\ \hline
    \end{tabular}
    \label{tab:coco_decomposition}
\end{table}

\begin{figure}[t]
    \centering
    \includegraphics[width=\textwidth,trim={0cm 4.445cm 0.5cm 4.445cm},clip,page=12]{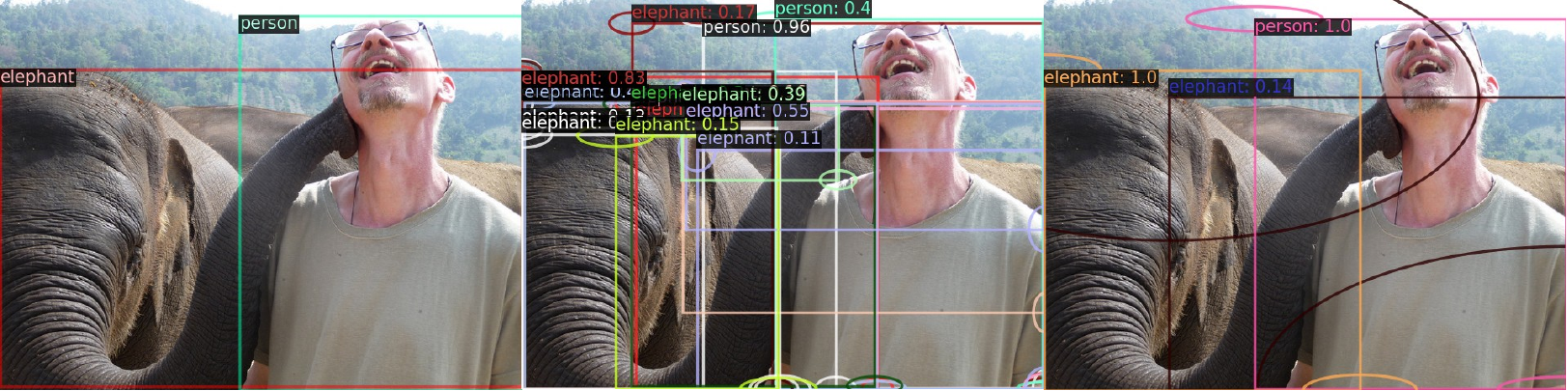}
    \caption{Ground truth (left), predictions from DETR ES (middle) and DETR MB-NLL (right). Predictions with a score less than 0.1 not shown. Models not trained with MB-NLL tend to produce many false positive detections.}
    \label{fig:result-example}
\end{figure}

Inspecting the decomposition of PMB-NLL in Table \ref{tab:coco_decomposition} gives further insights into the strengths and weaknesses of the detectors. Models that have not been trained with MB-NLL, show high penalties for producing many false positives. This is exemplified in Fig. \ref{fig:result-example}, where the model trained with MB-NLL produces a single prediction per object, and fewer false detections. Rather than producing multiple plausible predictions per object, where each prediction has low spatial uncertainty, they are compiled into a single hypothesis with slightly larger uncertainty. More examples of this are available in the Appendix. We theorize that the ES and NLL training is optimized toward mAP evaluation, where low confidence predictions are not penalized as heavily, and that the MB-NLL loss instead encourages models to produce plausible set predictions. Comparing across architectures, we can see from FP penalties that RetinaNet generally assigns lower existence probabilities to its incorrect predictions compared to DETR. For the matched predictions, DETR instead has the strongest classification performance, indicating that DETR generally has higher existence probabilities for its predictions, regardless of them being assigned to a ground truth or not.

The example in Fig. \ref{fig:result-example} also underlines important advantages with our evaluation. For the person in the image, DETR ES has one prediction with good regression but low confidence (in turquoise with 0.4), and one confident (in white with 0.96) with bad regression. Depending on which prediction is assigned to the true object, the error is either related to classification or regression. As highlighted previously, confidence thresholds are in practice needed by PDQ and used explicitly in other methods \cite{harakeh2021estimating}. Thus, existing methods only consider the confident prediction and report large regression errors. In contrast, PMB-NLL evaluates both possibilities and seamlessly weighs their contribution based on their individual likelihood, where the most likely assignment is in fact the one with lower existence probability.

Further, it is interesting to study the balance between matched predictions, and ground truth objects matched to the PPP. For Faster-RCNN trained with MB-NLL, many objects are assigned to the PPP. However, its PPP is a reasonably good description of the missed objects, resulting in a total PMB-NLL which is lower than the other models. For an application where a high recall level is desirable, the RetinaNet ES model might be a better choice, at the cost of worse regression and classification performance.






\section{Conclusions}
We propose the use of random finite sets for probabilistic object detection. Instead of predicting a set of objects $\mathbb{X}$, we ask our models to predict the distribution over the set of objects $f(\mathbb{X})$. Using a distribution over sets enables us to evaluate model performance for the true set of objects $\mathbb{Y}$ by applying the proper scoring rule negative log-likelihood $-\log(f(\mathbb{Y}))$. Our proposed method is general enough to be applied to detectors with any type of regression or classification distribution. It handles the assignments between predictions and objects automatically and can be decomposed into different error types. We evaluate three types of detectors using our new scoring rule and highlight their strengths and weaknesses. Our method enables fair comparison between probabilistic object detectors and we hope this will encourage the creation of novel architectures that aim for accurate uncertainty estimates rather than just accurate means. Future directions include how to better optimize networks toward our scoring rule and exploring further scoring rules within the random finite set framework.

\subsubsection{Acknowledgements}
This work was partially supported by the Wallenberg AI, Autonomous Systems and Software Program (WASP) funded by the Knut and Alice Wallenberg Foundation. Computational resources were provided by the Swedish National Infrastructure for Computing at \href{https://www.c3se.chalmers.se/}{C3SE} and \href{https://www.nsc.liu.se/}{NSC}, partially funded by the Swedish Research Council, grant agreement no. 2018-05973.

\bibliographystyle{splncs04}
\bibliography{references.bib}

\clearpage
\appendix
\section{Cost Matrix} \label{appendix:cost_matrix}
Suppose that we have a PMB density with Poisson intensity $\lambda(\cdot)$ and with $m$ Bernoulli components where the $i$-th Bernoulli component has probability of existence $r_i$ and existence-conditioned object density $p_i(\cdot)$. Note that we can model an MB as a PMB with Poisson intensity $\lambda(\cdot)=0$, which means that it is enough to describe how to handle PMB densities. To evaluate the multi-object density of a PMB, we have to calculate all possible assignments, which can be computationally intractable when working with many elements. However, we can approximate the PMB likelihood by only considering the assignments with the highest likelihood. This section shows how to find these assignments by solving an optimal assignment problem and how to select the corresponding cost matrix.

Before formulating the optimal assignment problem, we remind ourselves of the problem setting. For an object with state $y_j\in\mathbb{Y}= \{y_1,\dots,y_n\}$, the single object likelihood is proportional to $\lambda(y_j)$ if it is associated to the PPP and proportional to $r_i p_i(y_j)$ if it is associated to the $i$-th Bernoulli component. If this Bernoulli component is not associated to any object states, then the likelihood is $1-r_i$. 

Next, to formulate the optimization problem of finding the assignment with highest likelihood we introduce an association variable. Define a surjective association $\theta: \{1, \dots, n\} \rightarrow \{0, 1, \dots, m\} $ such that $\theta(i)=\theta(j) \in \{1, \dots, m\}$ if and only if $i=j$. If $\theta(j)=0$, object $y_j$ is associated to the PPP, while $\theta(j) = i > 0$ means that object state $y_j$ is associated to the $i$-th Bernoulli component. Further, let $\Theta$ be the set of all such $\theta$. Then, the PMB likelihood for the set of objects $\mathbb{Y}$ can be expressed as
\begin{equation}
\label{eq:pmb_likelihood_propto}
\begin{split}
    f_\text{PMB}(\mathbb{Y})
    &= 
    \sum_{\theta \in \Theta}
    \prod_{j:\theta(j)>0}r_{\theta(j)}p_{\theta(j)}(y_j) 
    \prod_{i:\nexists\theta(j)=i \forall j} 1-r_i 
    \prod_{j:\theta(j)=0}\lambda(y_j)
    \exp\left(-\Bar{\lambda}\right),
    \\
    & \propto 
    \sum_{\theta \in \Theta}
    \prod_{j:\theta(j)>0}r_{\theta(j)}p_{\theta(j)}(y_j) 
    \prod_{i:\nexists\theta(j)=i \forall j} 1-r_i 
    \prod_{j:\theta(j)=0}\lambda(y_j).
\end{split}
\end{equation}
When searching for the most likely associations $\theta$, we disregard $\exp\left(-\Bar{\lambda}\right)=\exp\left(-\int \lambda(y')dy'\right)$ since it is independent of $\theta$.

We note that \eqref{eq:pmb_likelihood_propto} captures the association of every Bernoulli component. If the $i$-th Bernoulli component does not appear in the second product in \eqref{eq:pmb_likelihood_propto}, then it must appear in the first product in \eqref{eq:pmb_likelihood_propto}. We also note that the factor $\prod_{i=1}^m(1-r_i)$ is independent of the association $\theta$. This means that dividing \eqref{eq:pmb_likelihood_propto} by $\prod_{i=1}^m(1-r_i)$ yields
\begin{equation}
\label{eq:pmb_likelihood_propto2}
    f_\text{PMB}(\mathbb{Y}) \propto 
    \sum_{\theta \in \Theta}
    \prod_{j:\theta(j)>0} \frac{r_{\theta(j)}p_{\theta(j)}(y_j)}{1-r_{\theta(j)}}
    \prod_{j:\theta(j)=0}\lambda(y_j),
\end{equation}
and the association that maximizes the PMB likelihood can be found as
\begin{equation}
\label{eq:pmb_likelihood_propto3}
    \theta^* = \argmax_\theta \prod_{j:\theta(j)>0} \frac{r_{\theta(j)}p_{\theta(j)}(y_j)}{1-r_{\theta(j)}}
    \prod_{j:\theta(j)=0}\lambda(y_j).
\end{equation}
Equivalently, we can search for the association that minimises the negative logarithm of \eqref{eq:pmb_likelihood_propto3}, which gives us
\begin{equation}
\label{eq:pmb_likelihood_propto_log}
    \theta^* = \argmin_\theta
    -
    \sum_{j:\theta(j)>0} \log \frac{r_{\theta(j)}p_{\theta(j)}(y_j)}{1-r_{\theta(j)}}
    -
    \sum_{j:\theta(j)=0} \log \lambda(y_j).
\end{equation}

In order to formulate this minimization as an optimal assignment problem, we make a slight change in notation. Each association map $\theta$ can be represented by a $(m+n)\times n$ assignment matrix $A$ consisting of 0s or 1s. There is a bijective mapping between $\theta$ and $A$: for $i=1,\dots,m$, $j=1,\dots,n$, $A_{i,j}=1$ if and only if $\theta(j)=i$. For $i=m+j$, $j=1,\dots,n$, $A_{i,j}=1$ if and only if $\theta(j)=0$, whereas $A_{i,j}$ is always $0$ when $i>m$ and $i \neq m+j$. The assignment matrix hence must satisfy the constraints  $\sum_i A_{i,j} = 1, \, \forall j$, and $\sum_j A_{i,j} \leq 1, \, \forall i$.

The cost matrix of the corresponding assignment matrix $A$ is the $(m+n)\times n$ matrix $C$ where
\begin{equation}
    C_{i,j} = - \log \frac{r_i p_i (y_j)}{1-r_i}, \ i=1,\dots,m, \ j=1,\dots,n,
\end{equation}
and where the lower part of $C$ is a ``diagonal'' matrix such that $C_{m+j,j}=-\log \lambda(y_j)$, $j=1,\dots,n$, and entries not on the diagonal are $\infty$. The cost of assignment matrix $A$ is then given by the Frobenius inner product

\begin{equation}
    \label{eq:frobenius_ip}
    \text{trace}(A^TC)=\sum_{i=1}^{m+n} \sum_{j=1}^{n} C_{i,j} A_{i,j},
\end{equation}
and the problem of finding the optimal assignment $A^*$ becomes
\begin{subequations}
\label{eq:assignment_problem-appendix}
\begin{align}
    A^* = \argmin_A & \sum_{i=1}^{m+n} \sum_{j=1}^{n} C_{i, j}A_{i,j}
    \\
    \label{eq:assignment_problem_constraints-appendix}
    \text{s.t.} \quad & \sum_{i=1}^{m+n} A_{i,j}=1,\, \sum_{j=1}^{n} A_{i,j}\leq 1,
    \\
    \label{eq:cost_matrix_mb-appendix}
    & C_{i, j} = \begin{cases} 
    -\log\left(\frac{p_i(y_j)}{1-r_i}r_i\right) &\text{if }i\leq m,
    \\
    -\log\lambda(y_j) & \text{if } i=m+j,
    \\
    \infty & \text{otherwise.}
    \end{cases}
\end{align}
\end{subequations}

\section{Experimental Details}

\subsection{Model Implementation}
For implementing the DETR, RetinaNet and Faster-RCNN models, we used the probabilistic extension \cite{harakeh2021estimating} of the Detectron2 \cite{wu2019detectron2} object detection framework. In that framework, models are trained to predict the covariance matrix $\bm{\Sigma}_b$ for the corresponding bounding box $b$. Specifically, models output the parameters of a lower triangular matrix $\bm{L}$ of the Cholesky decomposition $\bm{\Sigma}_b=\bm{L}\bm{L}^T$. While originally trained with a Gaussian distribution, we found that using an independent Laplace distribution for each parameter in $b$ yielded better results. Using the diagonal elements of $\bm{L}$ as $[\sigma_1,\sigma_2,\sigma_3,\sigma_4]$, we find the scale of each Laplace distribution as $s_i = \sigma_i/\sqrt{2}$. The choice of a diagonal matrix is partially motivated by the evaluations in \cite{harakeh2021estimating}. While their study was limited to Gaussian distributions, they found that diagonal covariance matrices perform on par with, or better than, their full equivalent.

\subsection{Training Details}
Fine-tuning toward MB-NLL was done given the pre-trained weights in \cite{harakeh2021estimating}\footnote{\url{https://github.com/asharakeh/probdet}}. For DETR and Faster-RCNN, models trained with ES were used as a starting point, while the model trained with NLL was used for RetinaNet. Faster-RCNN and RetinaNet were fine-tuned for 135,000 gradient steps, where the learning rate was dropped by a factor of 10 at 105,000 iterations, and again at 125,000 iterations. The initial learning rate was set to 0.001 for RetinaNet and 0.0025 for Faster-RCNN. DETR was also fine-tuned for 135,000 iterations, but with an initial learning rate of $5\cdot 10^{-5}$ and with learning rate drops at 60,000 and 100,000 iterations. Otherwise, no changes to hyperparameters from the standard Detectron2 framework were done. 

As both RetinaNet and Faster-RCNN rely on non-maximum suppression to remove duplicate detections, we applied NMS when training with the MB-NLL loss. Here, we used the standard IoU threshold of 0.5 and used the top 100 detections. For DETR, this was not necessary since it predicts the set of objects directly.

\subsection{Inference Details}
Following the COCO standard, detectors are limited to 100 predictions per image. We do not apply any confidence thresholding, but for RetinaNet and Faster-RCNN the 100 predictions with highest existence probability after NMS are used. For DETR, no selection is needed as the model only produces 100 predictions per image. 

\section{Additional Results}

\subsection{Qualitative Results}
\begin{figure}[thp]
\centering
\begin{subfigure}{\textwidth}
  \centering
  \includegraphics[width=4cm]{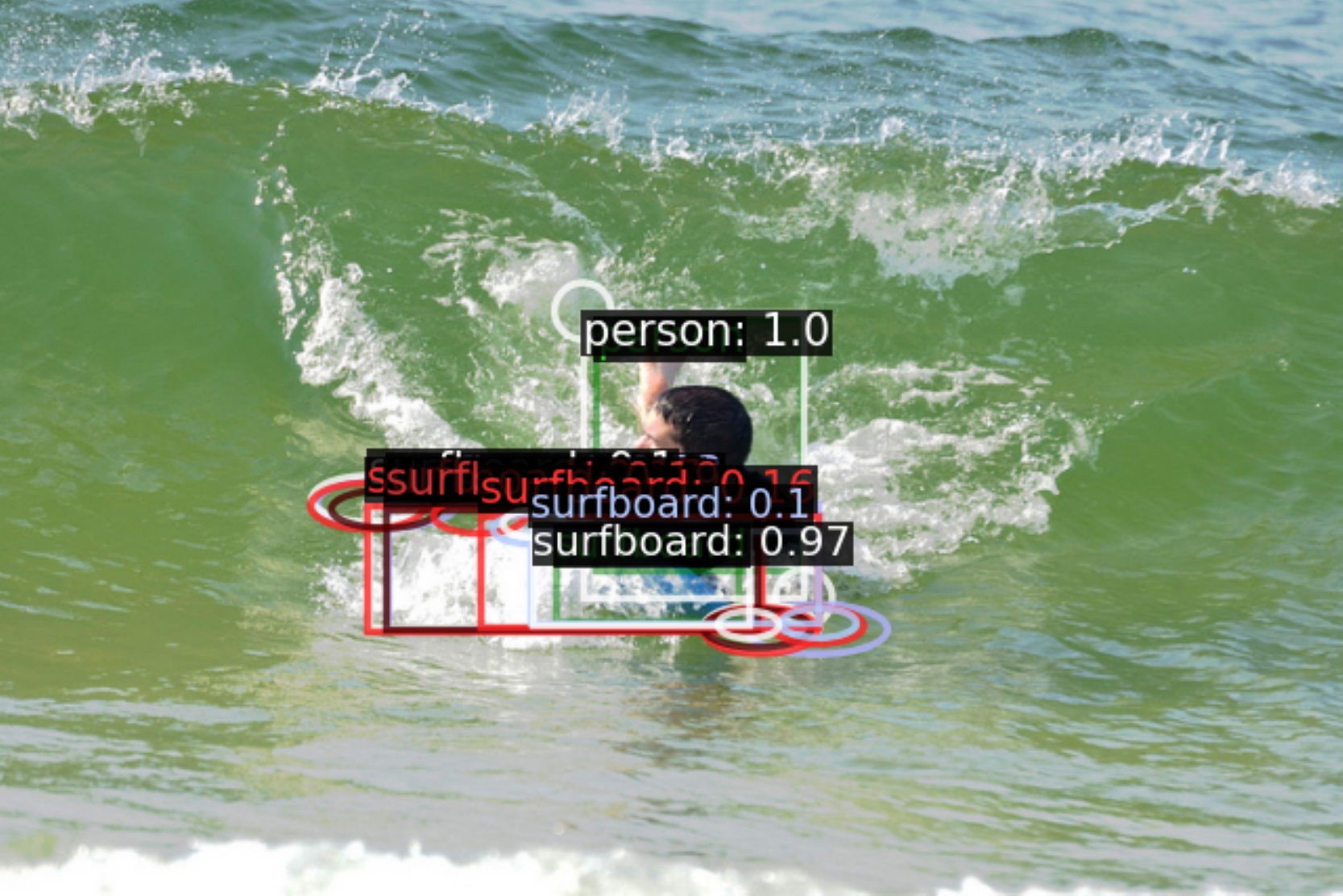}
  \includegraphics[width=4cm]{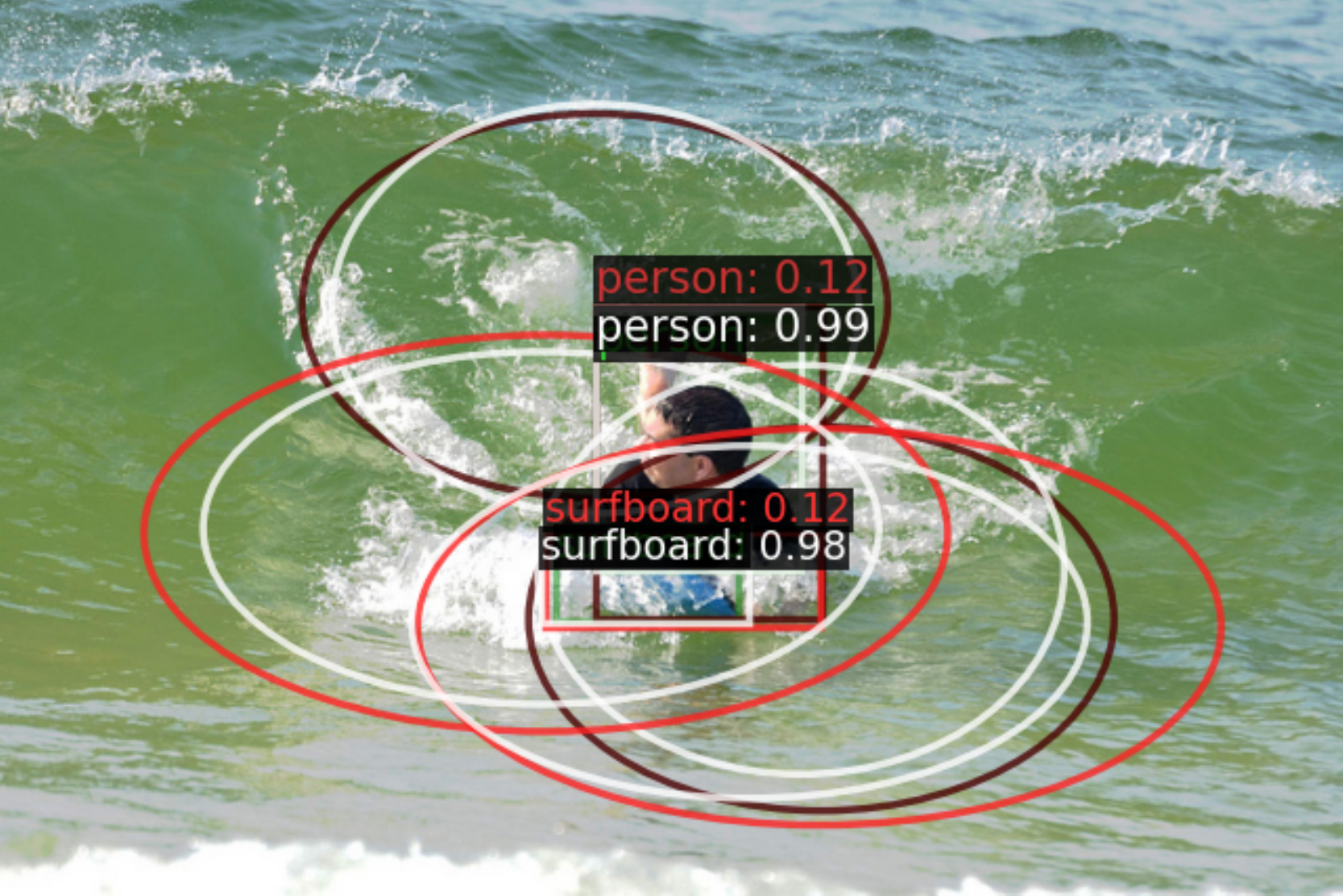}
  \includegraphics[width=4cm]{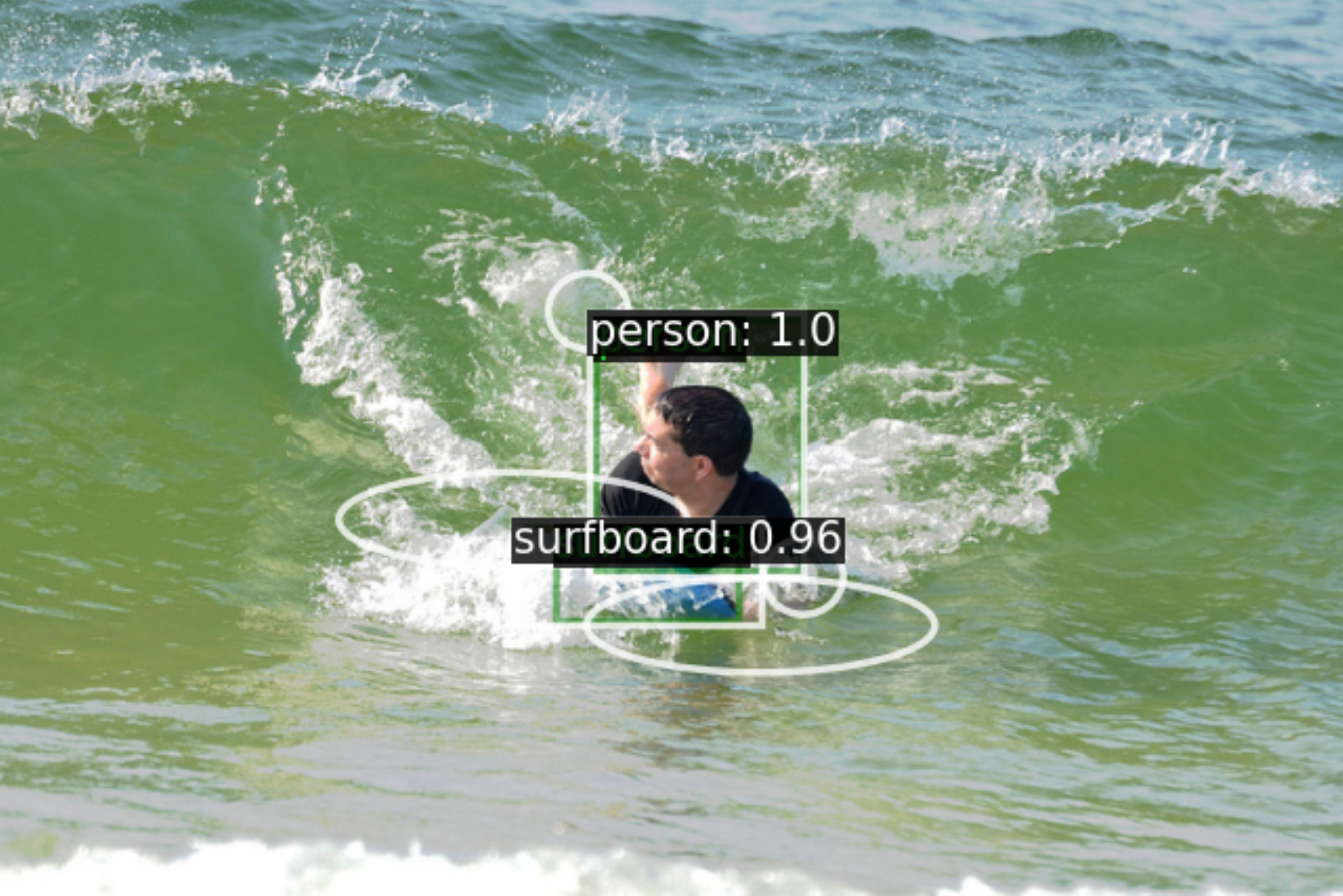}
  \caption{Example 1.}
  \label{fig:detr-example-1}
\end{subfigure}
\newline
\begin{subfigure}{\textwidth}
  \centering
  \includegraphics[width=4cm]{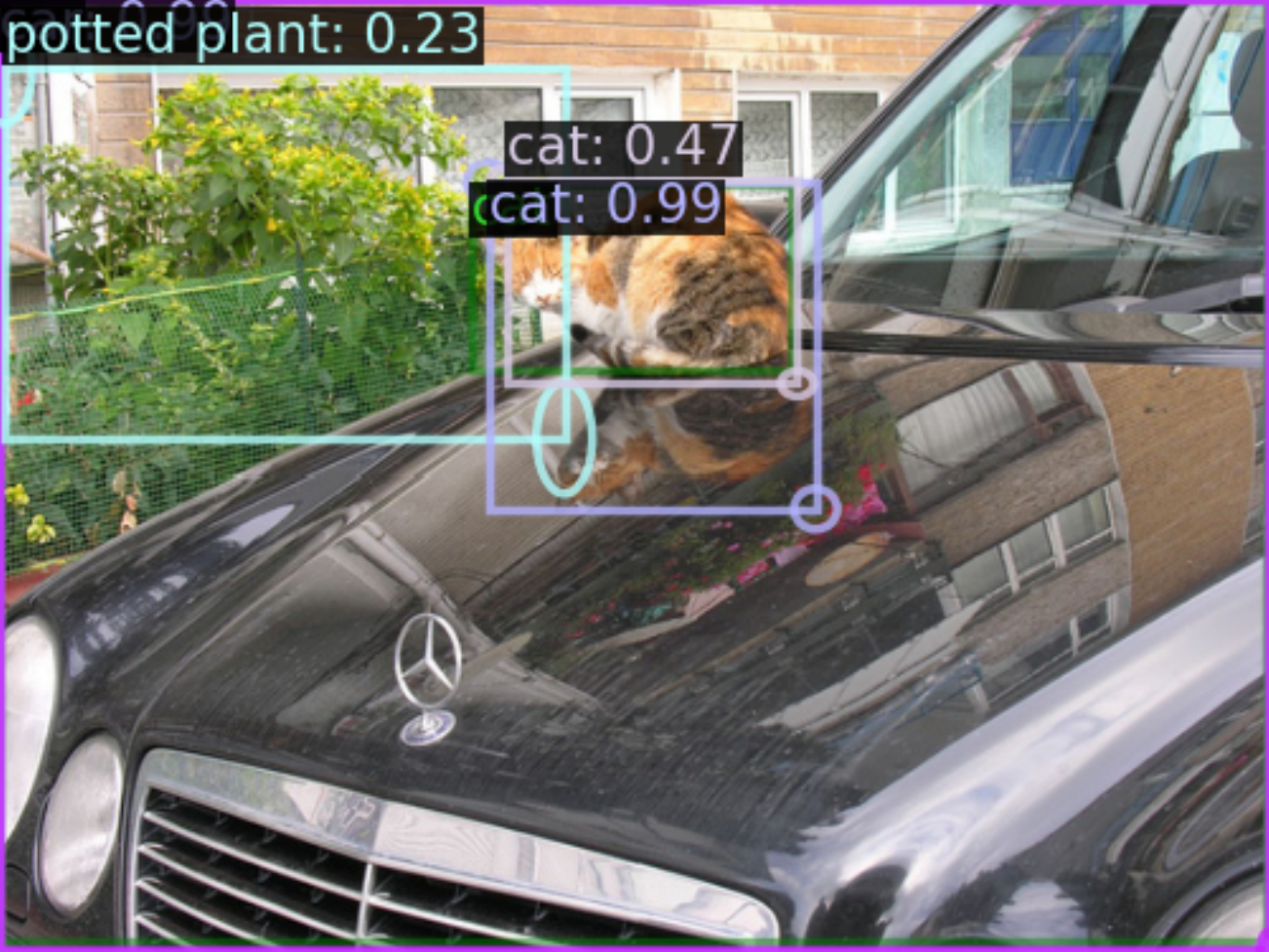}
  \includegraphics[width=4cm]{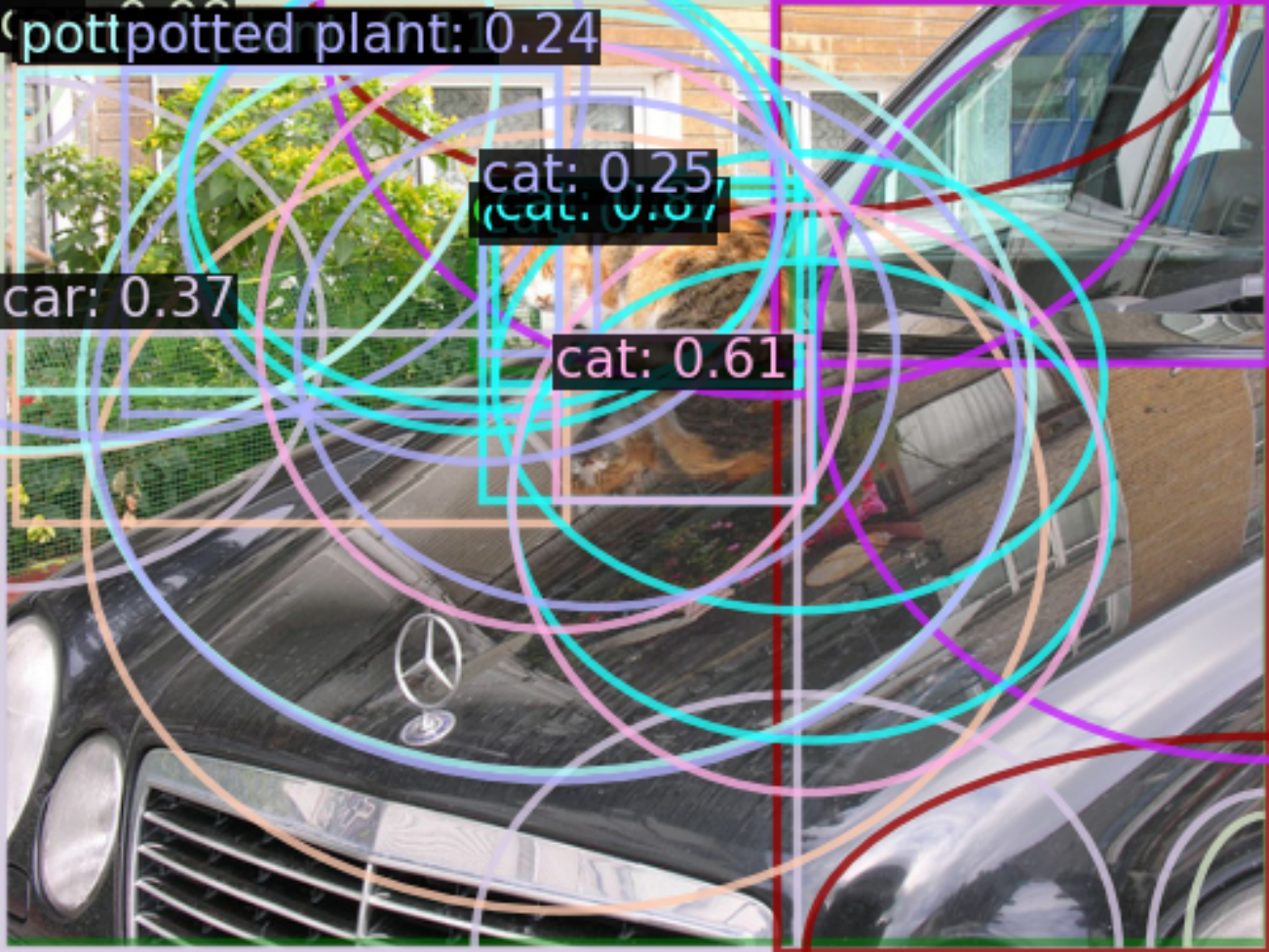}
  \includegraphics[width=4cm]{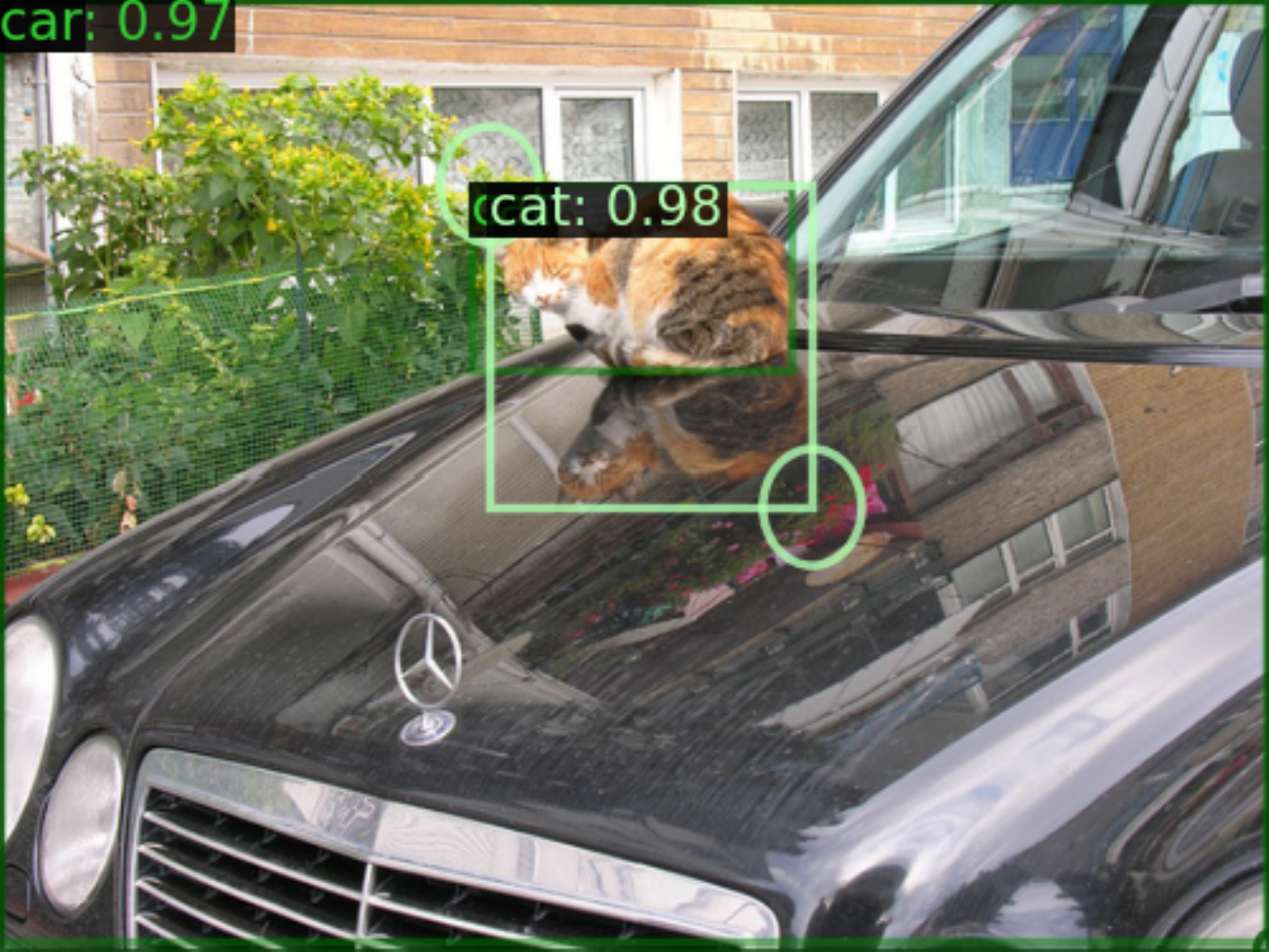}
  \caption{Example 2.}
  \label{fig:detr-example-2}
\end{subfigure}
\newline
\begin{subfigure}{\textwidth}
  \centering
  \includegraphics[width=4cm]{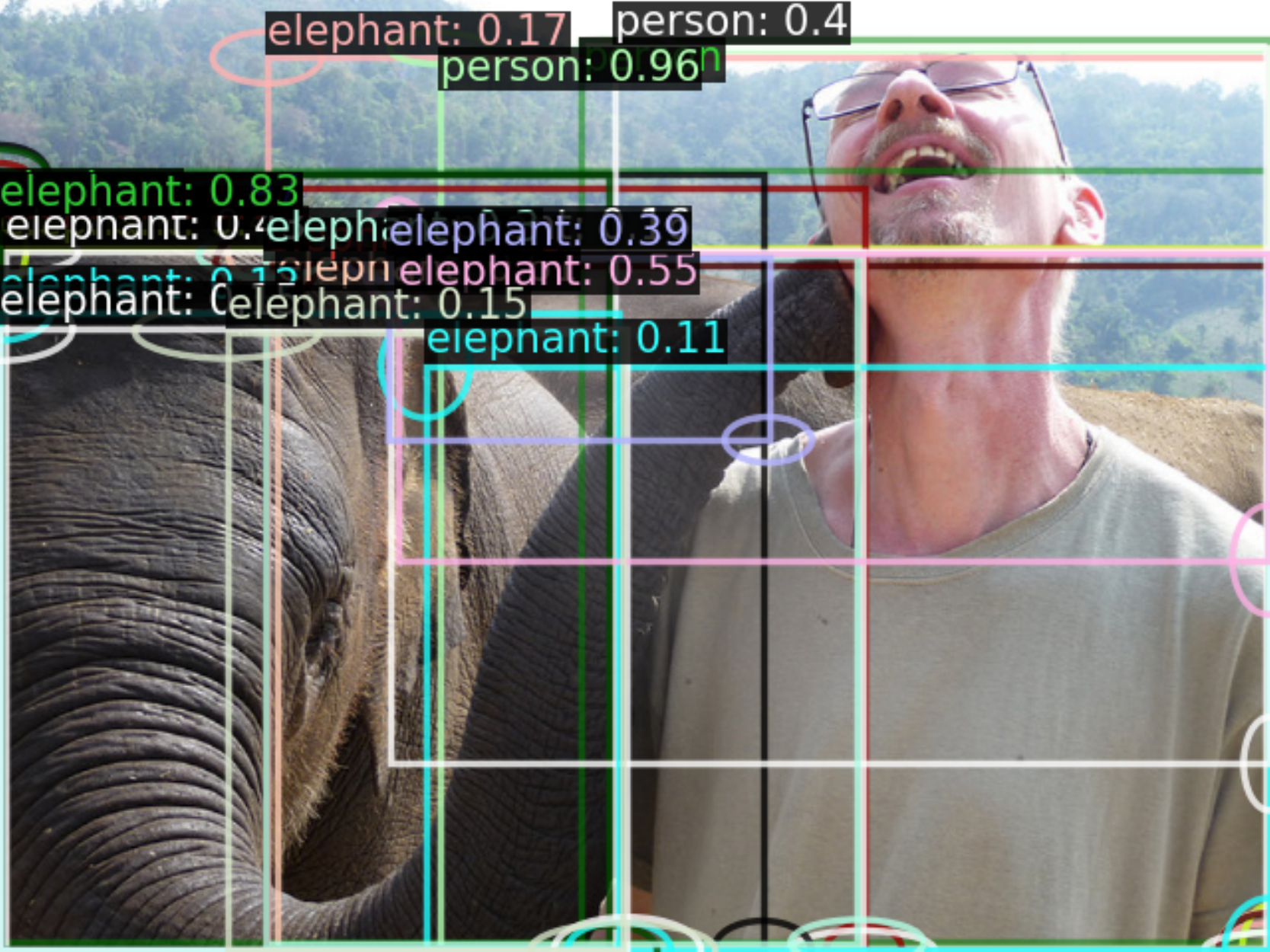}
  \includegraphics[width=4cm]{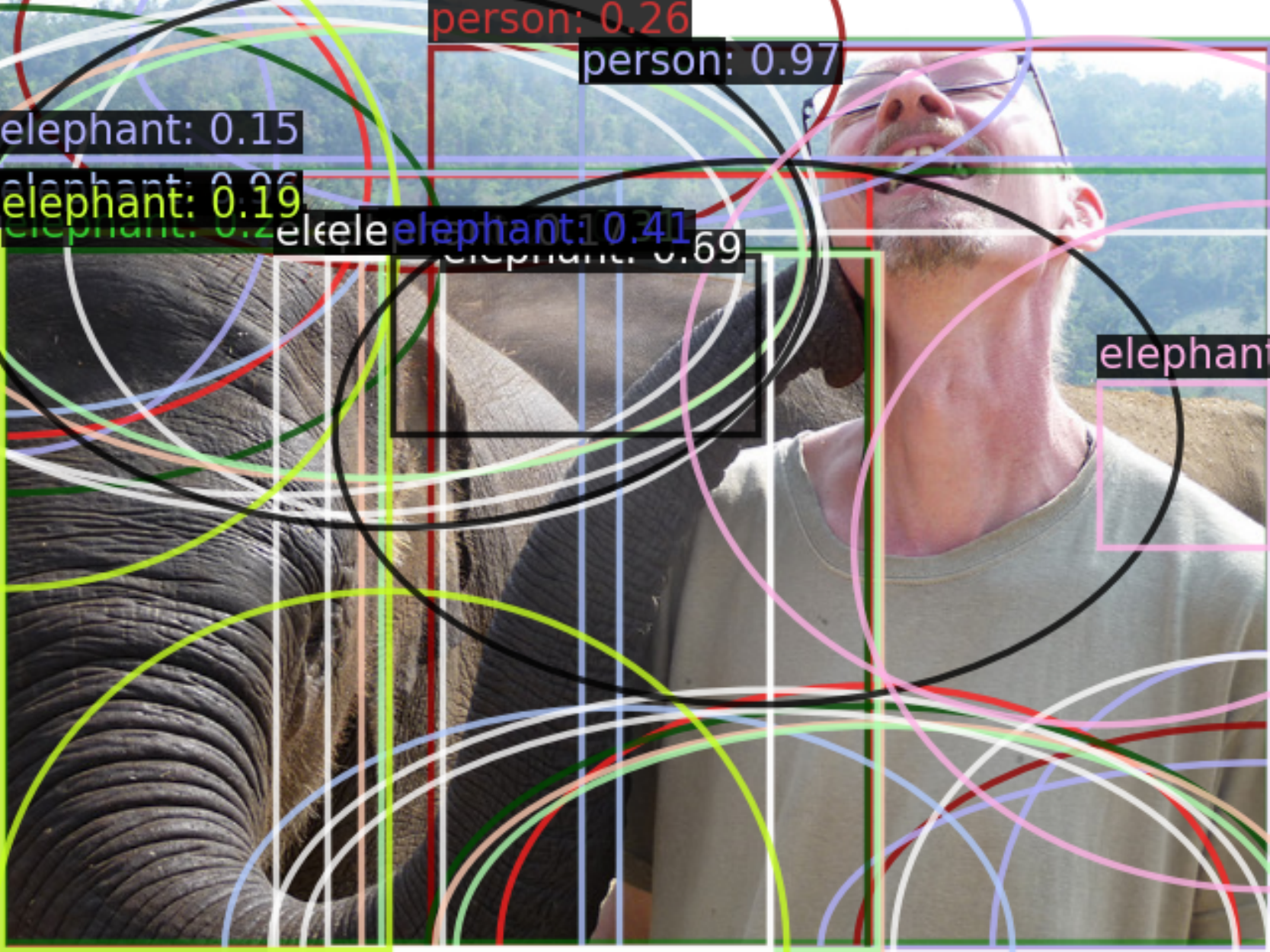}
  \includegraphics[width=4cm]{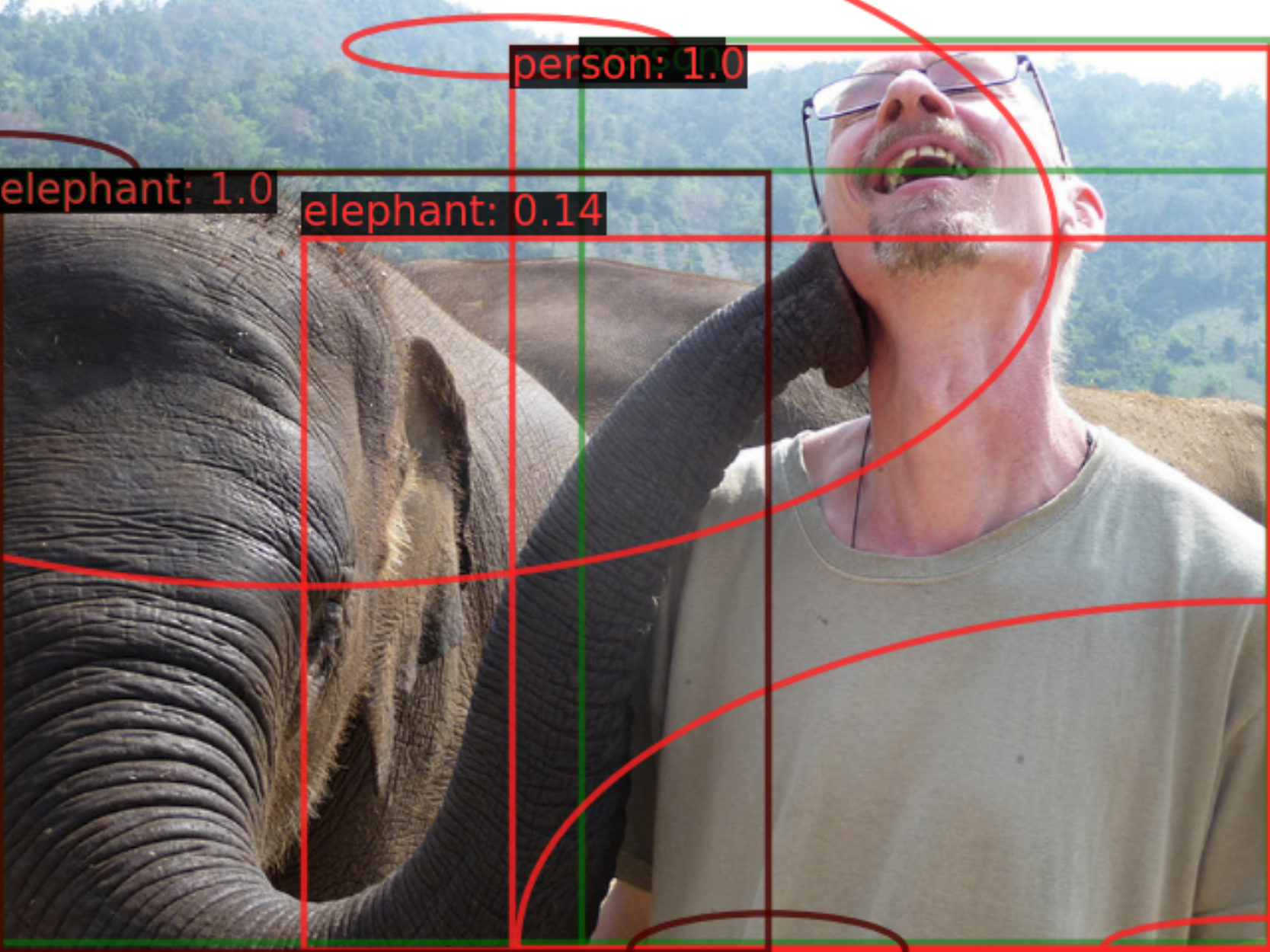}
  \caption{Example 3.}
  \label{fig:detr-example-3}
\end{subfigure}
\caption{Examples from COCO validation data with predictions made by DETR detectors trained with ES (left), NLL (middle), and MB-NLL (right). True objects are shown in green and without confidence values. Predictions with $r<0.1$ are not shown.}
\label{fig:detr-examples}
\end{figure}

\begin{figure}[thp]
\centering
\begin{subfigure}{\textwidth}
  \centering
  \includegraphics[width=4cm]{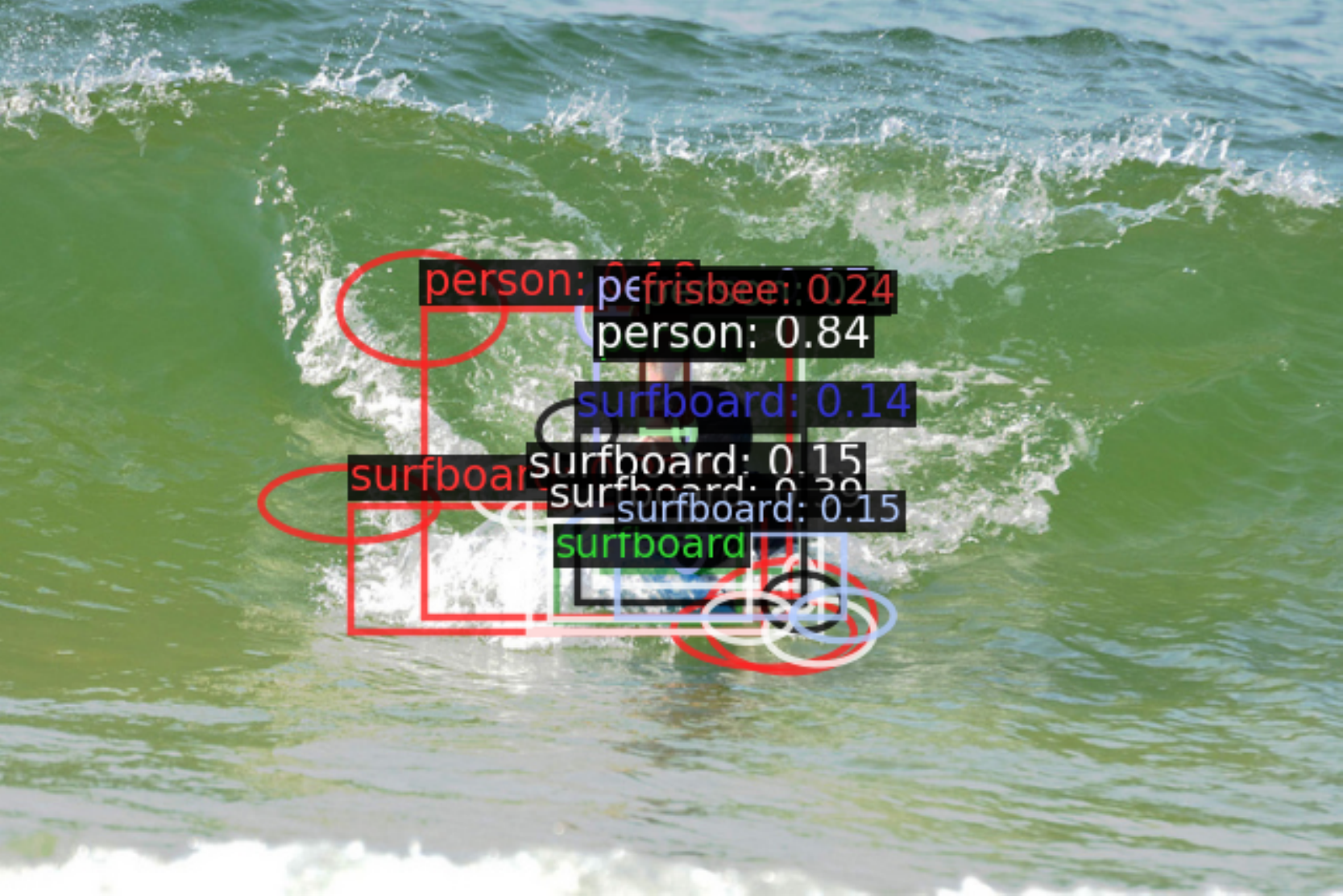}
  \includegraphics[width=4cm]{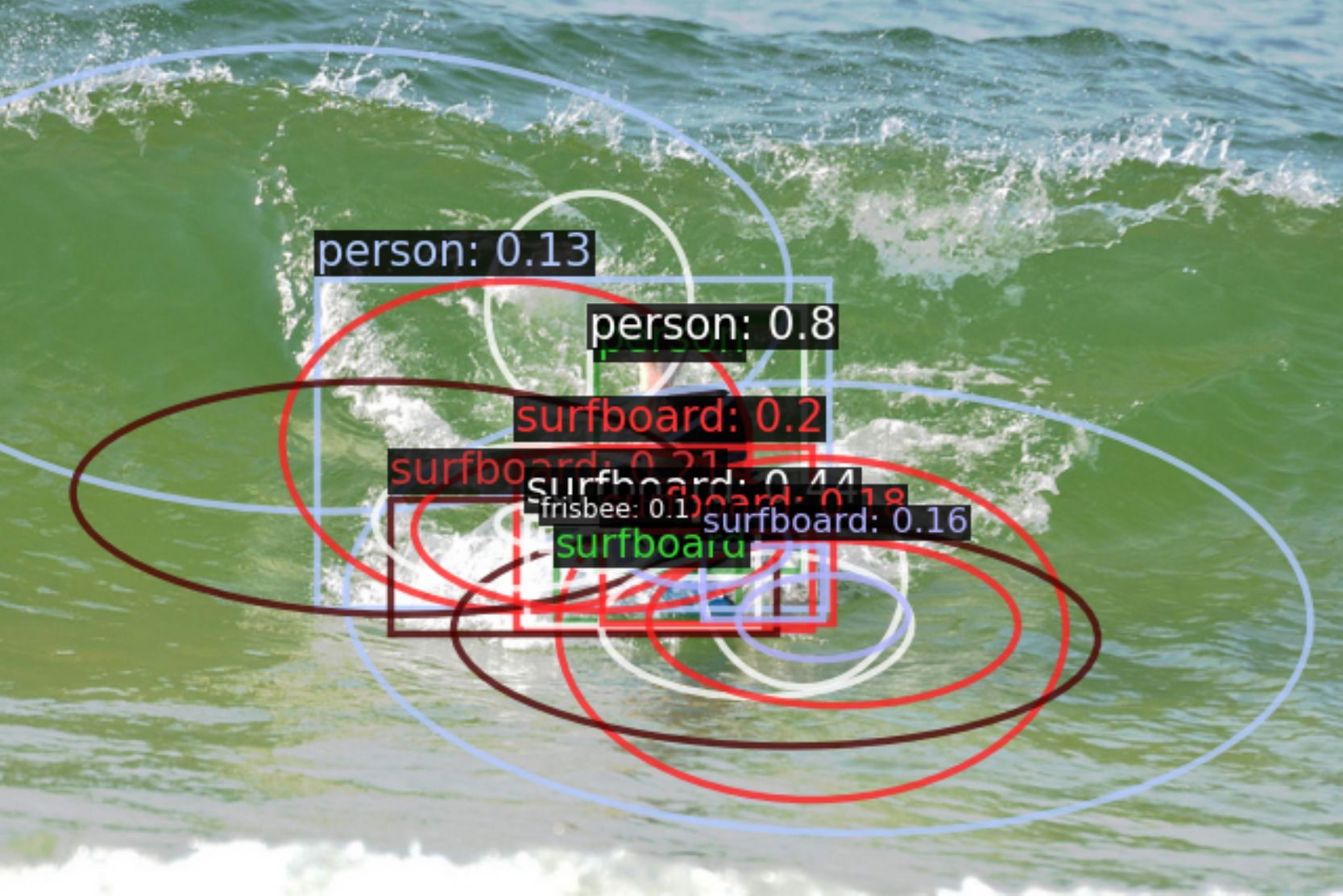}
  \includegraphics[width=4cm]{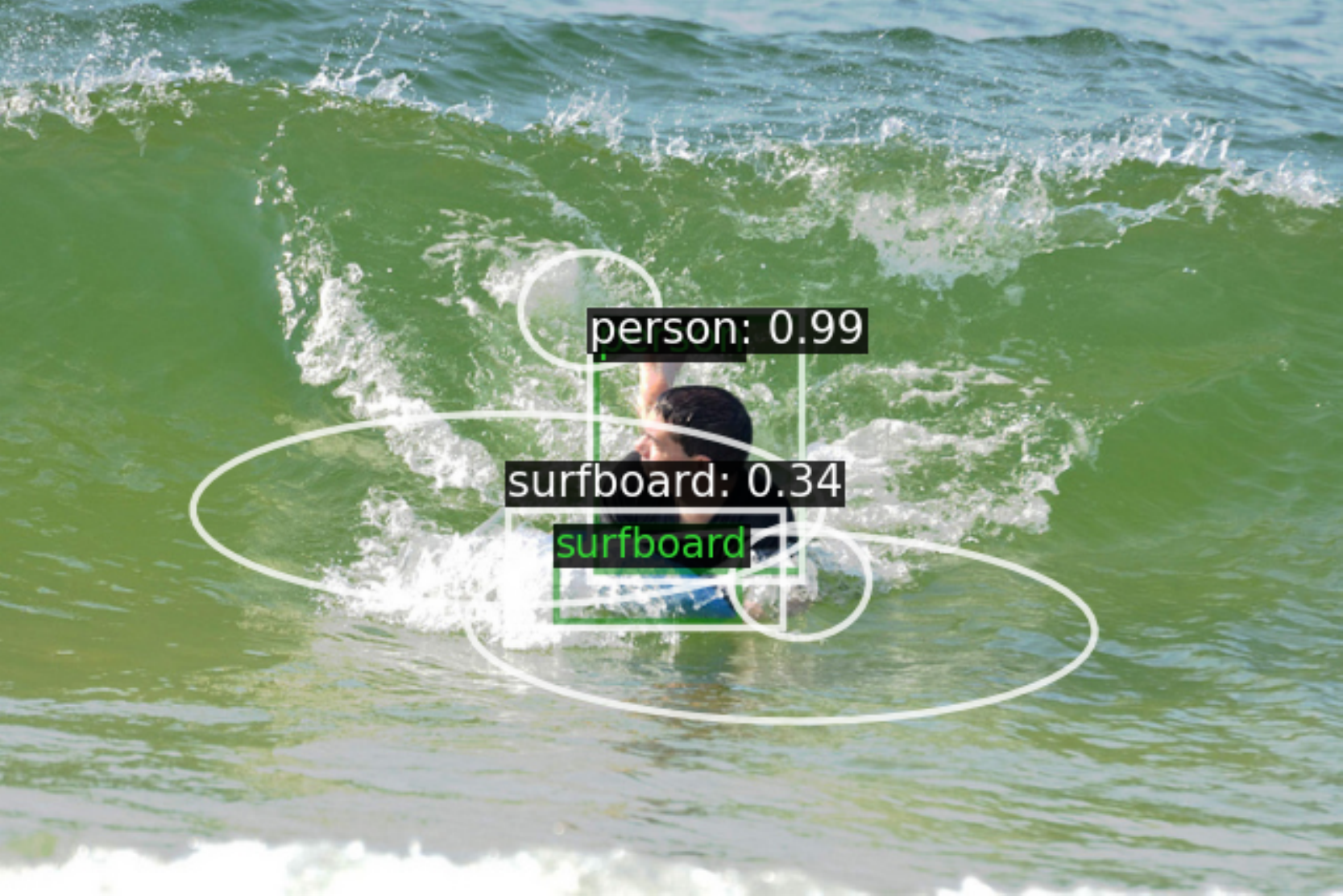}
  \caption{Example 1.}
  \label{fig:retinanet-example-1}
\end{subfigure}
\newline
\begin{subfigure}{\textwidth}
  \centering
  \includegraphics[width=4cm]{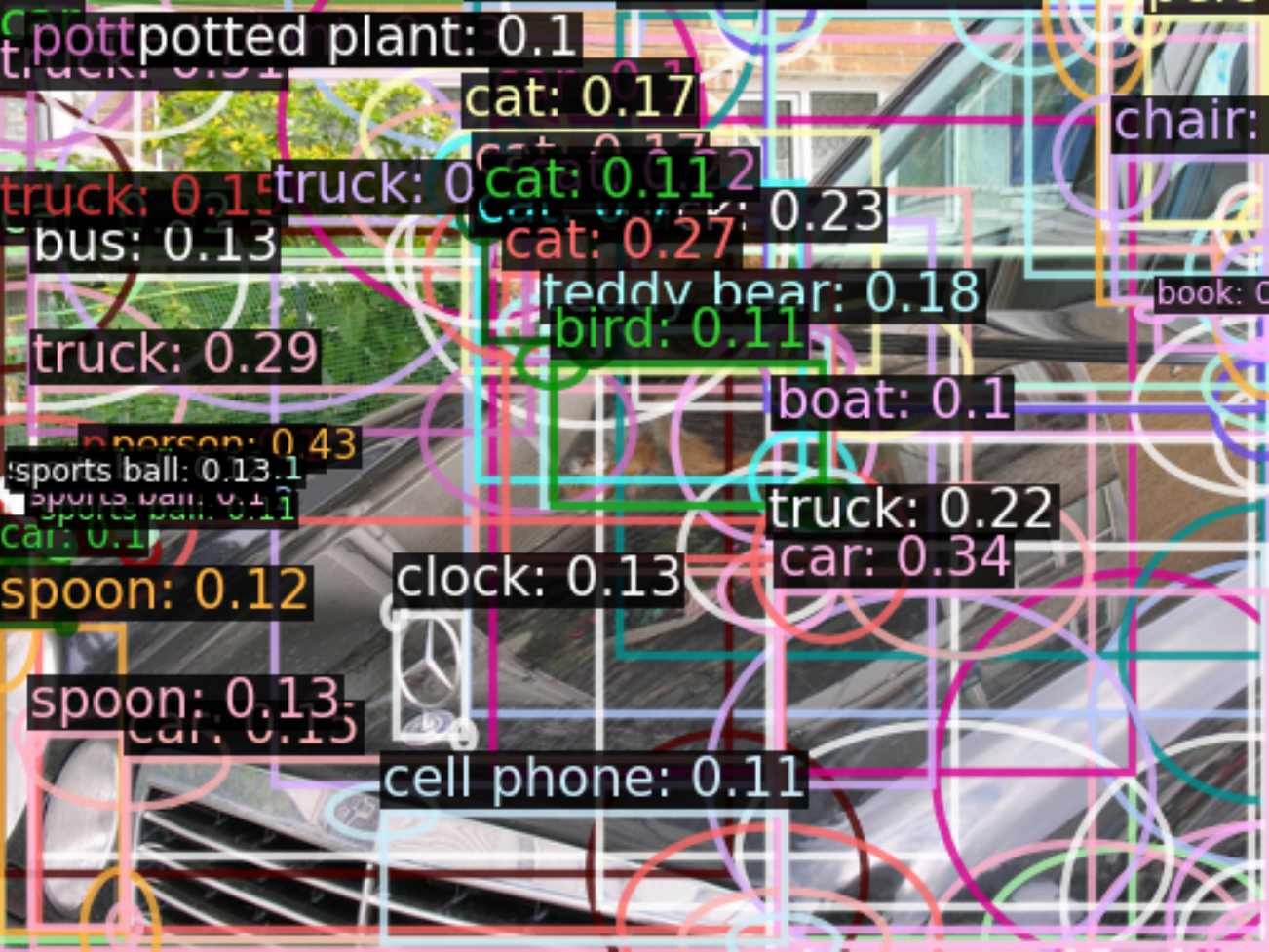}
  \includegraphics[width=4cm]{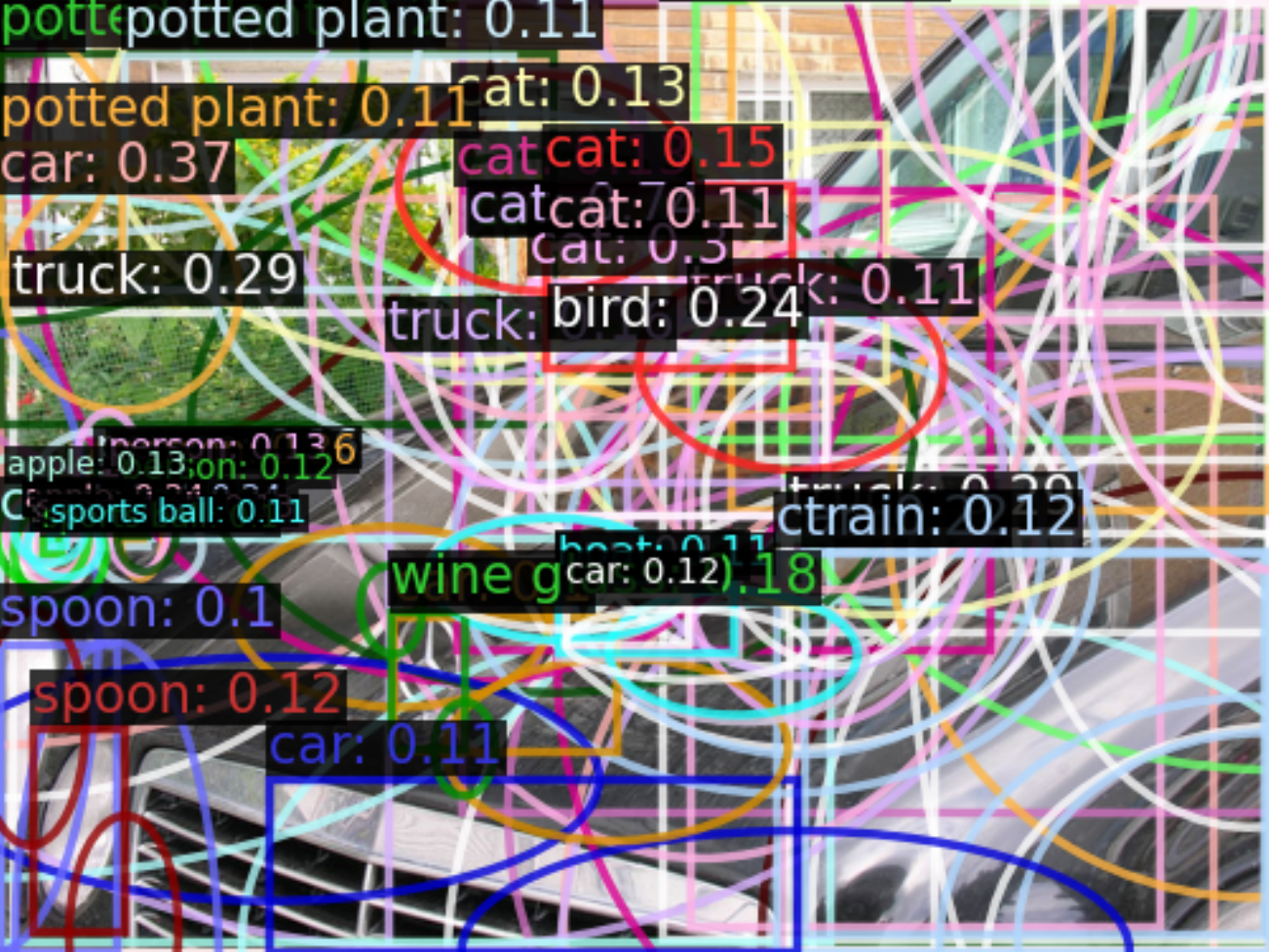}
  \includegraphics[width=4cm]{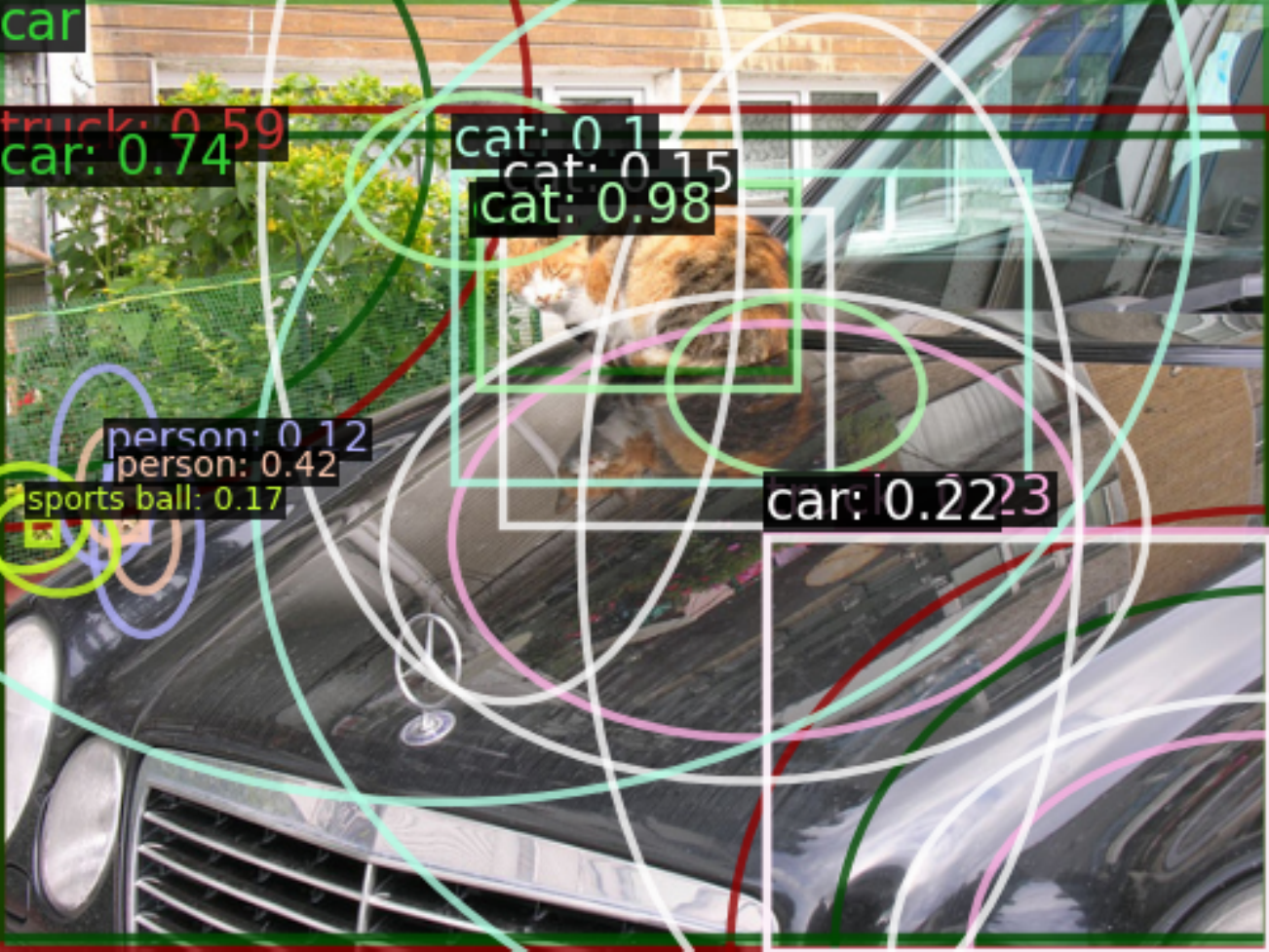}
  \caption{Example 2.}
  \label{fig:retinanet-example-2}
\end{subfigure}
\newline
\begin{subfigure}{\textwidth}
  \centering
  \includegraphics[width=4cm]{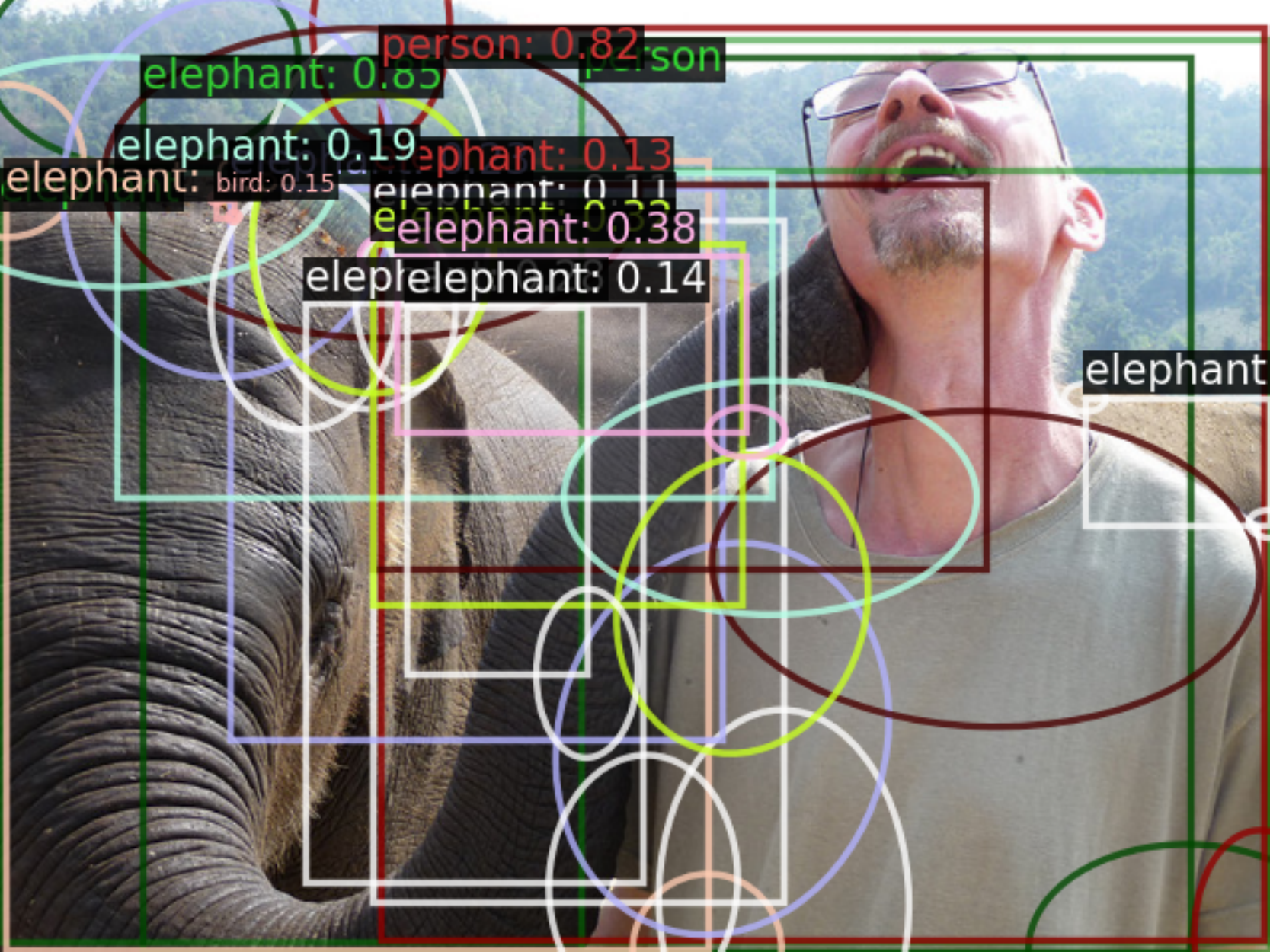}
  \includegraphics[width=4cm]{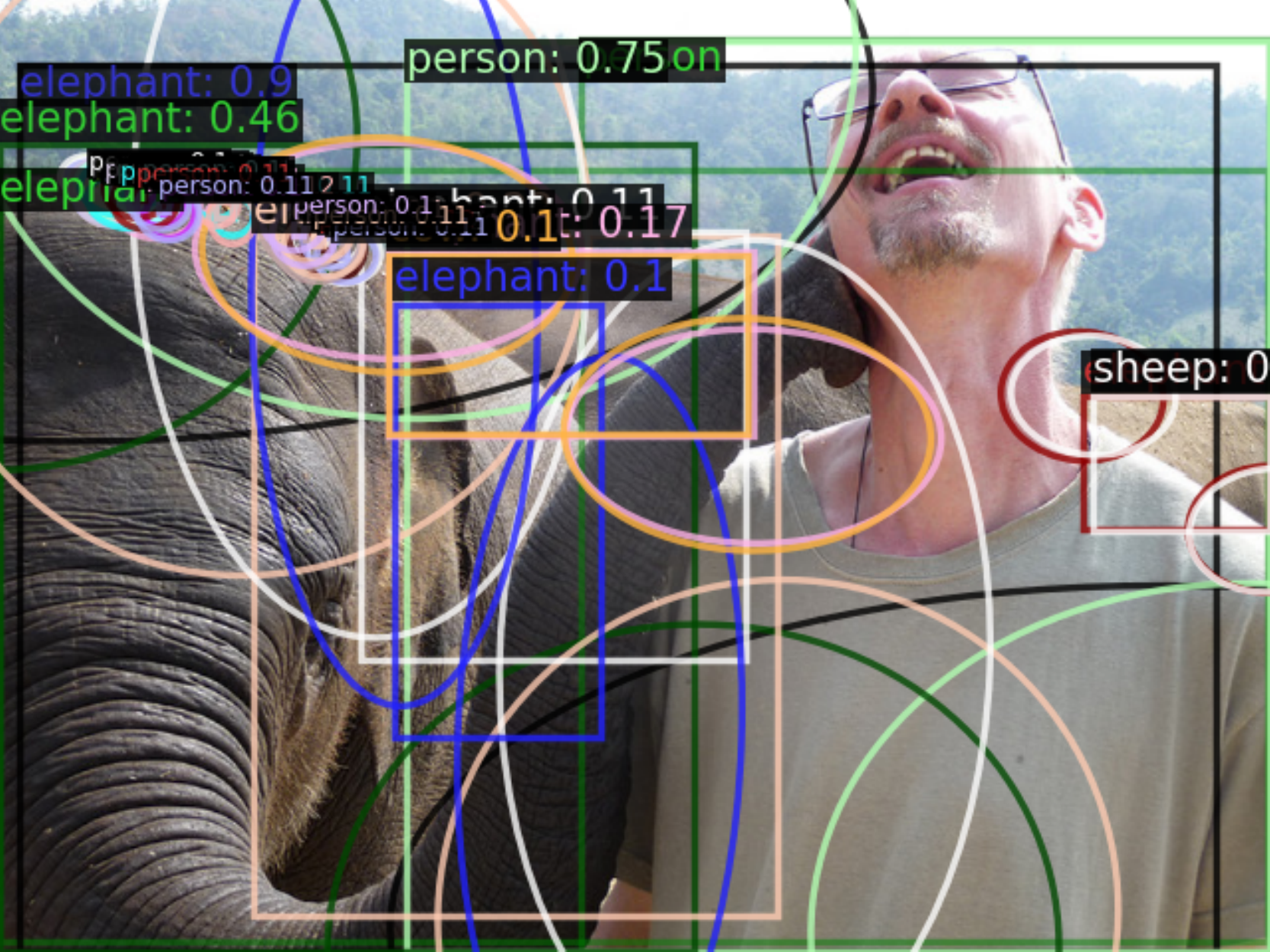}
  \includegraphics[width=4cm]{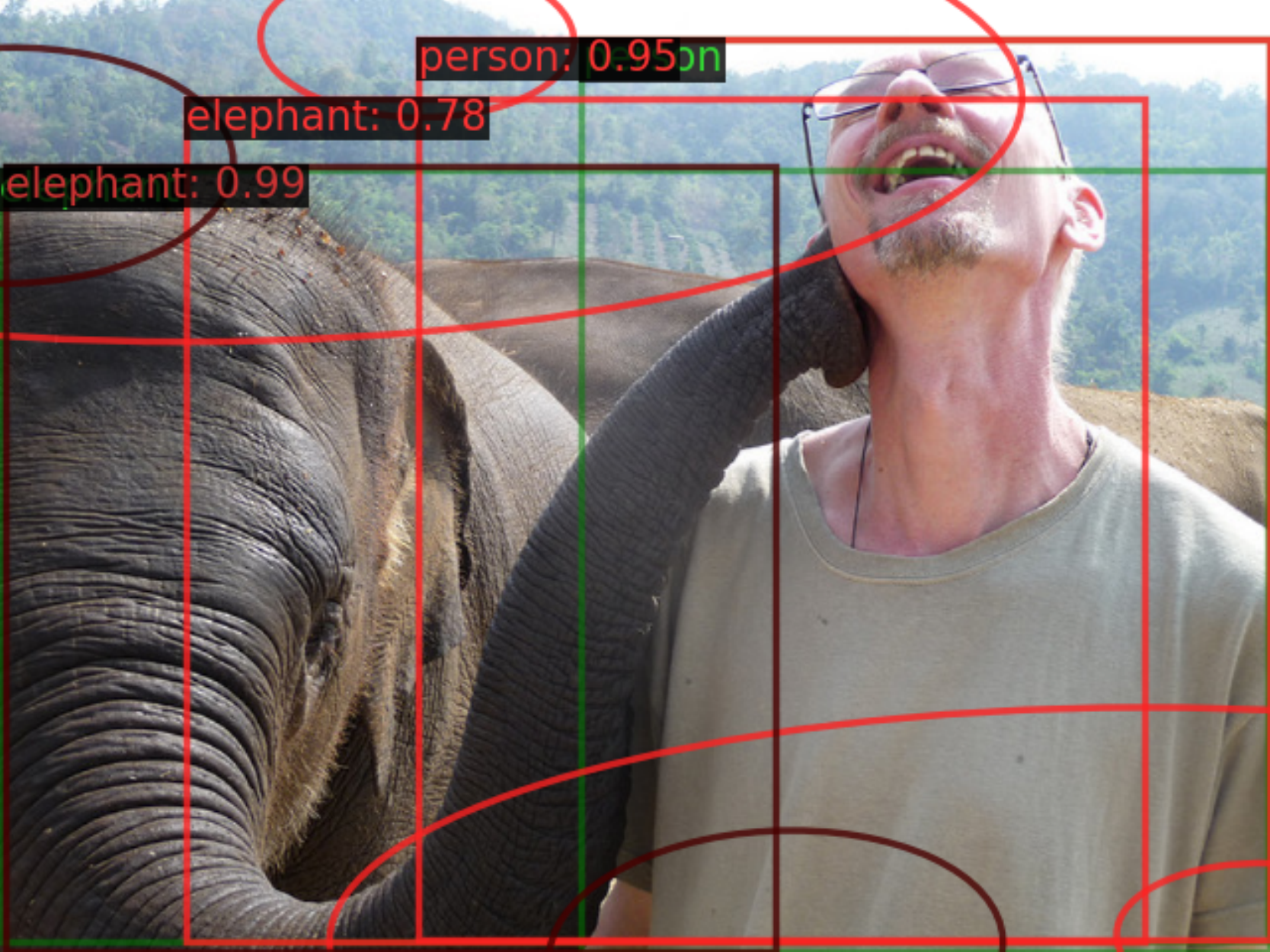}
  \caption{Example 3.}
  \label{fig:retinanet-example-3}
\end{subfigure}
\caption{Examples from COCO validation data with predictions made by RetinaNet detectors trained with ES (left), NLL (middle), and MB-NLL (right). True objects are shown in green and without confidence values. Predictions with $r<0.1$ are not shown.}
\label{fig:retinanet-examples}
\end{figure}

\begin{figure}[thp]
\centering
\begin{subfigure}{\textwidth}
  \centering
  \includegraphics[width=4cm]{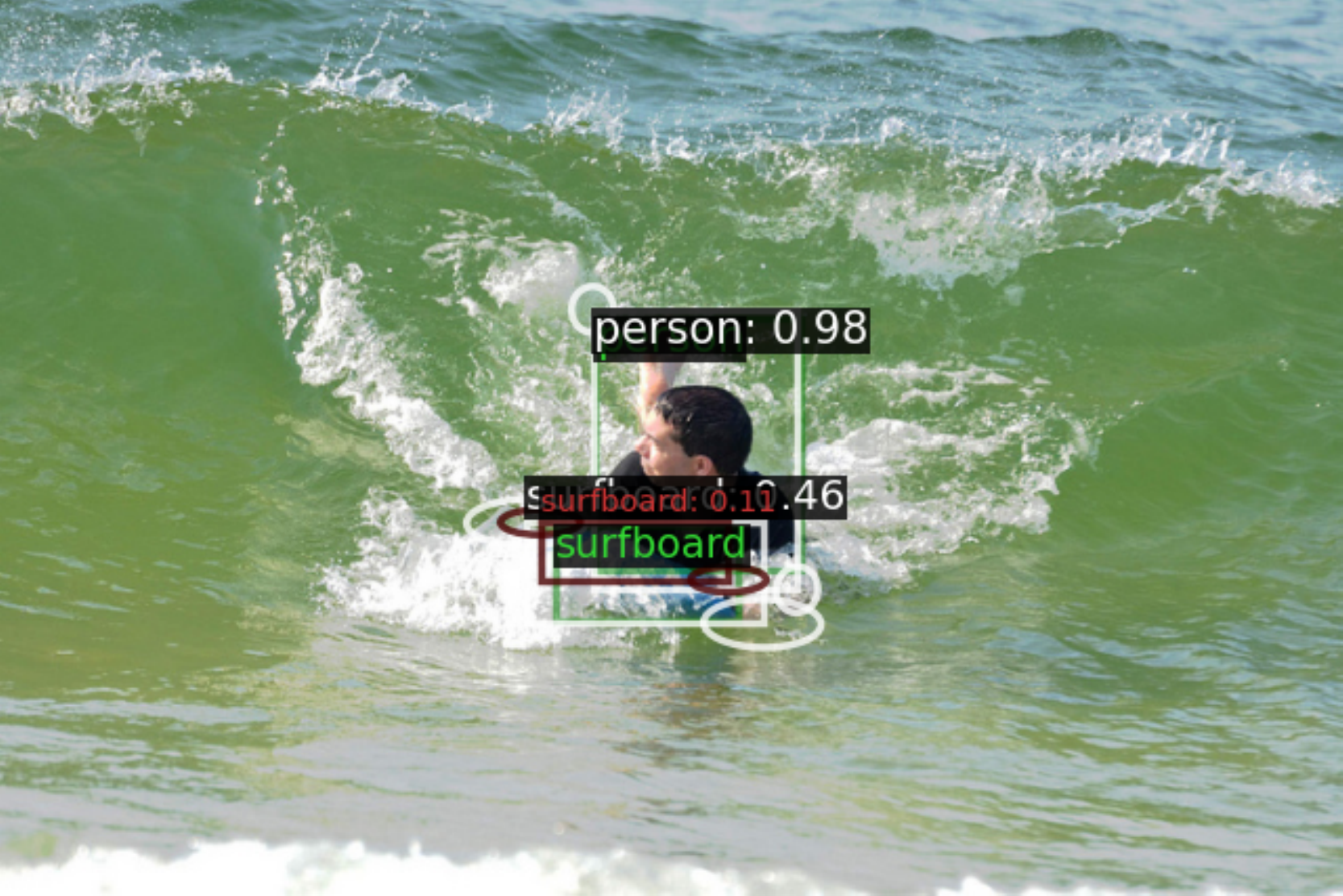}
  \includegraphics[width=4cm]{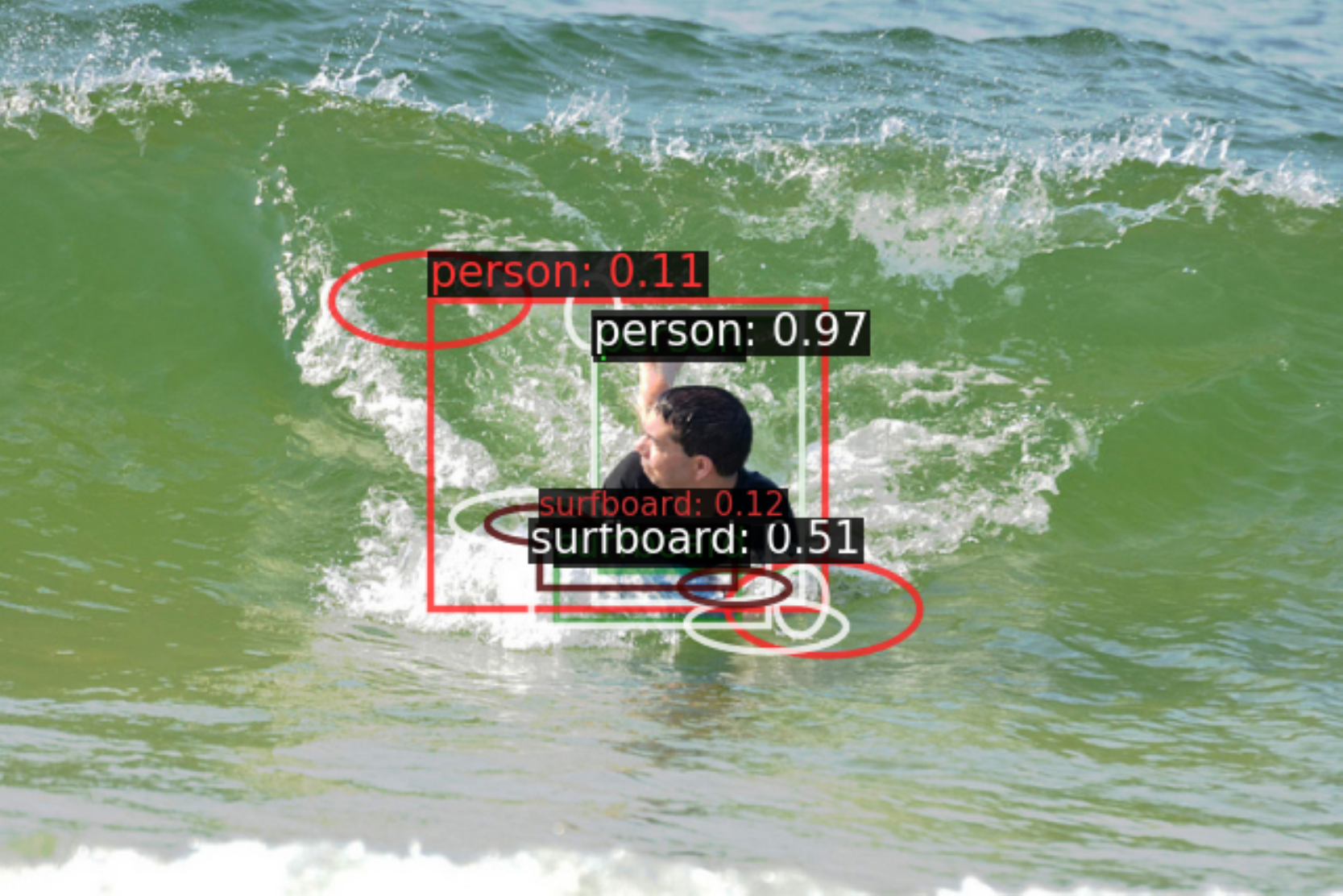}
  \includegraphics[width=4cm]{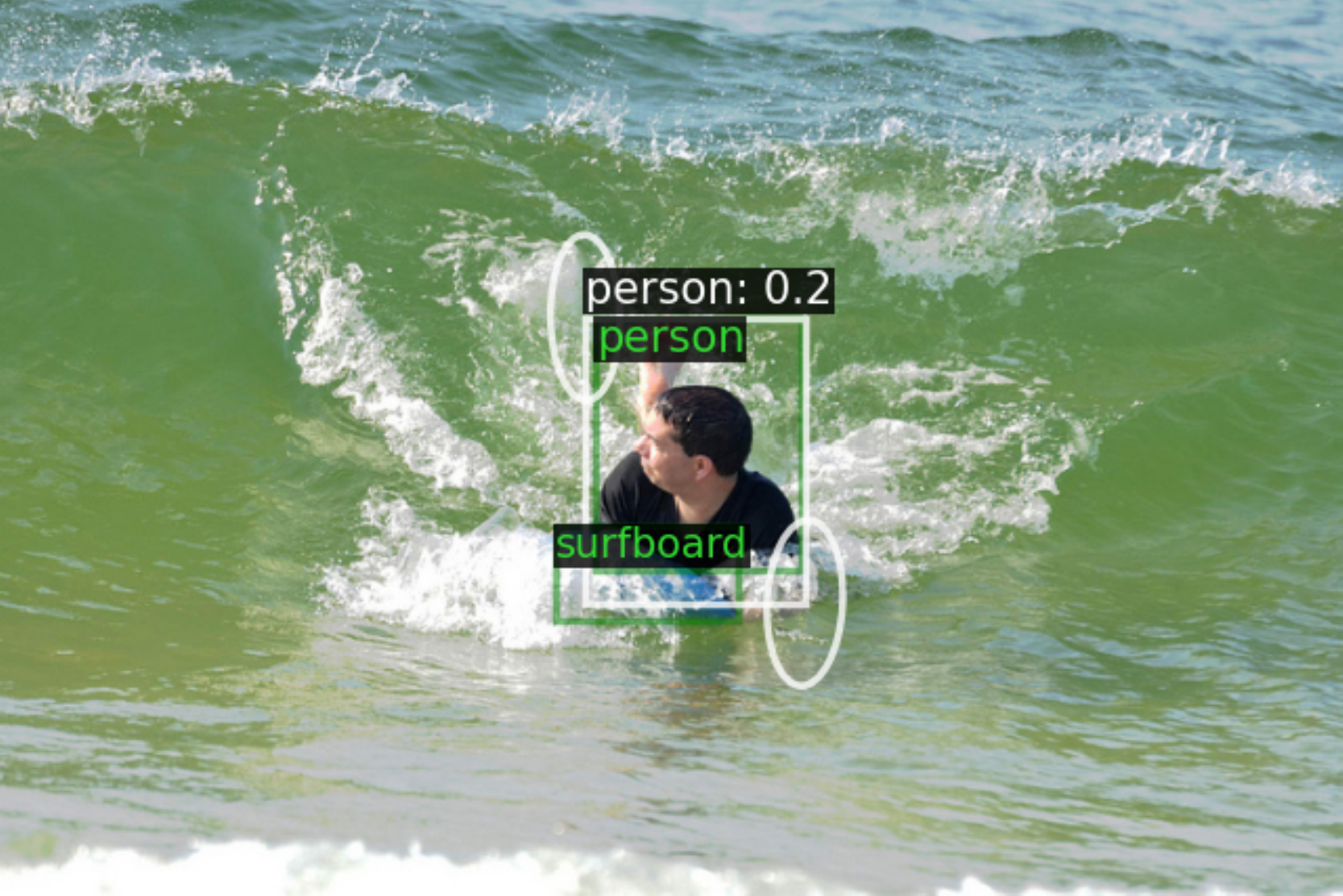}
  \caption{Example 1.}
  \label{fig:faster-rcnn-example-1}
\end{subfigure}
\newline
\begin{subfigure}{\textwidth}
  \centering
  \includegraphics[width=4cm]{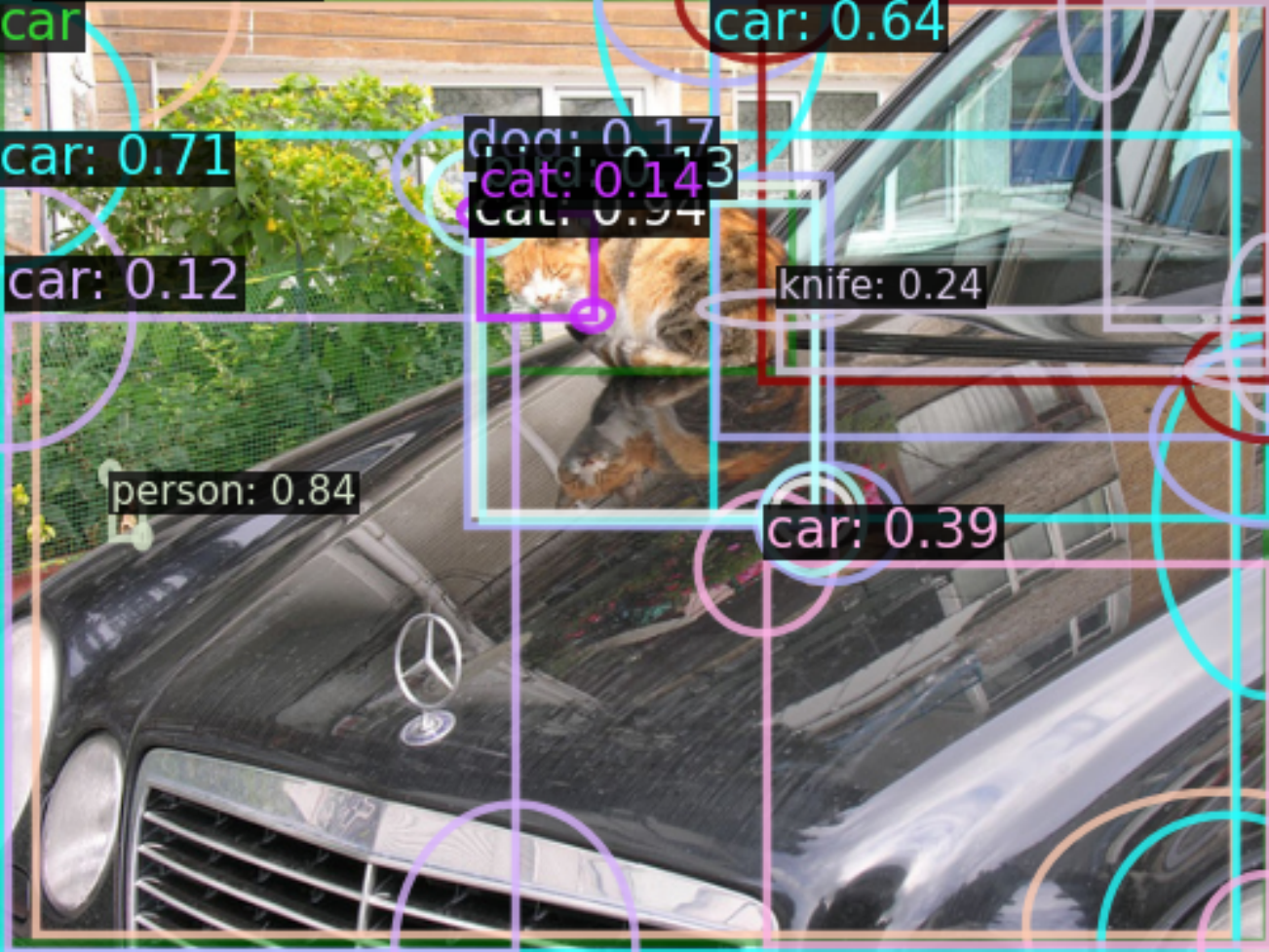}
  \includegraphics[width=4cm]{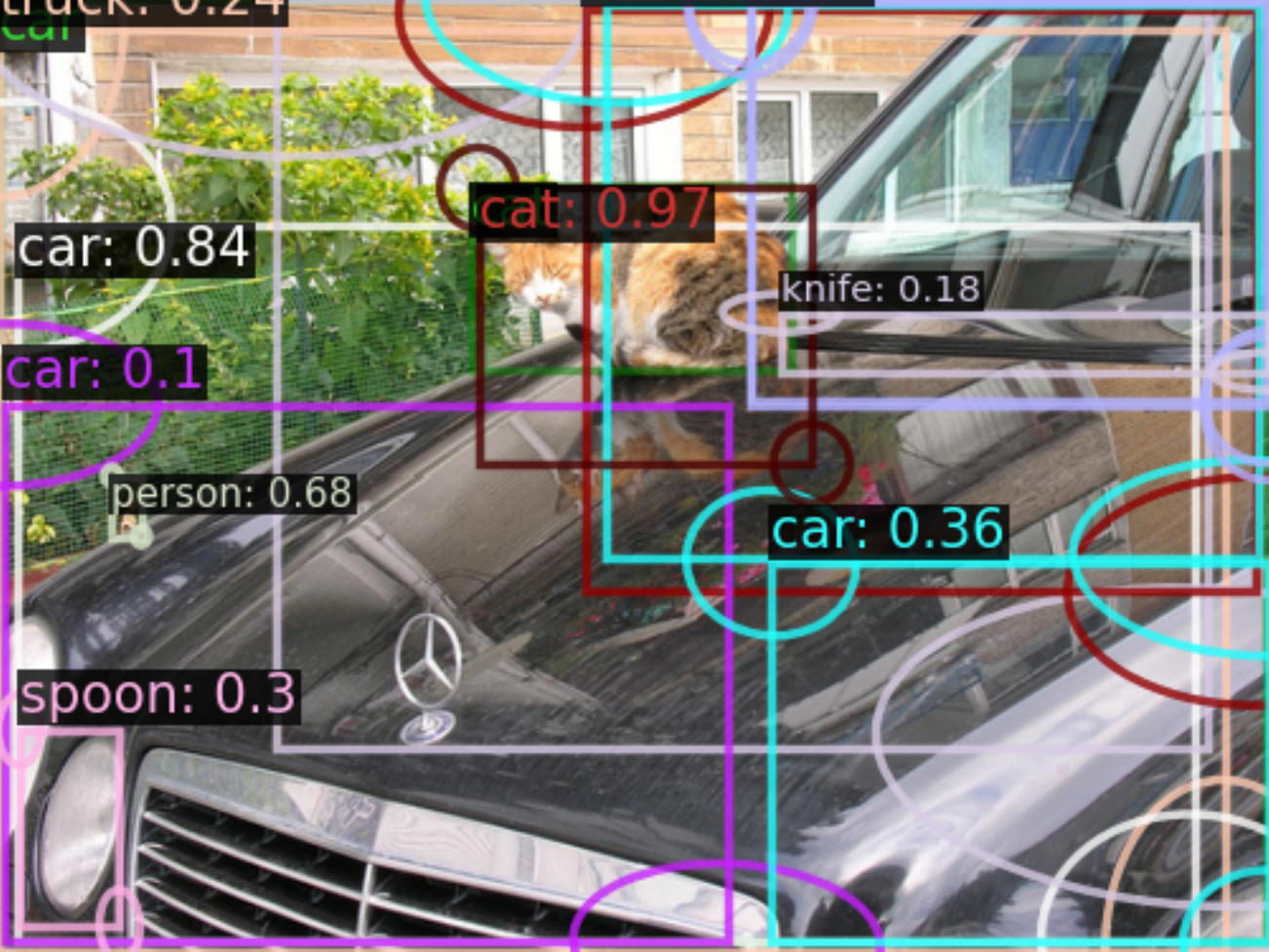}
  \includegraphics[width=4cm]{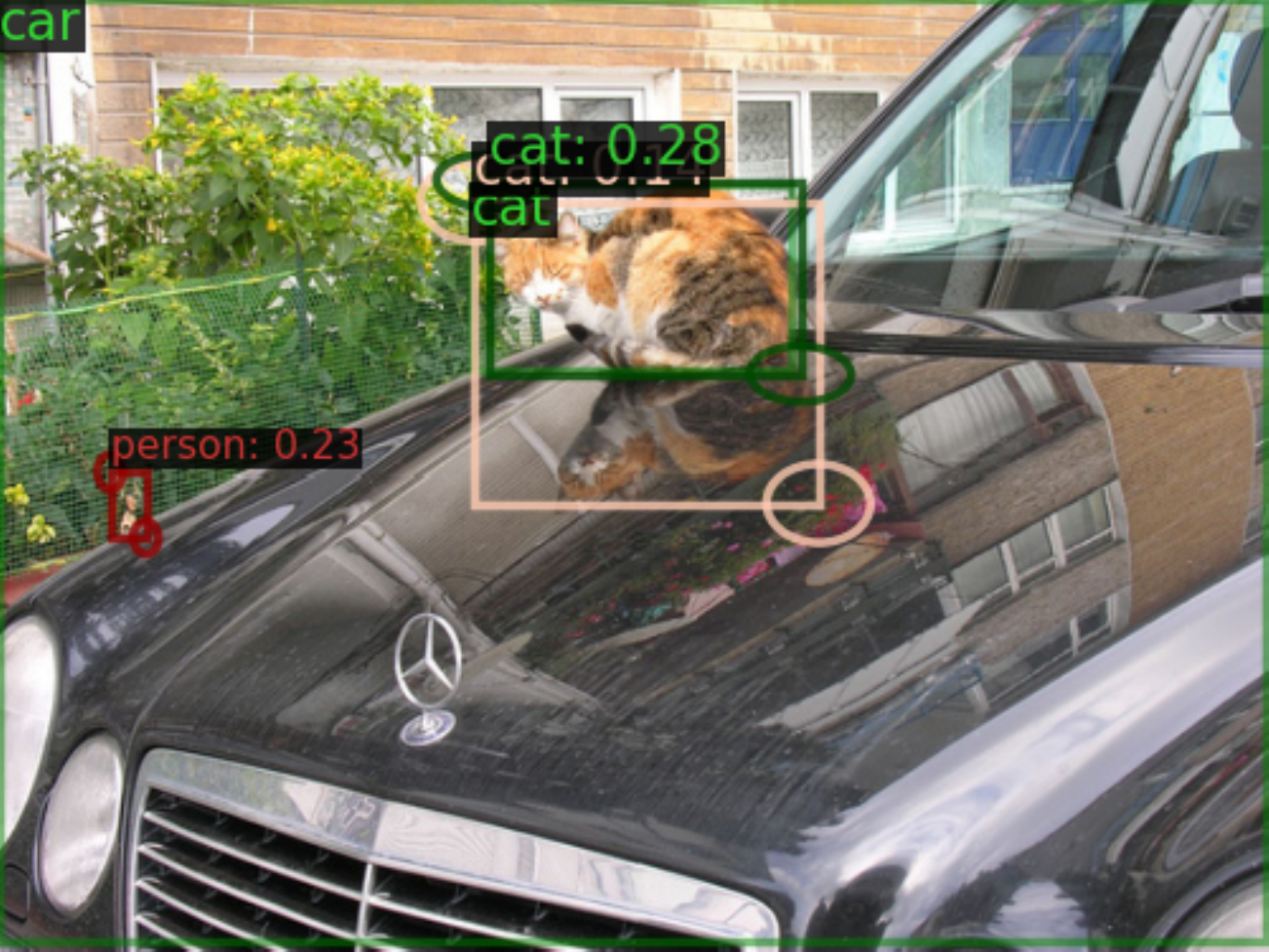}
  \caption{Example 2.}
  \label{fig:faster-rcnn-example-2}
\end{subfigure}
\newline
\begin{subfigure}{\textwidth}
  \centering
  \includegraphics[width=4cm]{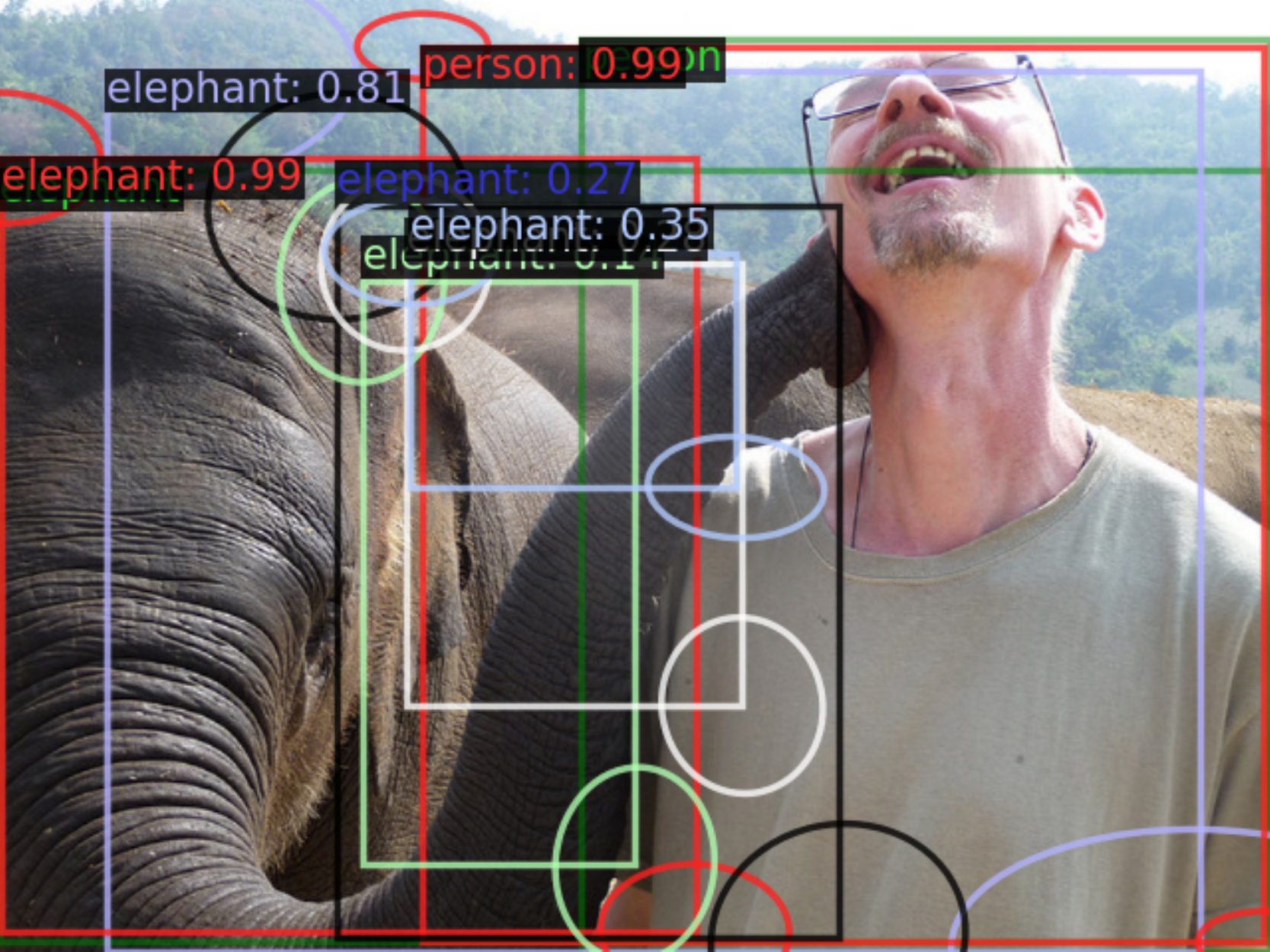}
  \includegraphics[width=4cm]{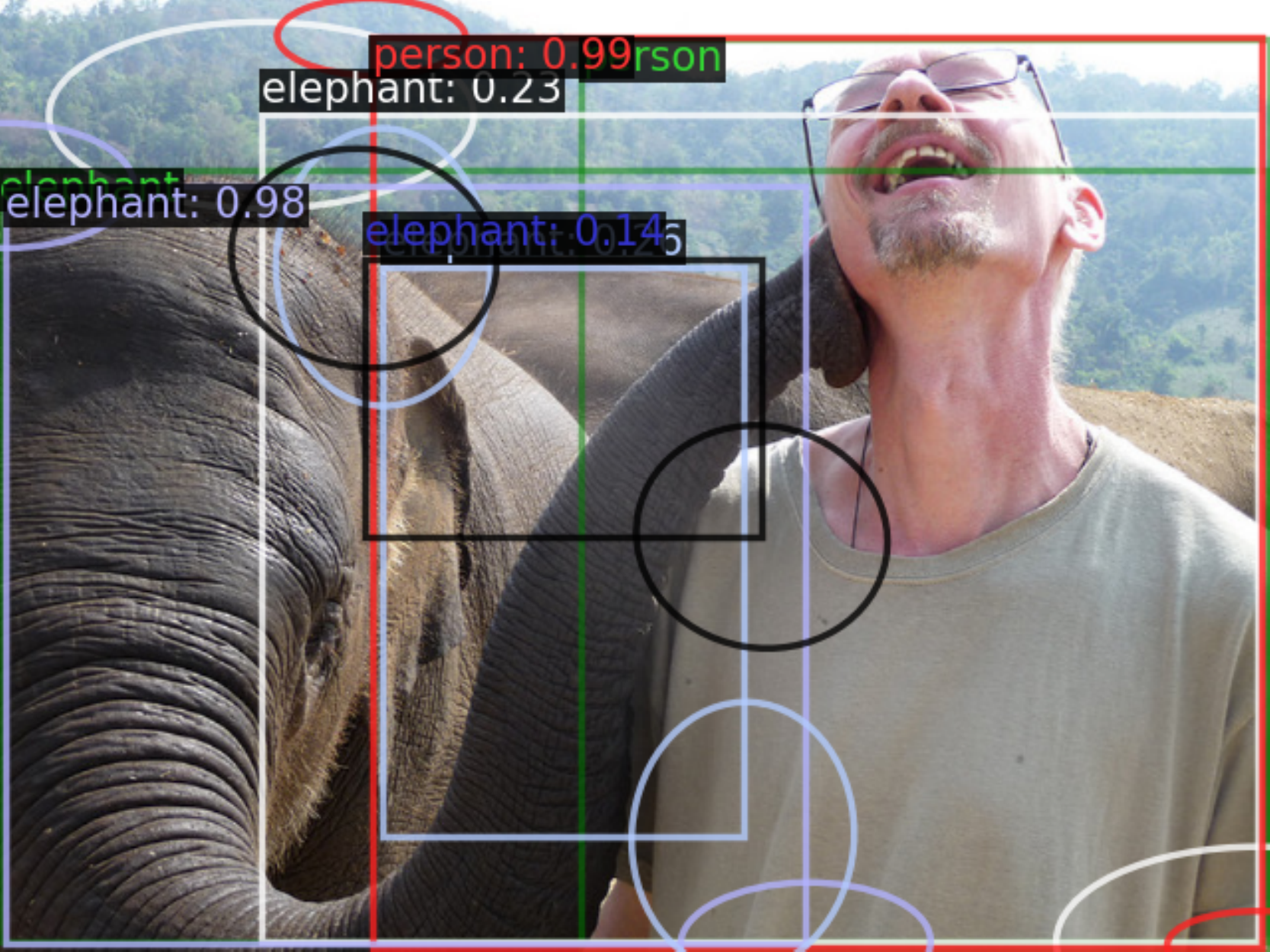}
  \includegraphics[width=4cm]{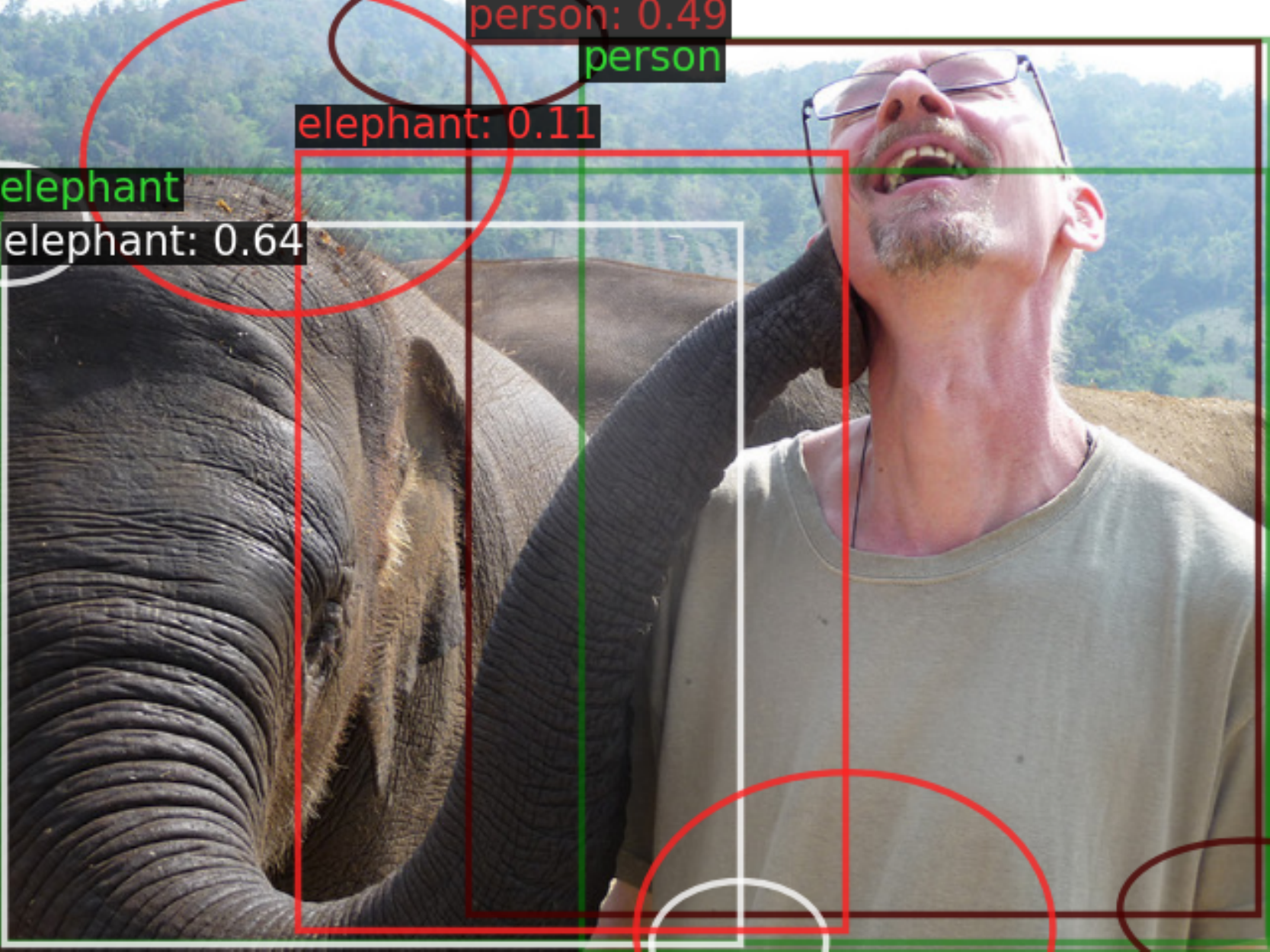}
  \caption{Example 3.}
  \label{fig:faster-rcnn-example-3}
\end{subfigure}
\caption{Examples from COCO validation data with predictions made by Faster-RCNN detectors trained with ES (left), NLL (middle), and MB-NLL (right). True objects are shown in green and without confidence values. Predictions with $r<0.1$ are not shown.}
\label{fig:faster-rcnn-examples}
\end{figure}

In addition to the example detections shown in Section 4, we provide further examples for DETR, RetinaNet and Faster-RCNN in figures \ref{fig:detr-examples}, \ref{fig:retinanet-examples} and \ref{fig:faster-rcnn-examples}. We can identify similar trends in these examples as in the ones described in our results. First, Fig. \ref{fig:faster-rcnn-example-2} shows additional examples where the assignment is ambiguous. There, the cat predictions for Faster-RCNN trained with MB-NLL either have large regression or classification errors, depending on which one of them is assigned to the true object. Second, we see that MB-NLL reduces the number of confident false detections across all models. When comparing the models trained with ES and MB-NLL, the reduction of false detections can also be interpreted as a different representation of spatial uncertainty. In both Fig. \ref{fig:detr-example-1} and Fig. \ref{fig:retinanet-example-1}, there is uncertainty in where the surfboard ends on the left side. The MB-NLL models have a single prediction with larger regression uncertainty, while the ES models have many detections, each with relatively small spatial uncertainty. Further, the MB-NLL loss is normalized with the number of predictions during training.

\subsection{PMB-NLL Decomposition}
\begin{figure}[thp]
\centering
\begin{subfigure}{\textwidth}
  \centering
  \adjustbox{trim=0.2cm 0.2cm 0.2cm 0.2cm}{\includegraphics[width=4cm]{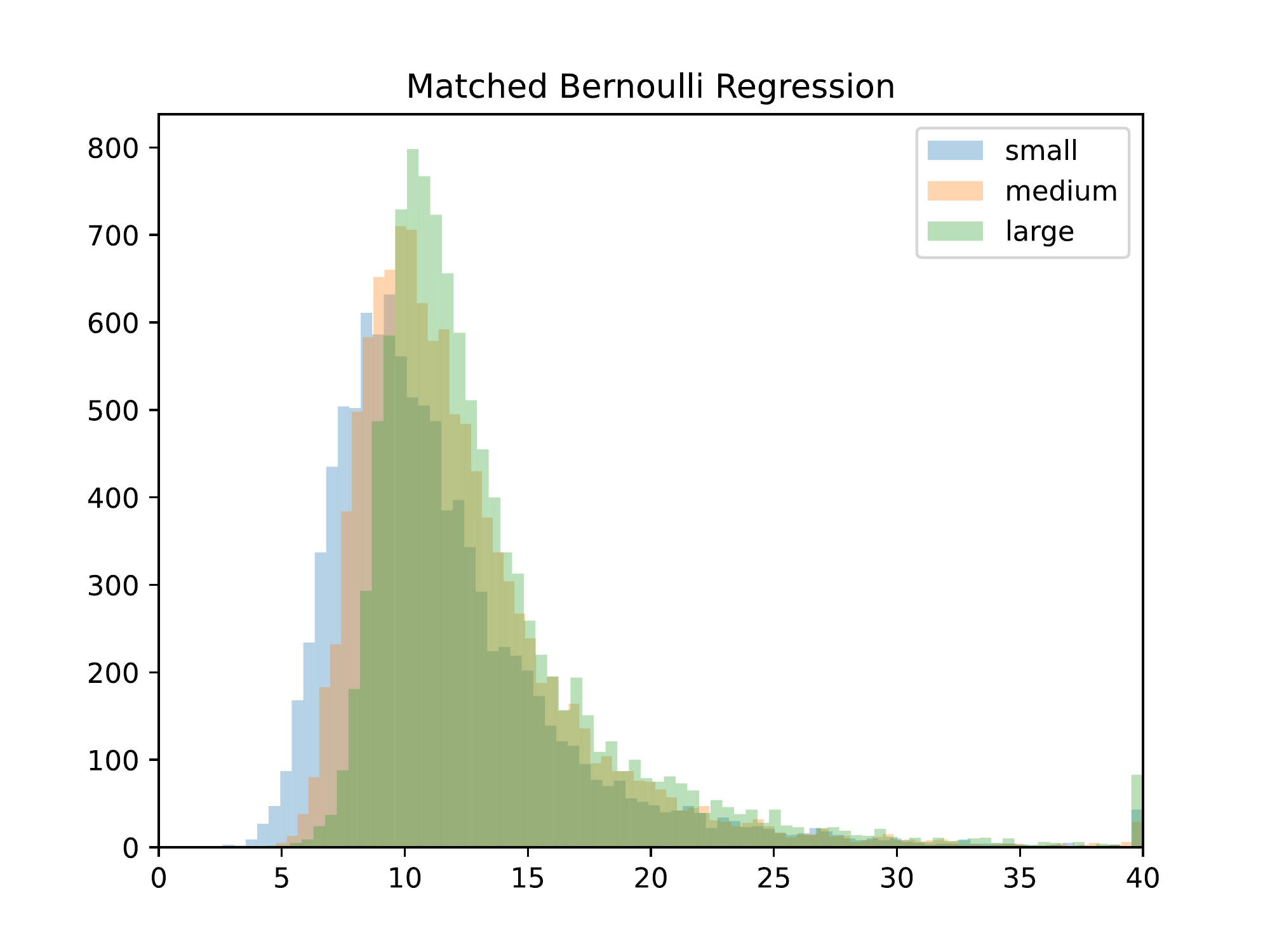}}
  \adjustbox{trim=0.2cm 0.2cm 0.2cm 0.2cm}{\includegraphics[width=4cm]{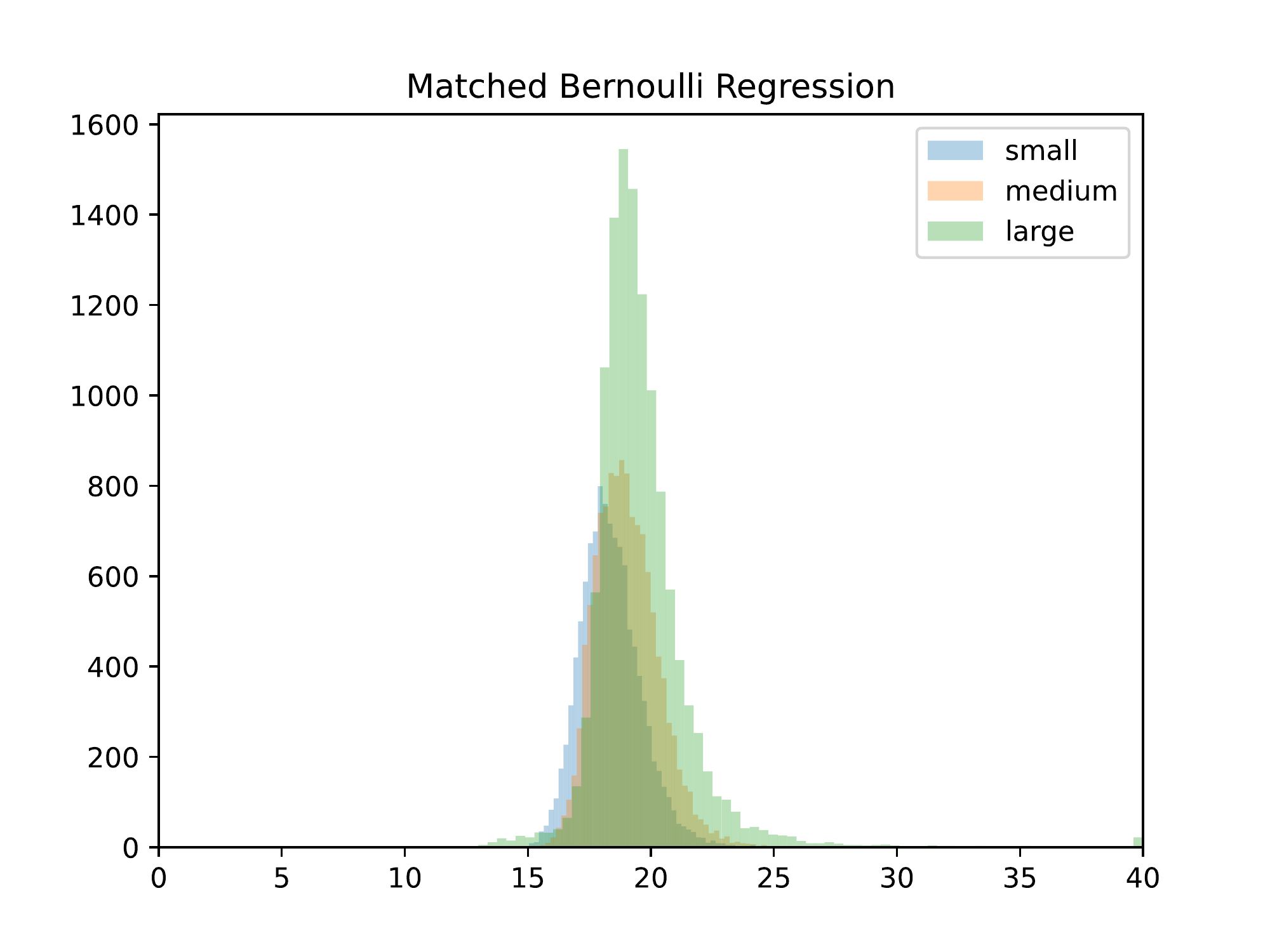}}
  \adjustbox{trim=0.2cm 0.2cm 0.2cm 0.2cm}{\includegraphics[width=4cm]{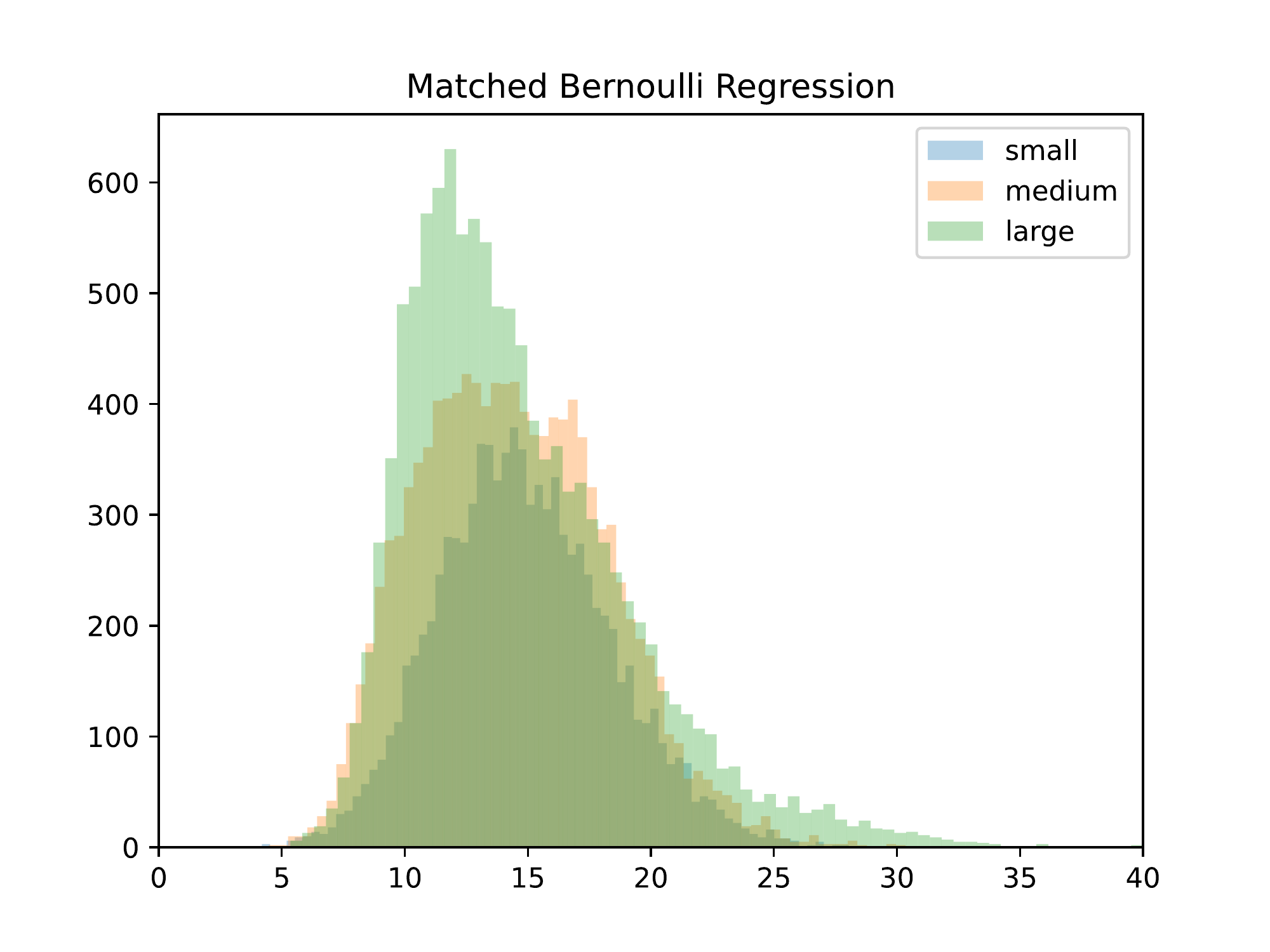}}
  \caption{Matched Bernoullis regression.}
  \label{fig:detr-reg}
\end{subfigure}
\newline
\begin{subfigure}{\textwidth}
  \centering
  \adjustbox{trim=0.2cm 0.2cm 0.2cm 0.2cm}{\includegraphics[width=4cm]{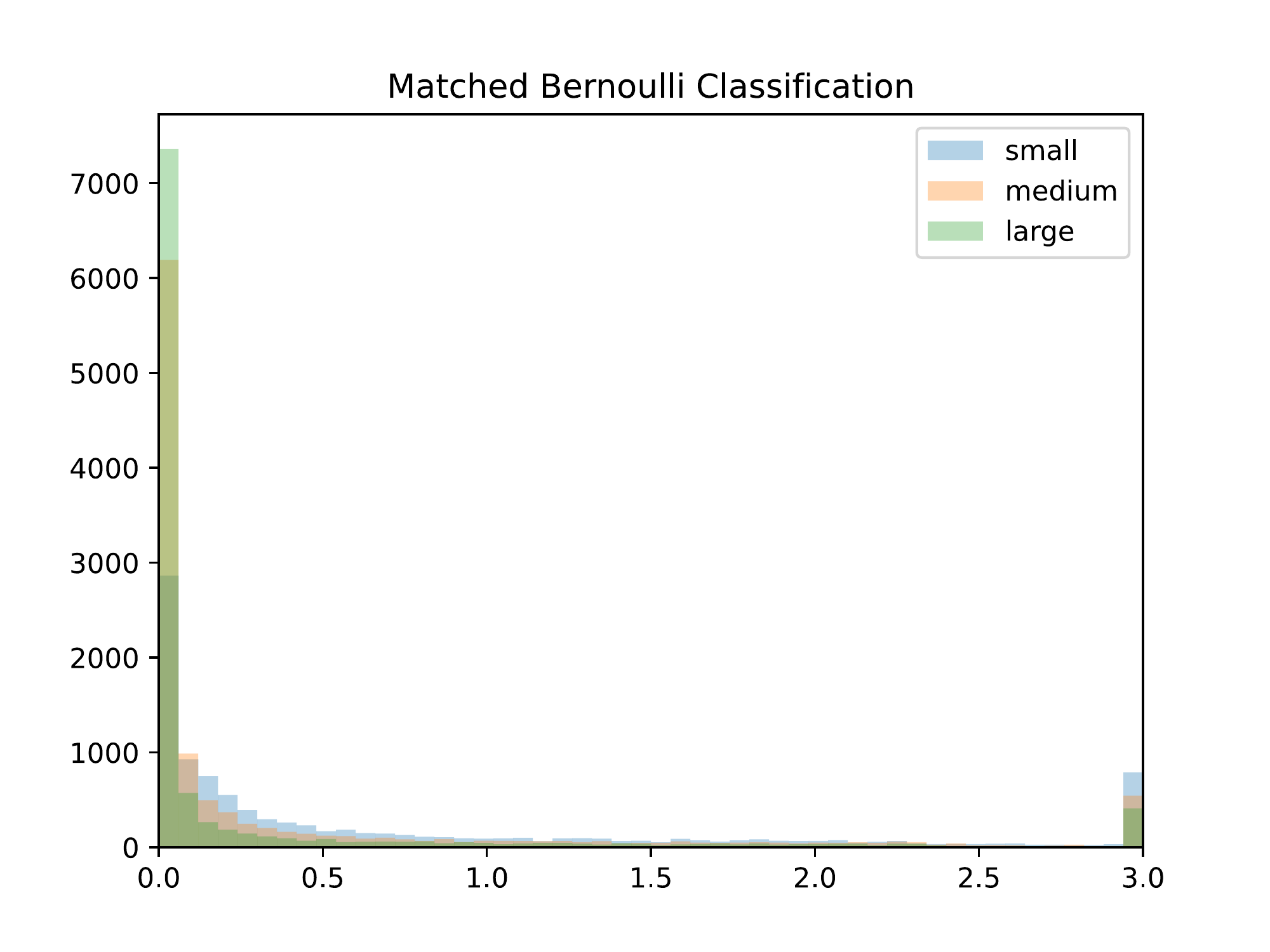}}
  \adjustbox{trim=0.2cm 0.2cm 0.2cm 0.2cm}{\includegraphics[width=4cm]{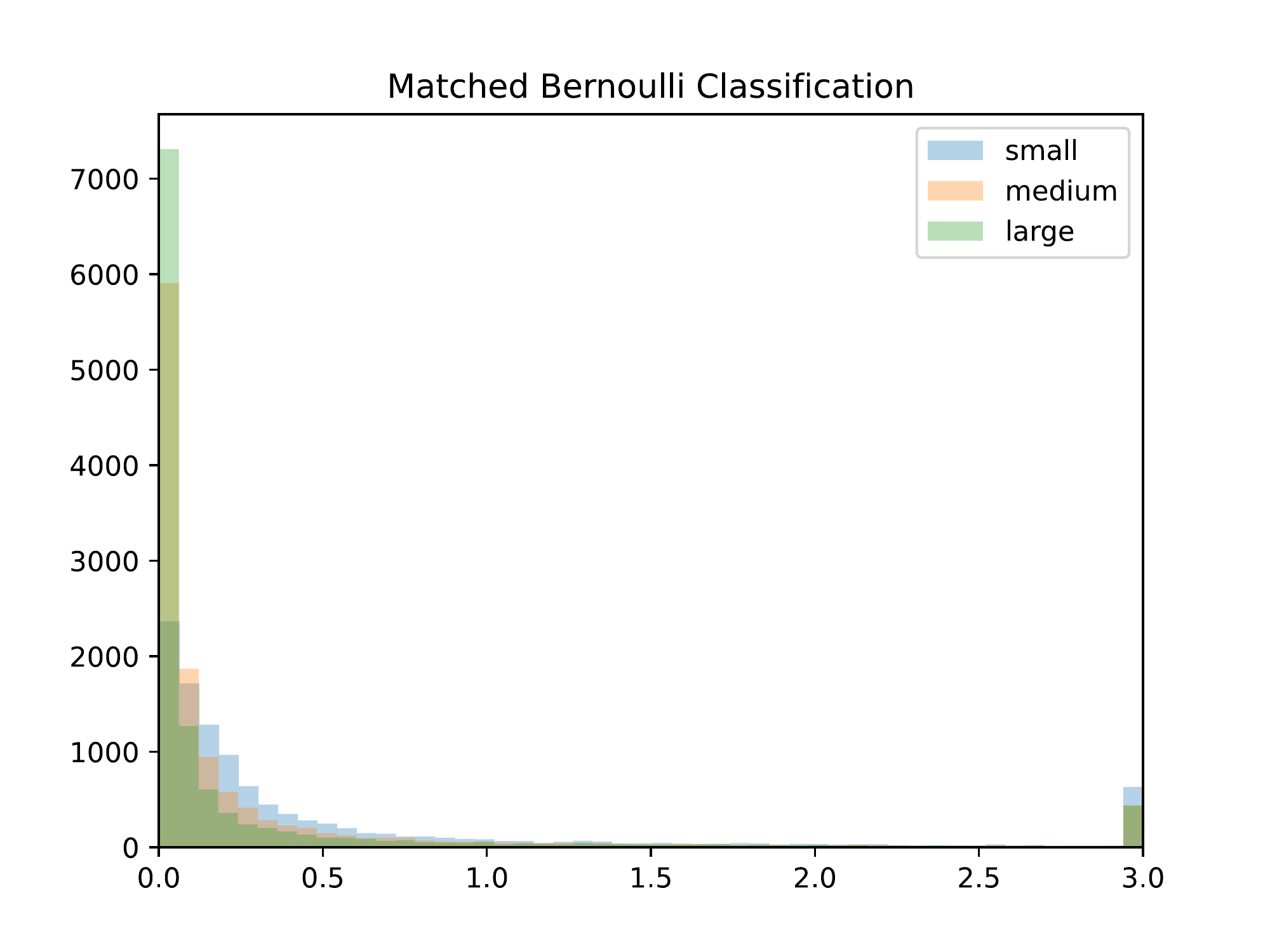}}
  \adjustbox{trim=0.2cm 0.2cm 0.2cm 0.2cm}{\includegraphics[width=4cm]{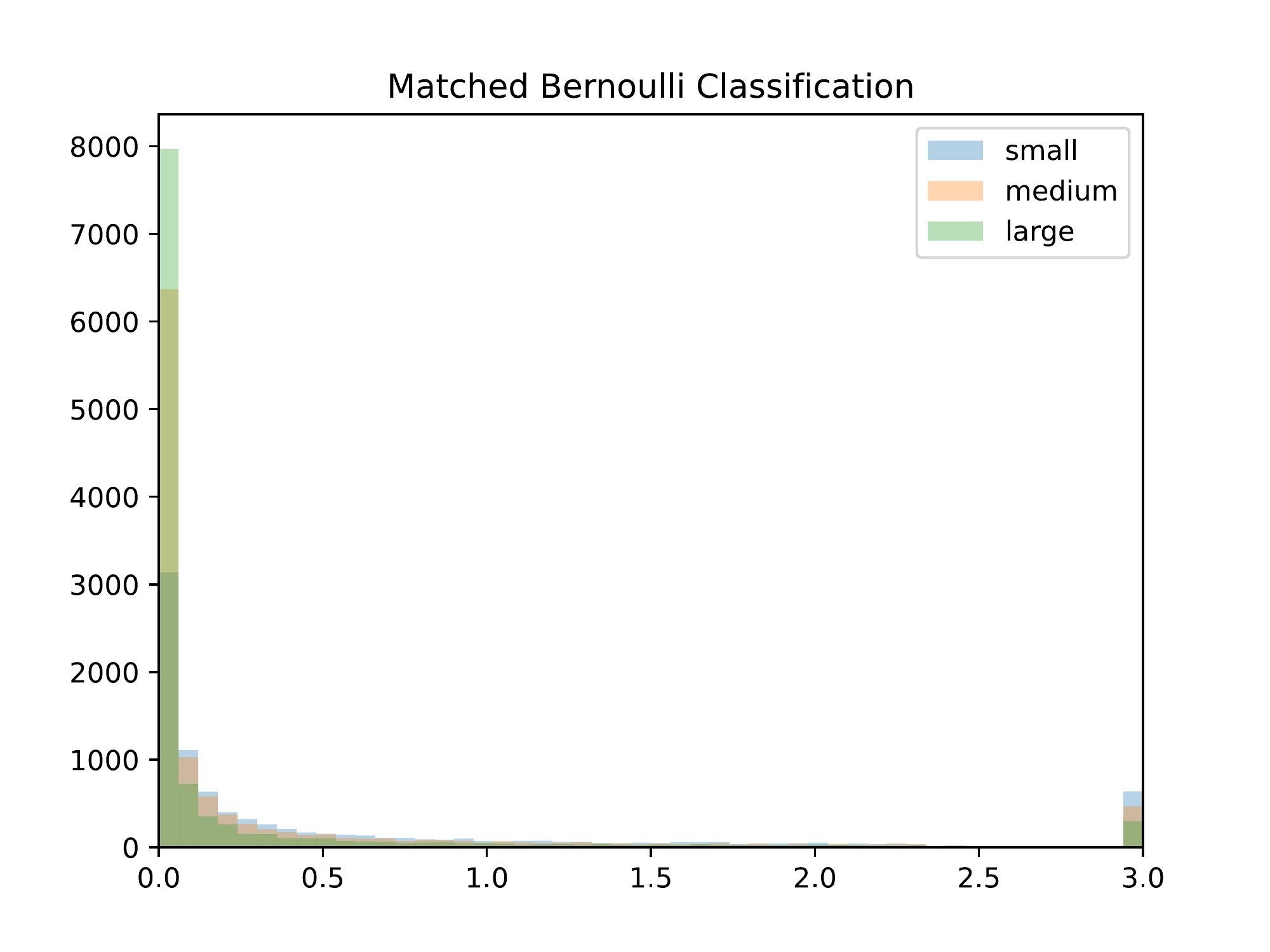}}
  \caption{Matched Bernoullis classification.}
  \label{fig:detr-cls}
\end{subfigure}
\newline
\begin{subfigure}{\textwidth}
  \centering
  \adjustbox{trim=0.2cm 0.2cm 0.2cm 0.2cm}{\includegraphics[width=4cm]{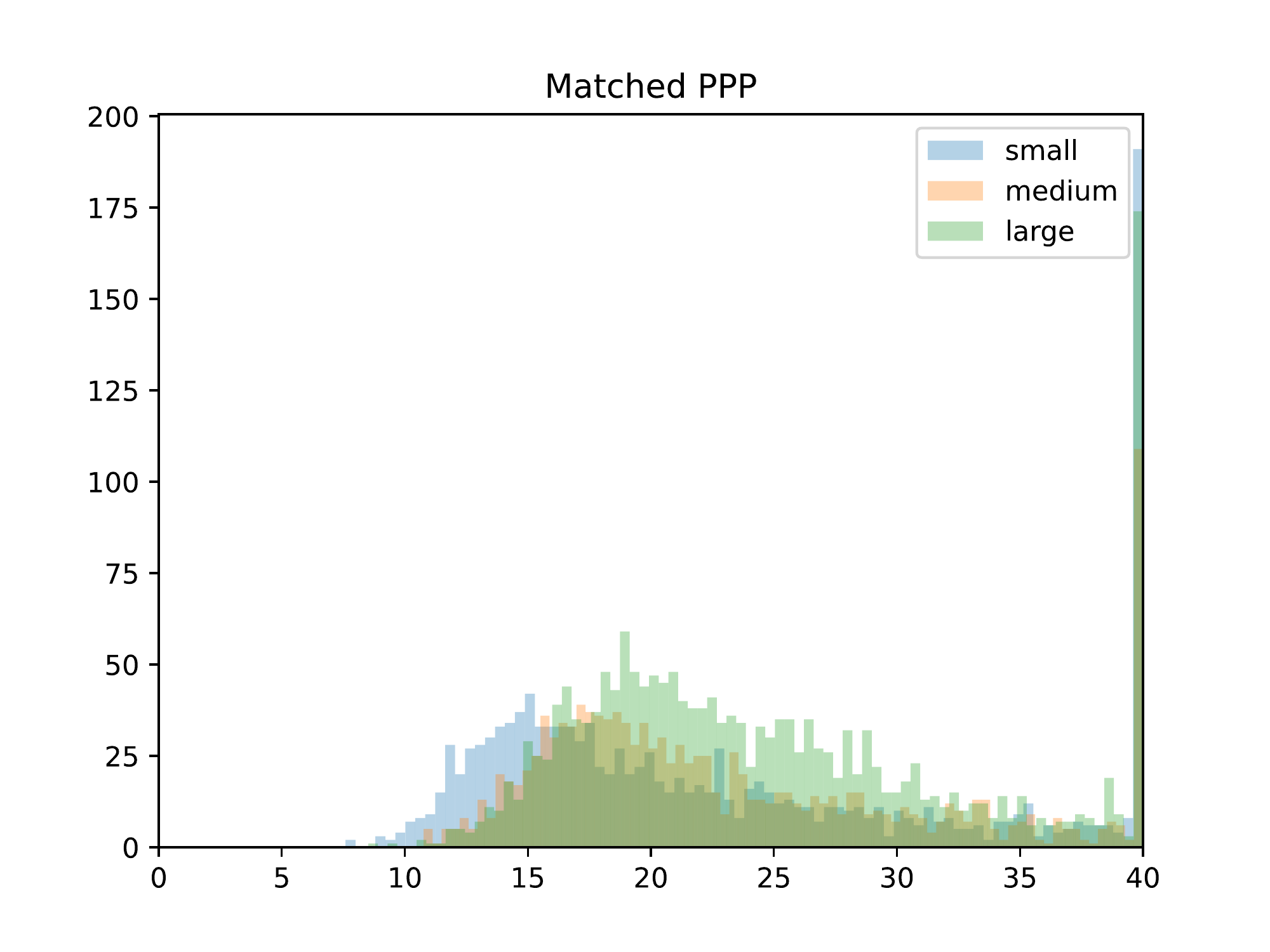}}
  \adjustbox{trim=0.2cm 0.2cm 0.2cm 0.2cm}{\includegraphics[width=4cm]{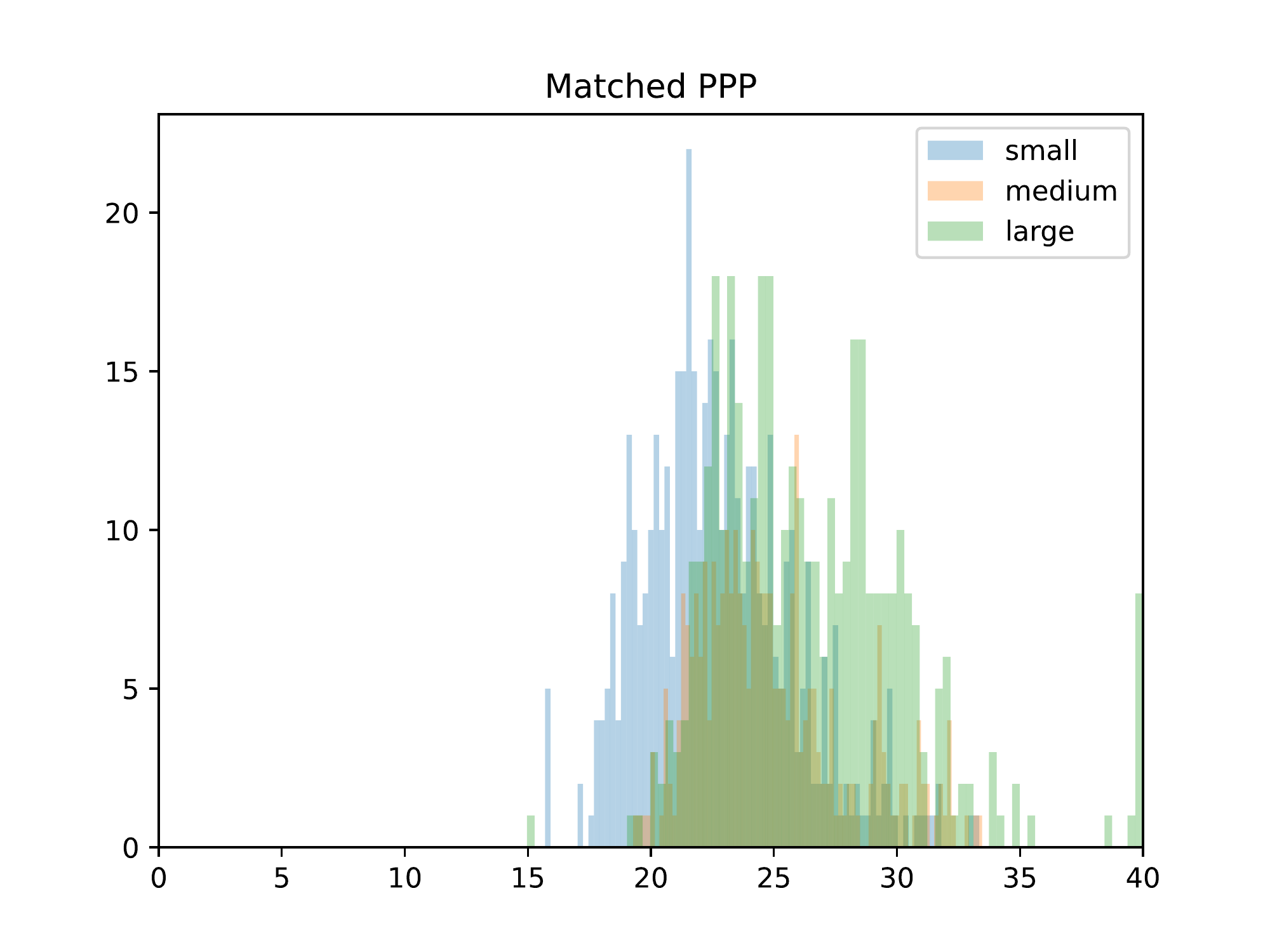}}
  \adjustbox{trim=0.2cm 0.2cm 0.2cm 0.2cm}{\includegraphics[width=4cm]{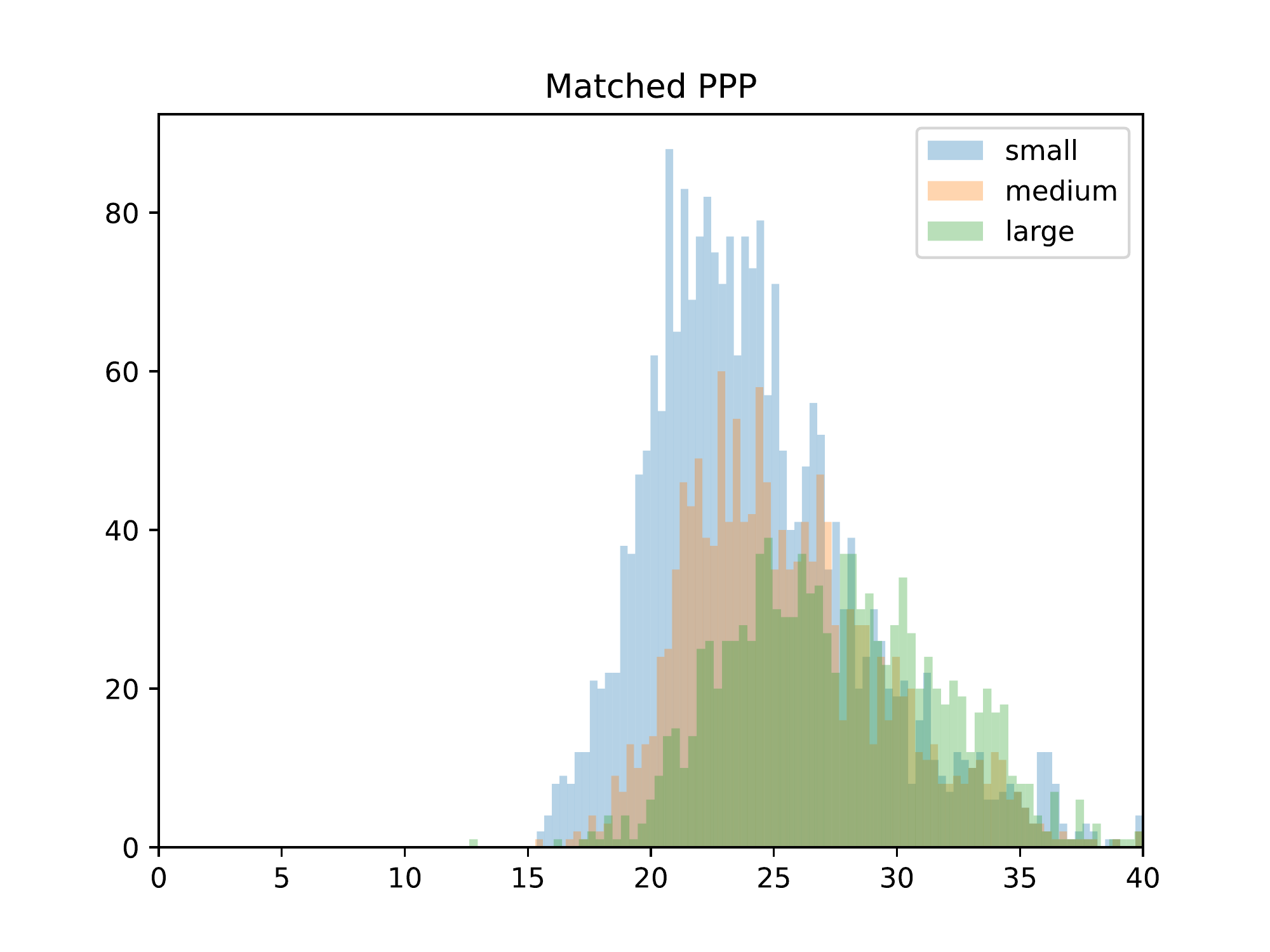}}
  \caption{Matched PPP.}
  \label{fig:detr-ppp}
\end{subfigure}
\newline
\begin{subfigure}{\textwidth}
  \centering
  \adjustbox{trim=0.2cm 0.2cm 0.2cm 0.2cm}{\includegraphics[width=4cm]{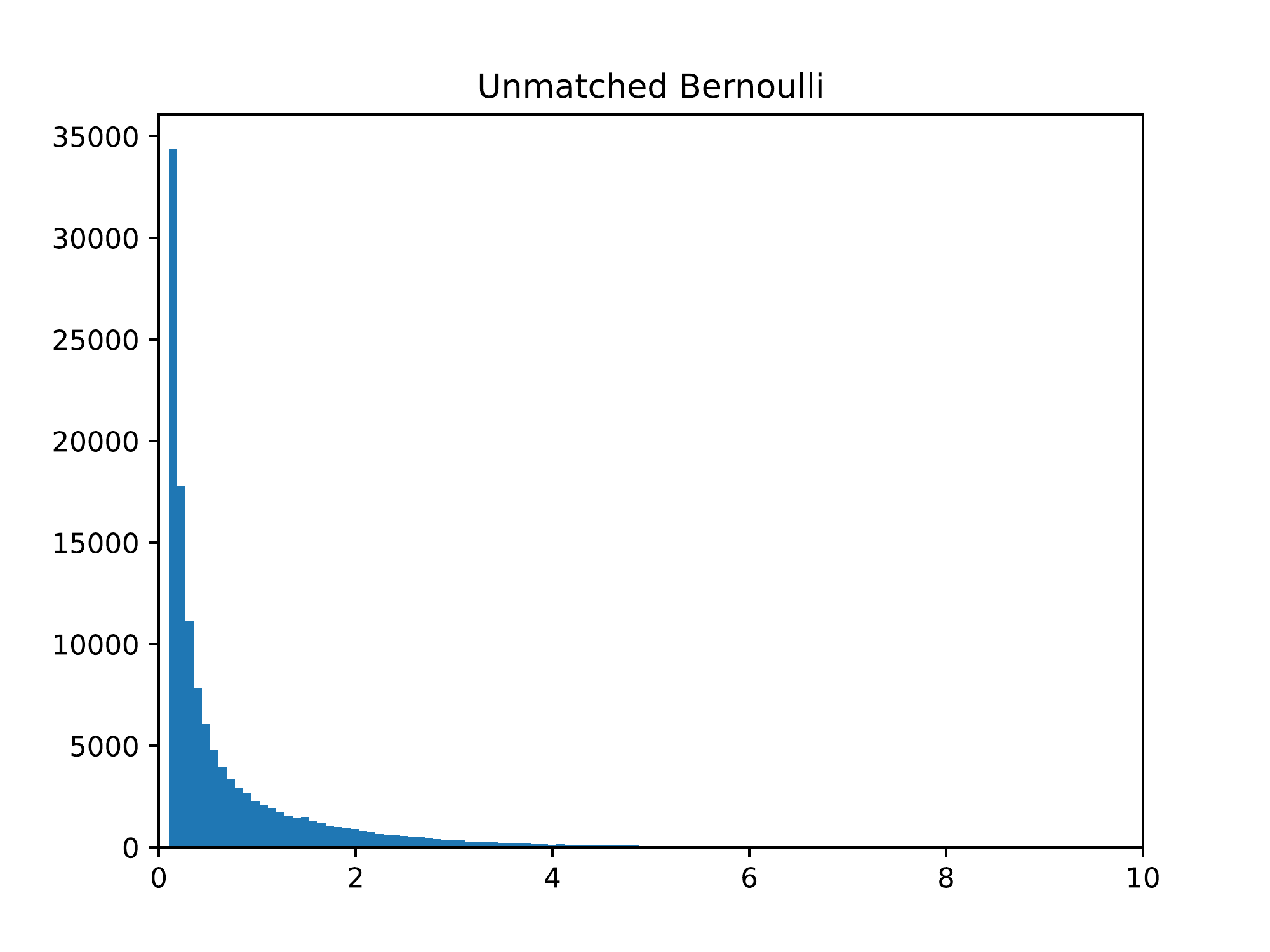}}
  \adjustbox{trim=0.2cm 0.2cm 0.2cm 0.2cm}{\includegraphics[width=4cm]{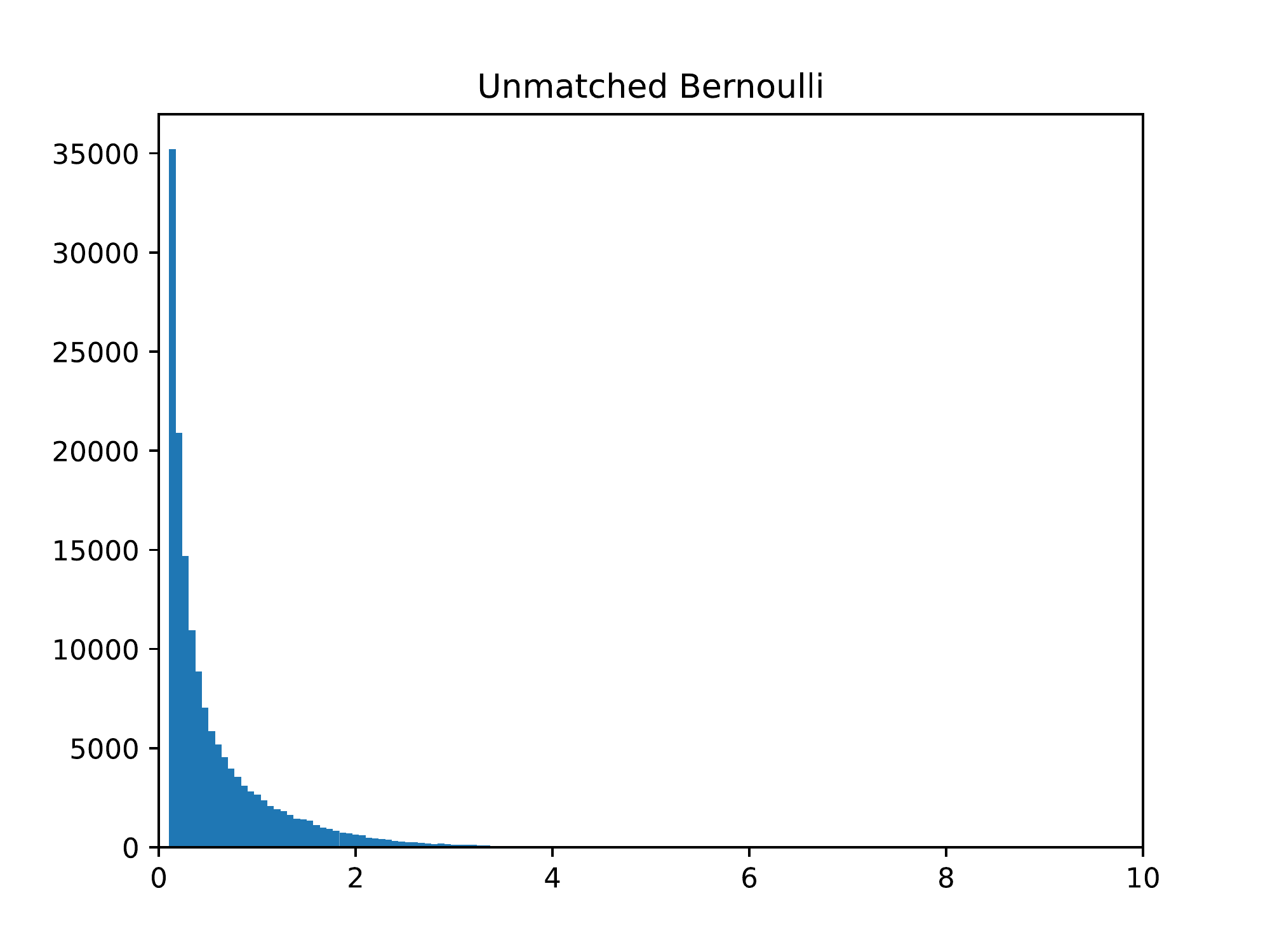}}
  \adjustbox{trim=0.2cm 0.2cm 0.2cm 0.2cm}{\includegraphics[width=4cm]{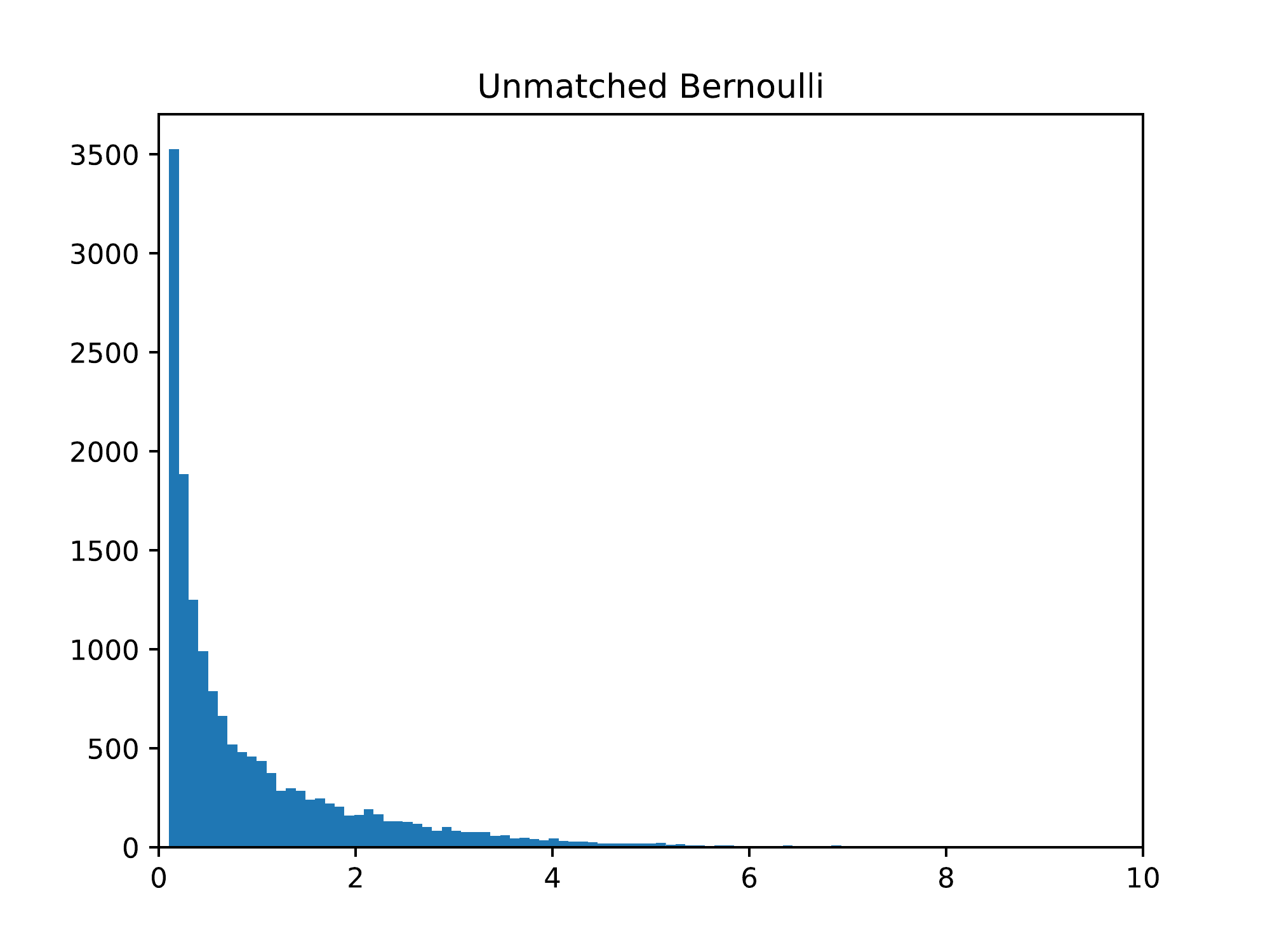}}
  \caption{Unmatched Bernoullis.}
  \label{fig:detr-unmatched}
\end{subfigure}
    \caption{Histograms over PMB-NLL decomposition for DETR trained with different loss functions: ES (left), NLL (middle), and MB-NLL (right). Note varying y-axes across models. }
    \label{fig:detr-hist}
\end{figure}

\begin{figure}[thp]
\centering
\begin{subfigure}{\textwidth}
  \centering
  \adjustbox{trim=0.2cm 0.2cm 0.2cm 0.2cm}{\includegraphics[width=4cm]{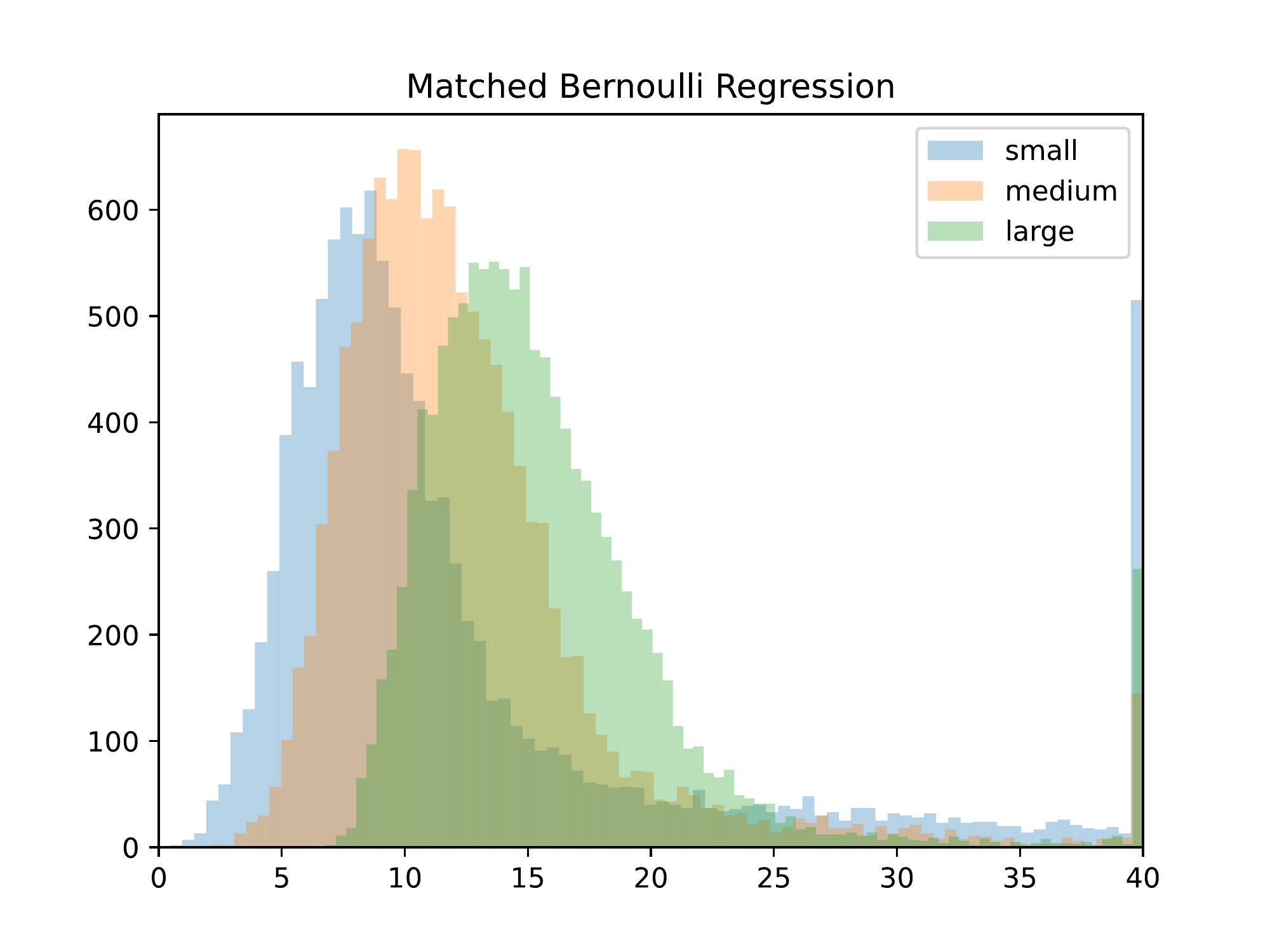}}
  \adjustbox{trim=0.2cm 0.2cm 0.2cm 0.2cm}{\includegraphics[width=4cm]{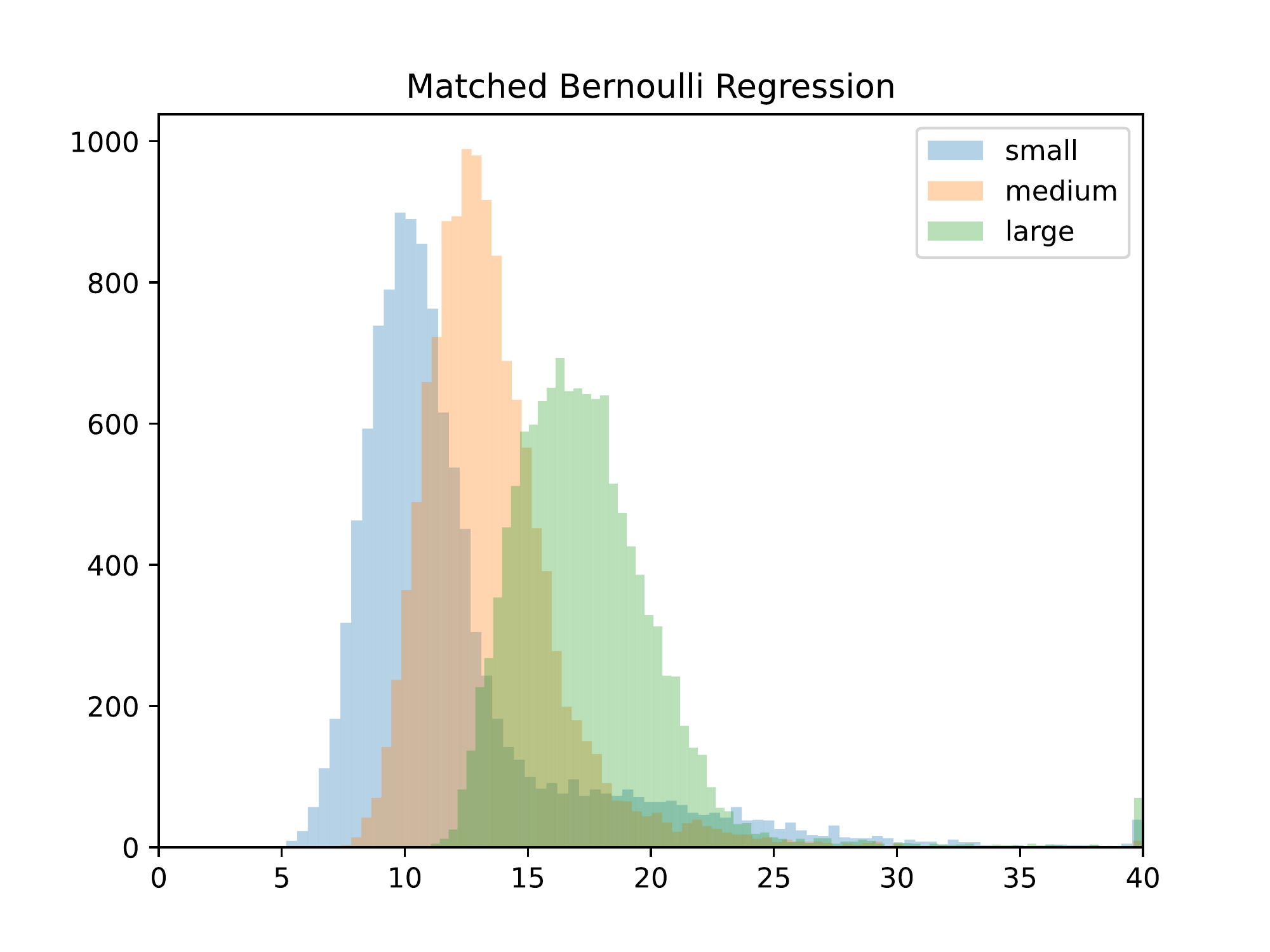}}
  \adjustbox{trim=0.2cm 0.2cm 0.2cm 0.2cm}{\includegraphics[width=4cm]{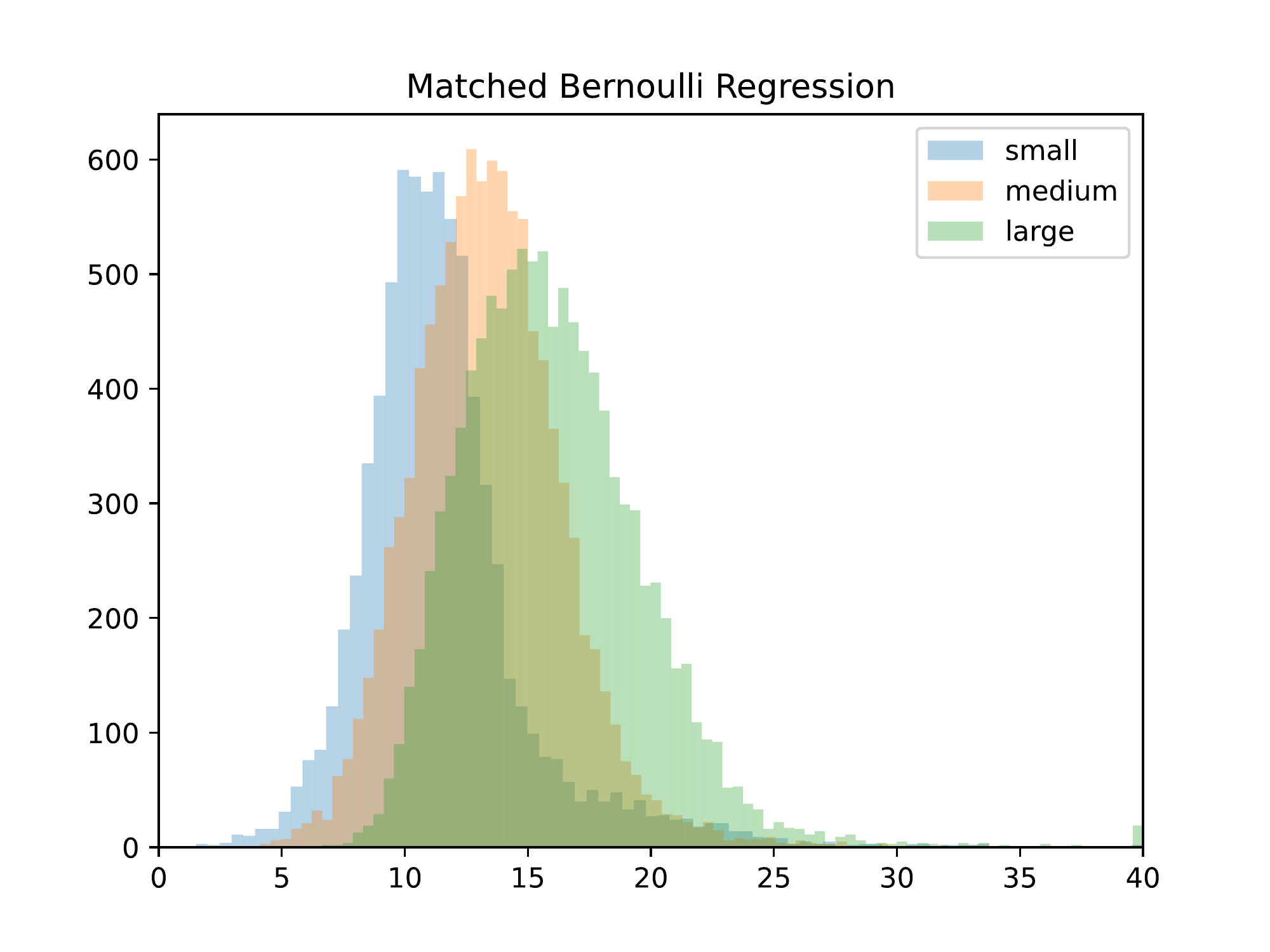}}
  \caption{Matched Bernoullis regression.}
  \label{fig:retinanet-reg}
\end{subfigure}
\newline
\begin{subfigure}{\textwidth}
  \centering
  \adjustbox{trim=0.2cm 0.2cm 0.2cm 0.2cm}{\includegraphics[width=4cm]{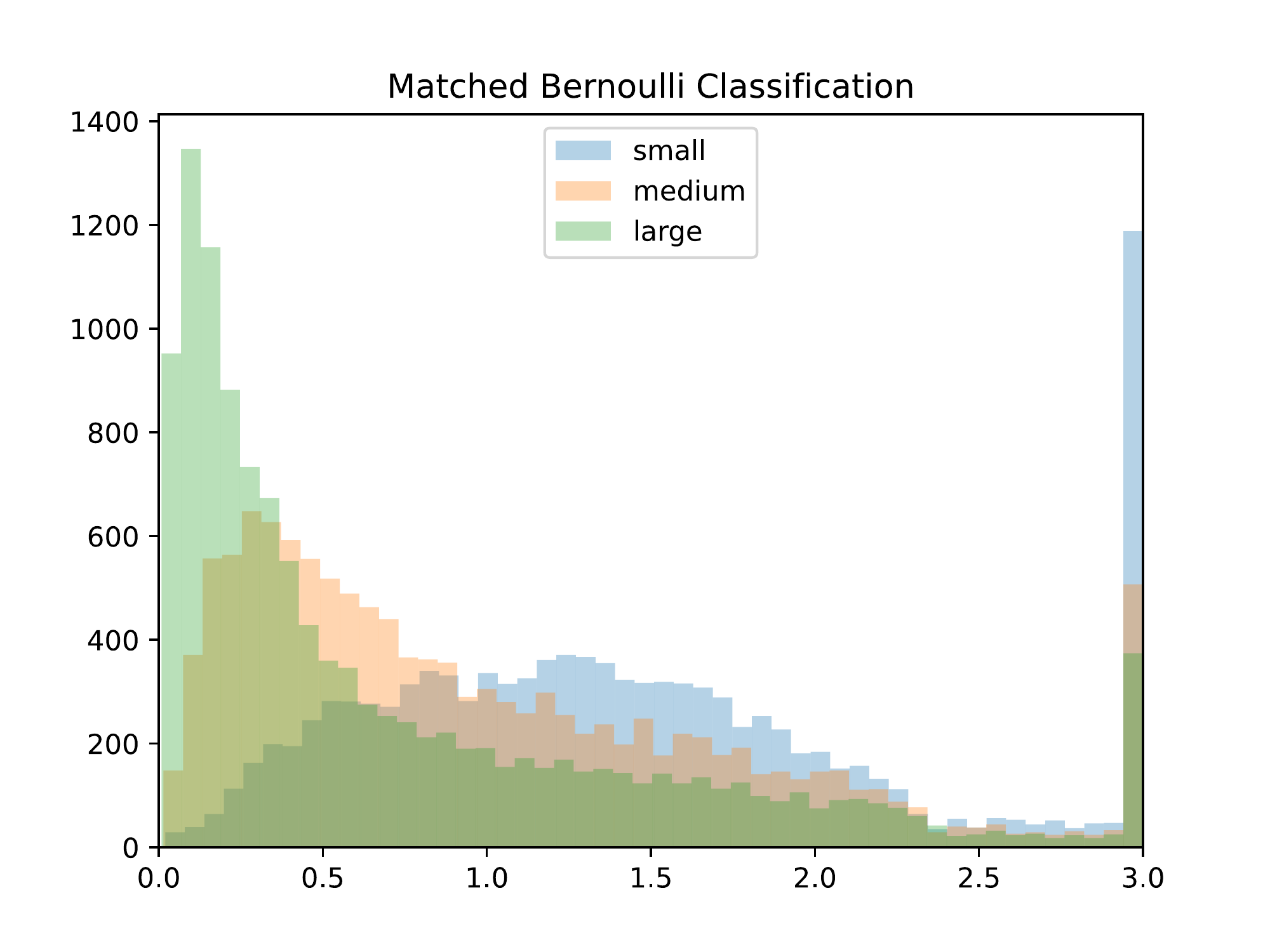}}
  \adjustbox{trim=0.2cm 0.2cm 0.2cm 0.2cm}{\includegraphics[width=4cm]{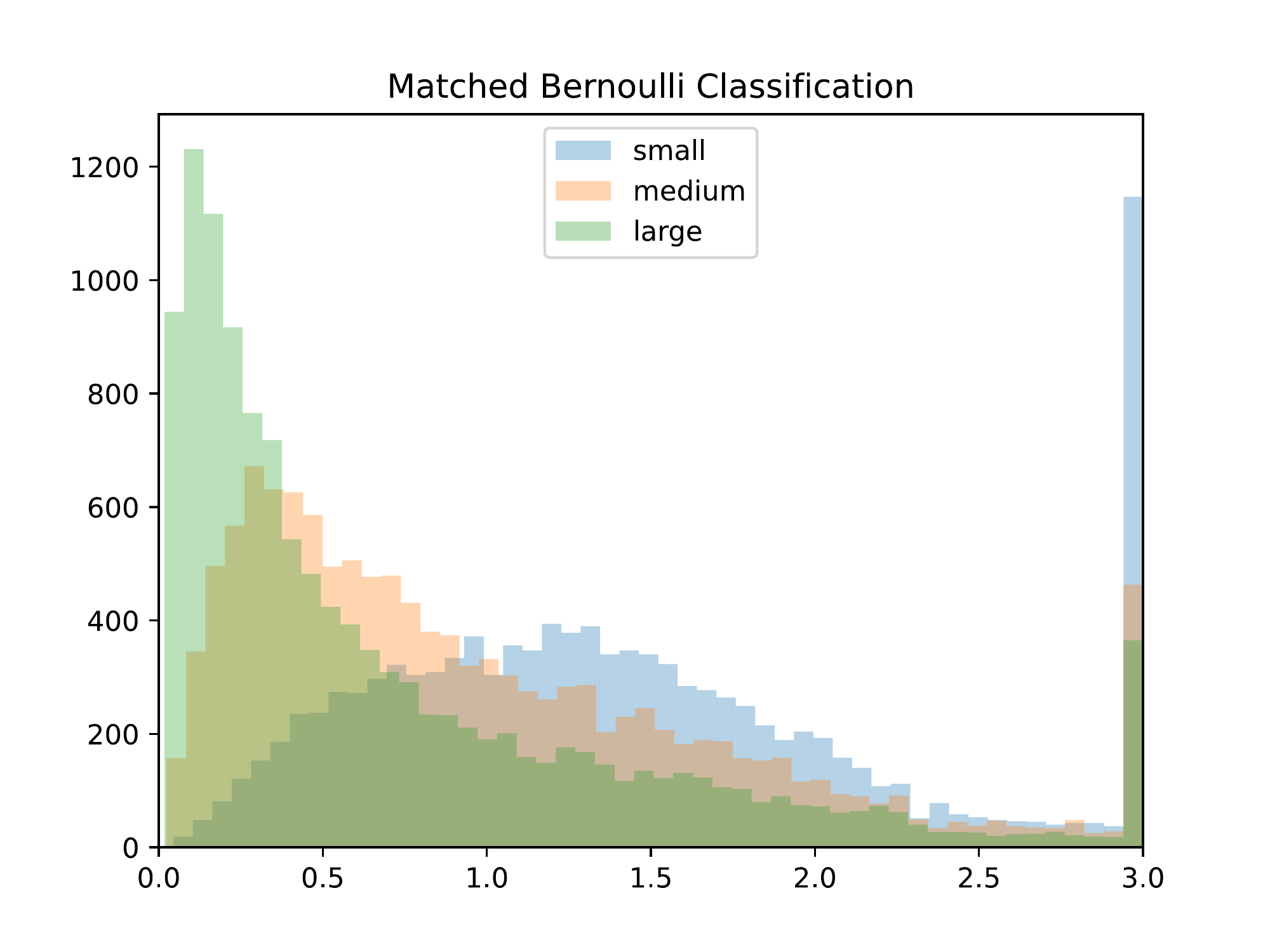}}
  \adjustbox{trim=0.2cm 0.2cm 0.2cm 0.2cm}{\includegraphics[width=4cm]{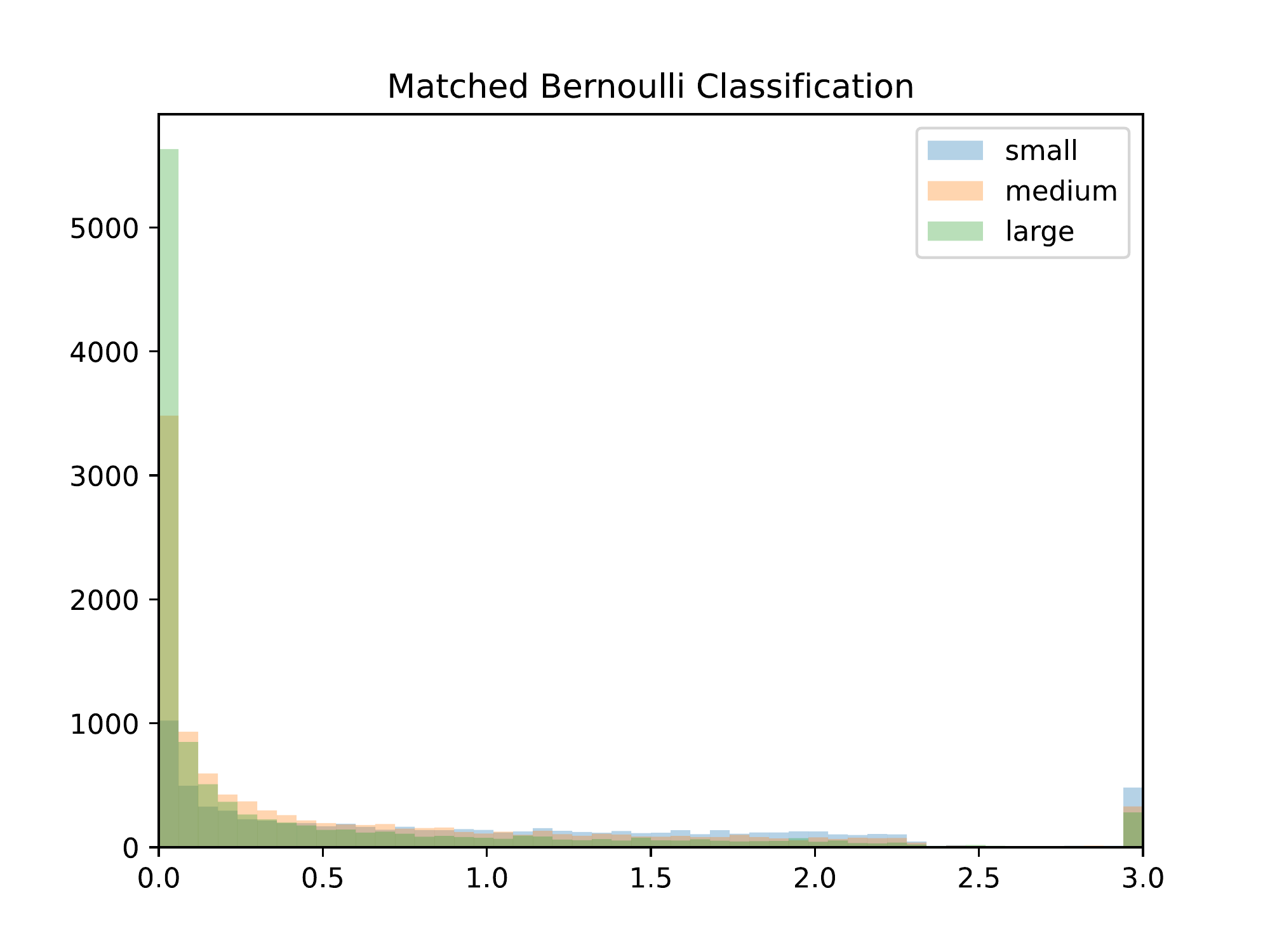}}
  \caption{Matched Bernoullis classification.}
  \label{fig:retinanet-cls}
\end{subfigure}
\newline
\begin{subfigure}{\textwidth}
  \centering
  \adjustbox{trim=0.2cm 0.2cm 0.2cm 0.2cm}{\includegraphics[width=4cm]{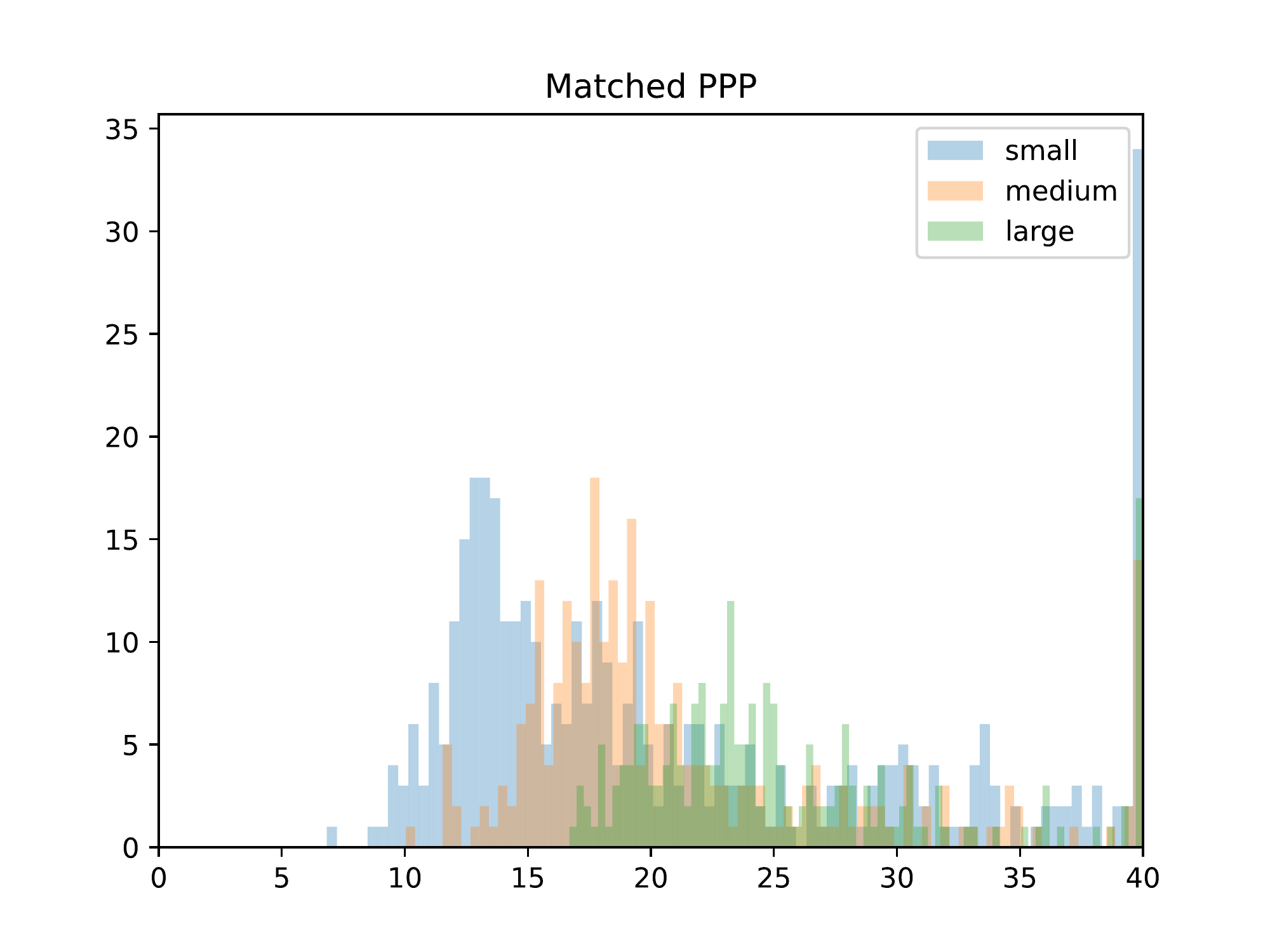}}
  \adjustbox{trim=0.2cm 0.2cm 0.2cm 0.2cm}{\includegraphics[width=4cm]{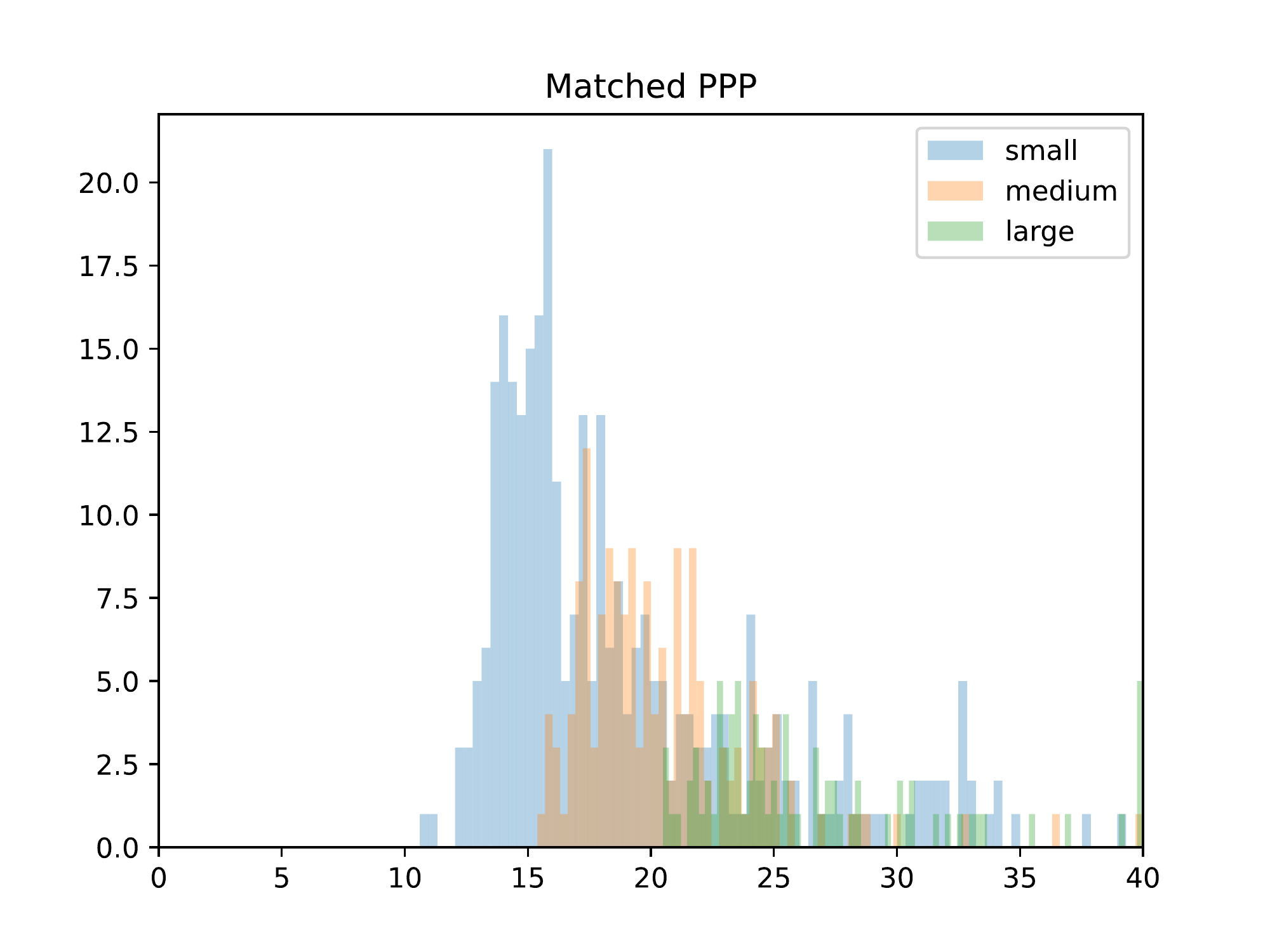}}
  \adjustbox{trim=0.2cm 0.2cm 0.2cm 0.2cm}{\includegraphics[width=4cm]{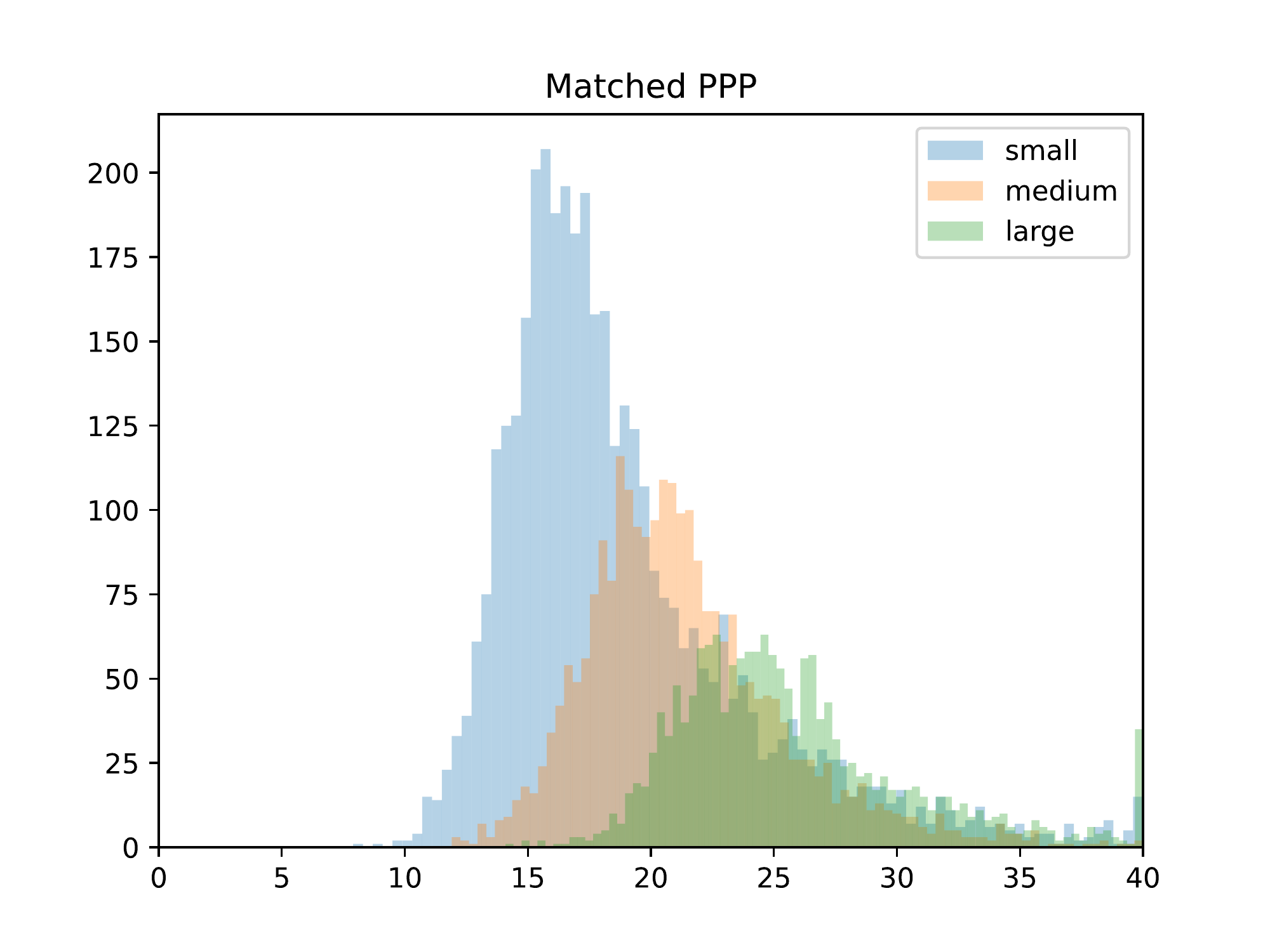}}
  \caption{Matched PPP.}
  \label{fig:retinanet-ppp}
\end{subfigure}
\newline
\begin{subfigure}{\textwidth}
  \centering
  \adjustbox{trim=0.2cm 0.2cm 0.2cm 0.2cm}{\includegraphics[width=4cm]{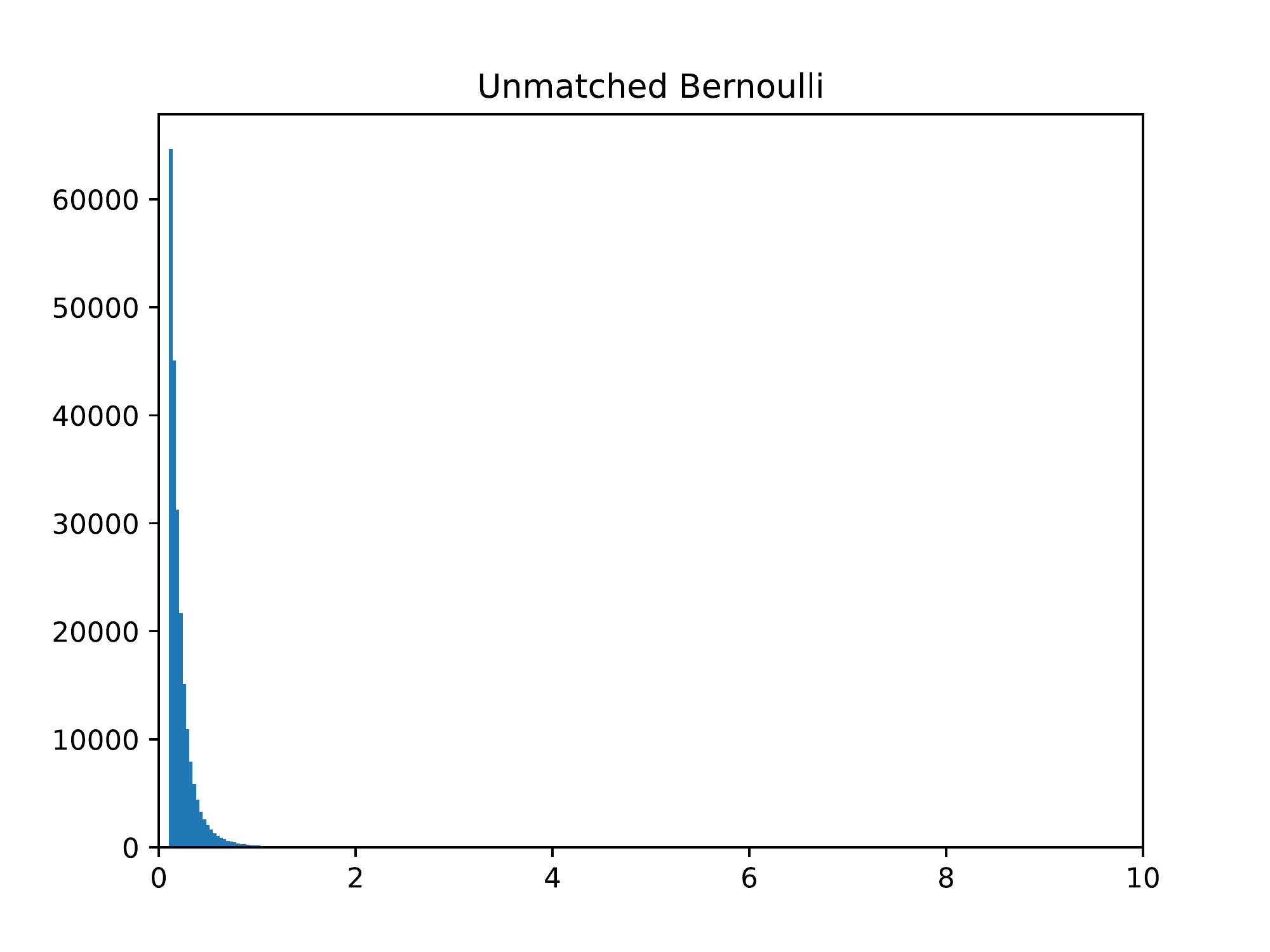}}
  \adjustbox{trim=0.2cm 0.2cm 0.2cm 0.2cm}{\includegraphics[width=4cm]{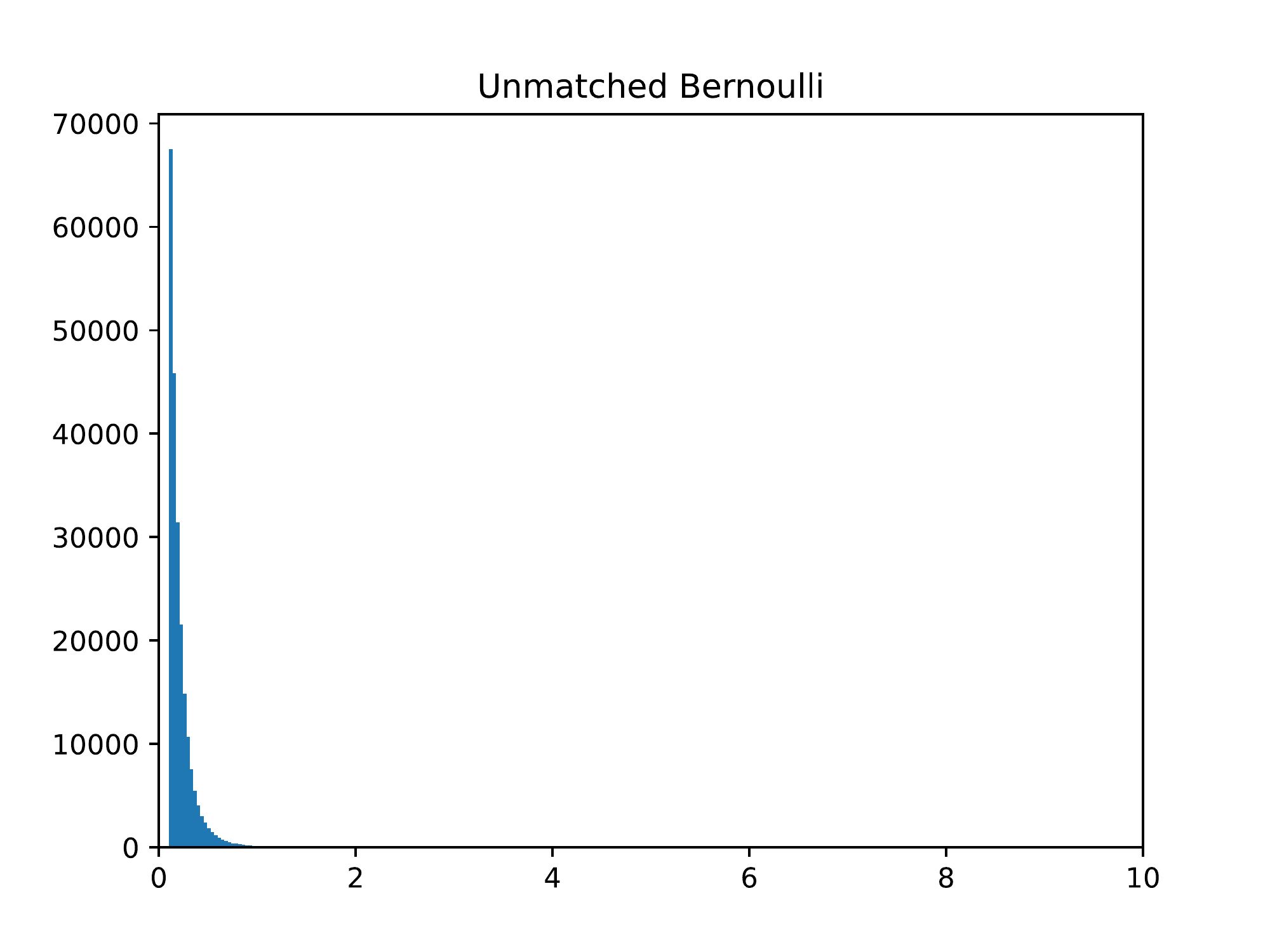}}
  \adjustbox{trim=0.2cm 0.2cm 0.2cm 0.2cm}{\includegraphics[width=4cm]{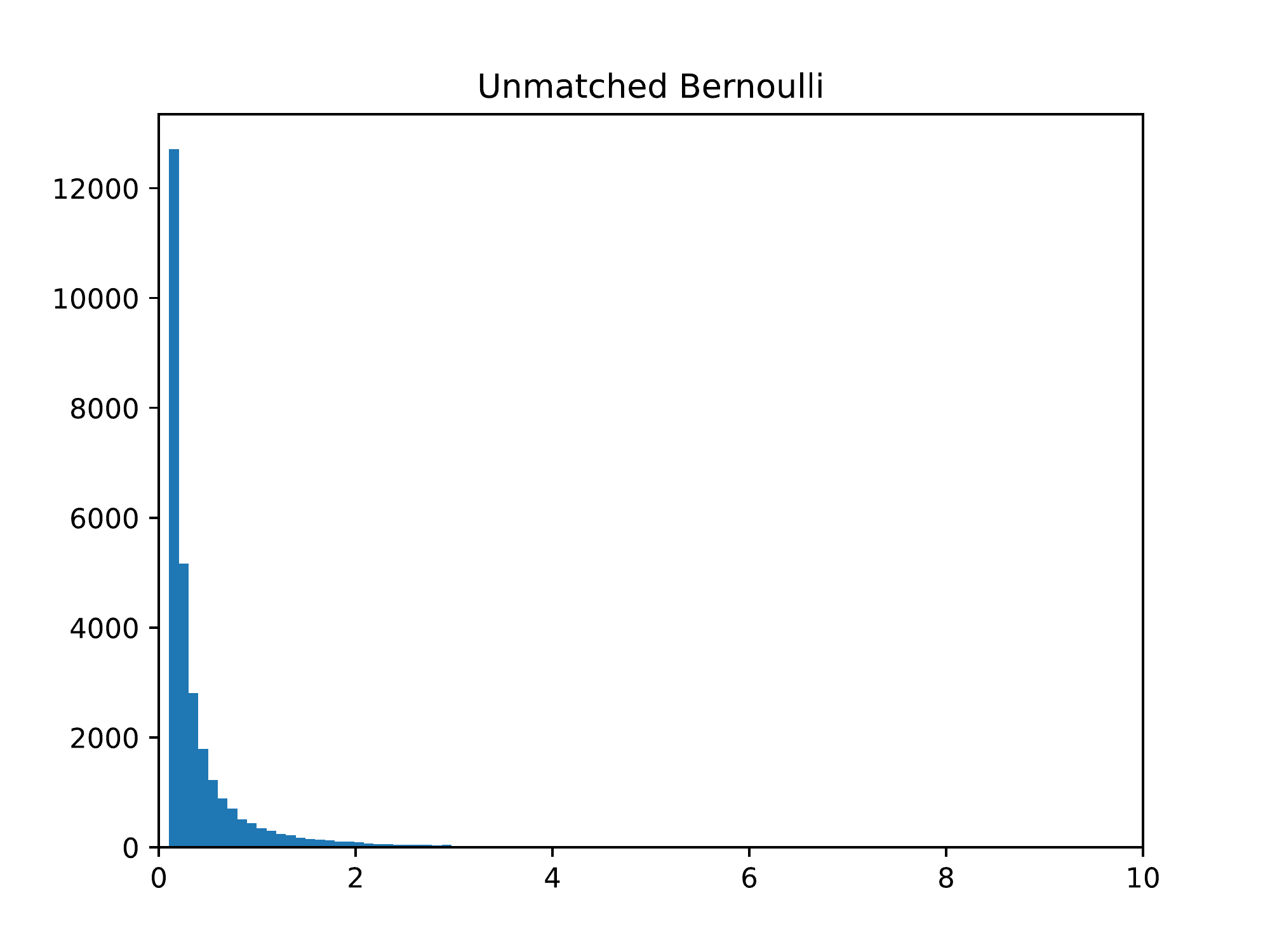}}
  \caption{Unmatched Bernoullis.}
  \label{fig:retinanet-unmatched}
\end{subfigure}
    \caption{Histograms over PMB-NLL decomposition for Retinanet trained with different loss functions: ES (left), NLL (middle), and MB-NLL (right). Note varying y-axes across models. }
    \label{fig:retinanet-hist}
\end{figure}

\begin{figure}[thp]
\centering
\begin{subfigure}{\textwidth}
  \centering
  \adjustbox{trim=0.2cm 0.2cm 0.2cm 0.2cm}{\includegraphics[width=4cm]{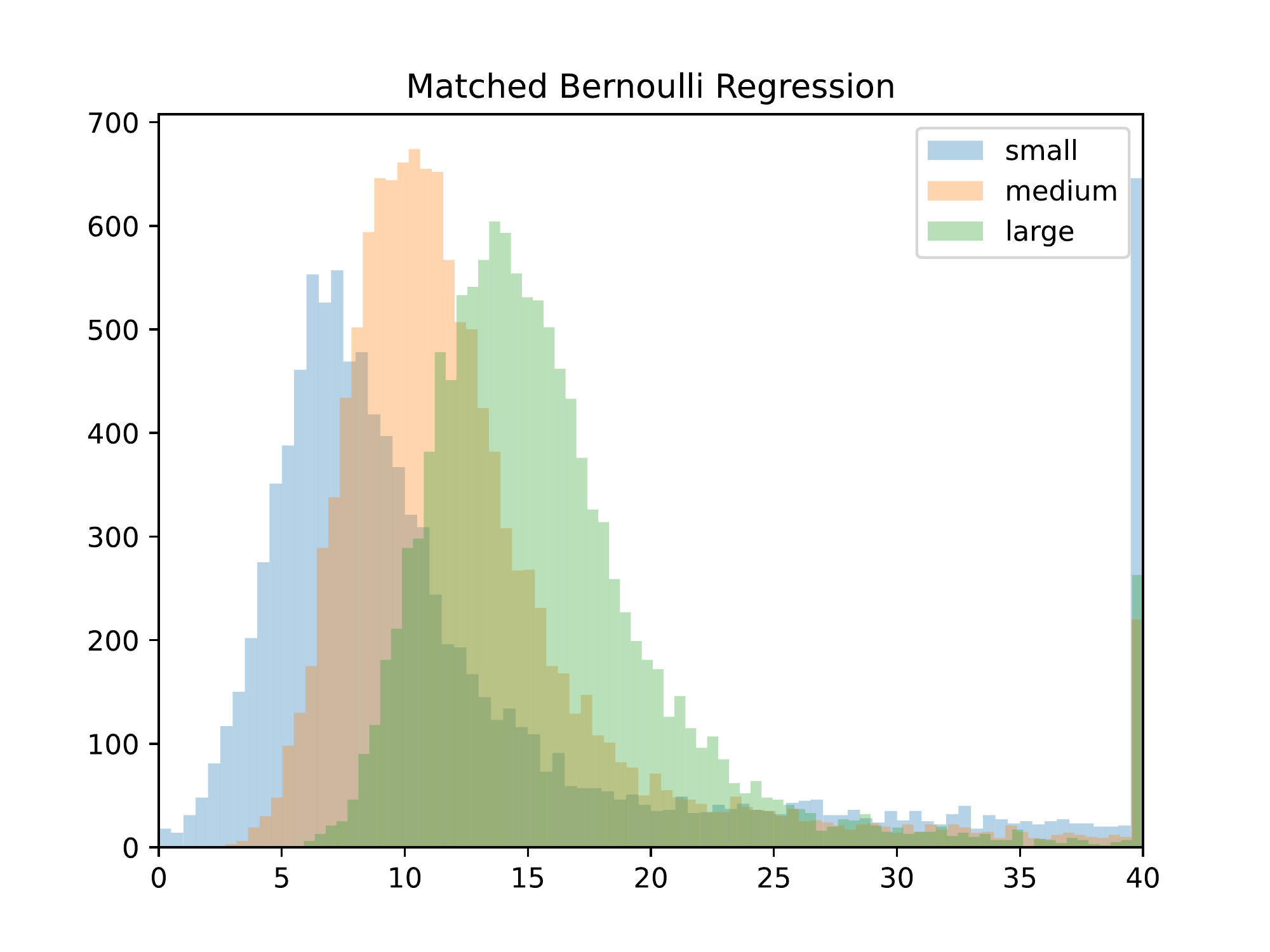}}
  \adjustbox{trim=0.2cm 0.2cm 0.2cm 0.2cm}{\includegraphics[width=4cm]{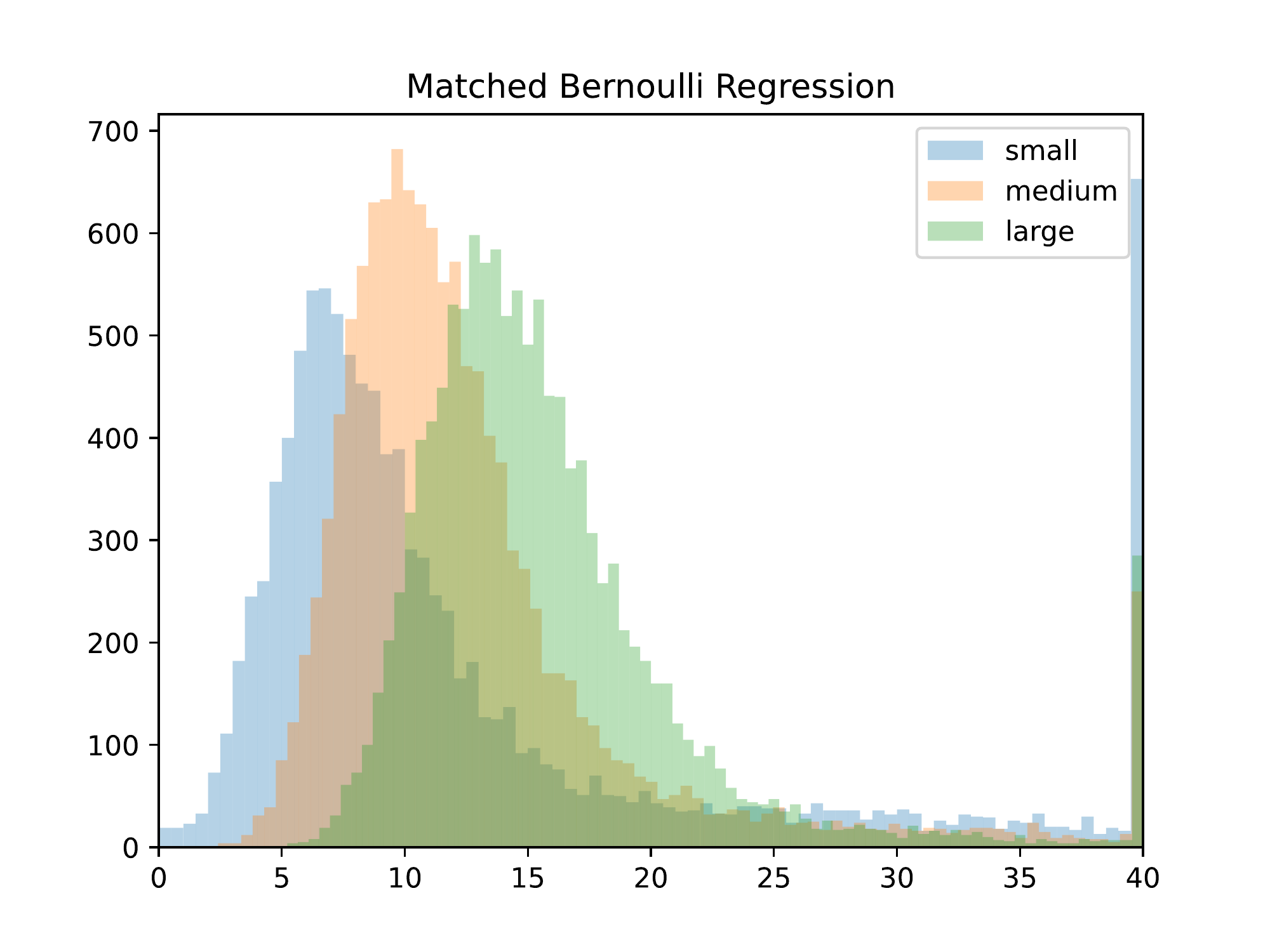}}
  \adjustbox{trim=0.2cm 0.2cm 0.2cm 0.2cm}{\includegraphics[width=4cm]{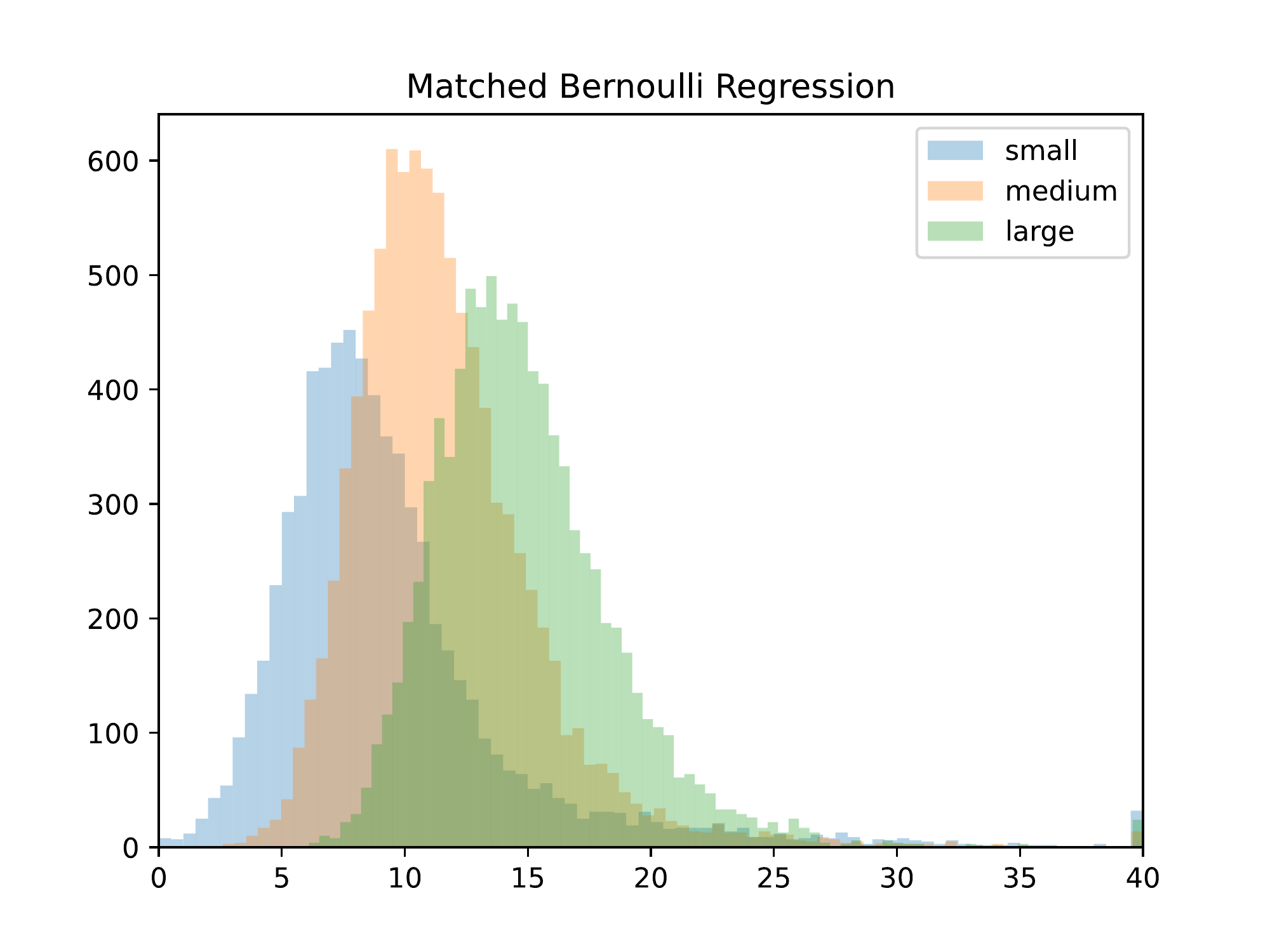}}
  \caption{Matched Bernoullis regression.}
  \label{fig:faster-rcnn-reg}
\end{subfigure}
\newline
\begin{subfigure}{\textwidth}
  \centering
  \adjustbox{trim=0.2cm 0.2cm 0.2cm 0.2cm}{\includegraphics[width=4cm]{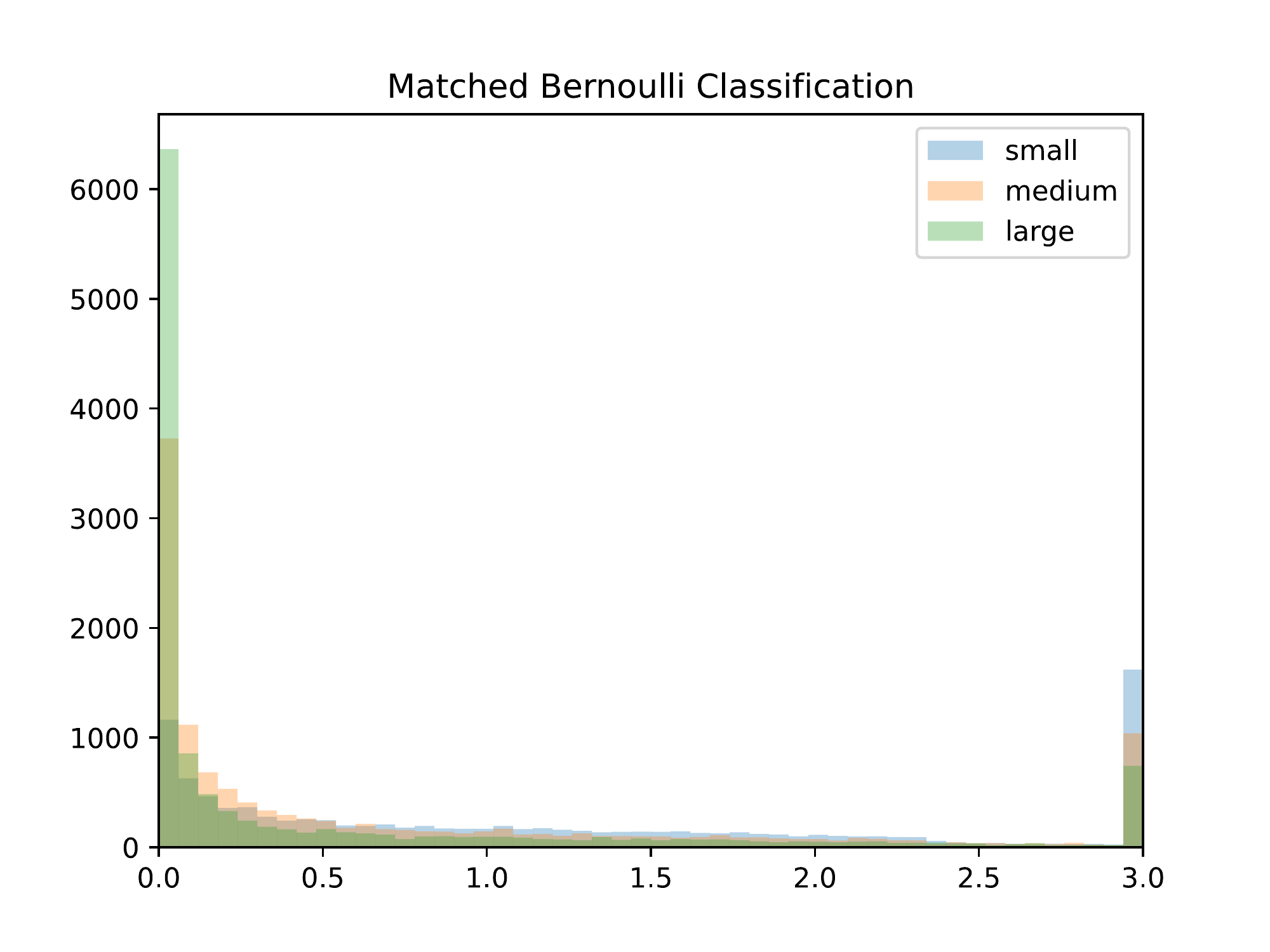}}
  \adjustbox{trim=0.2cm 0.2cm 0.2cm 0.2cm}{\includegraphics[width=4cm]{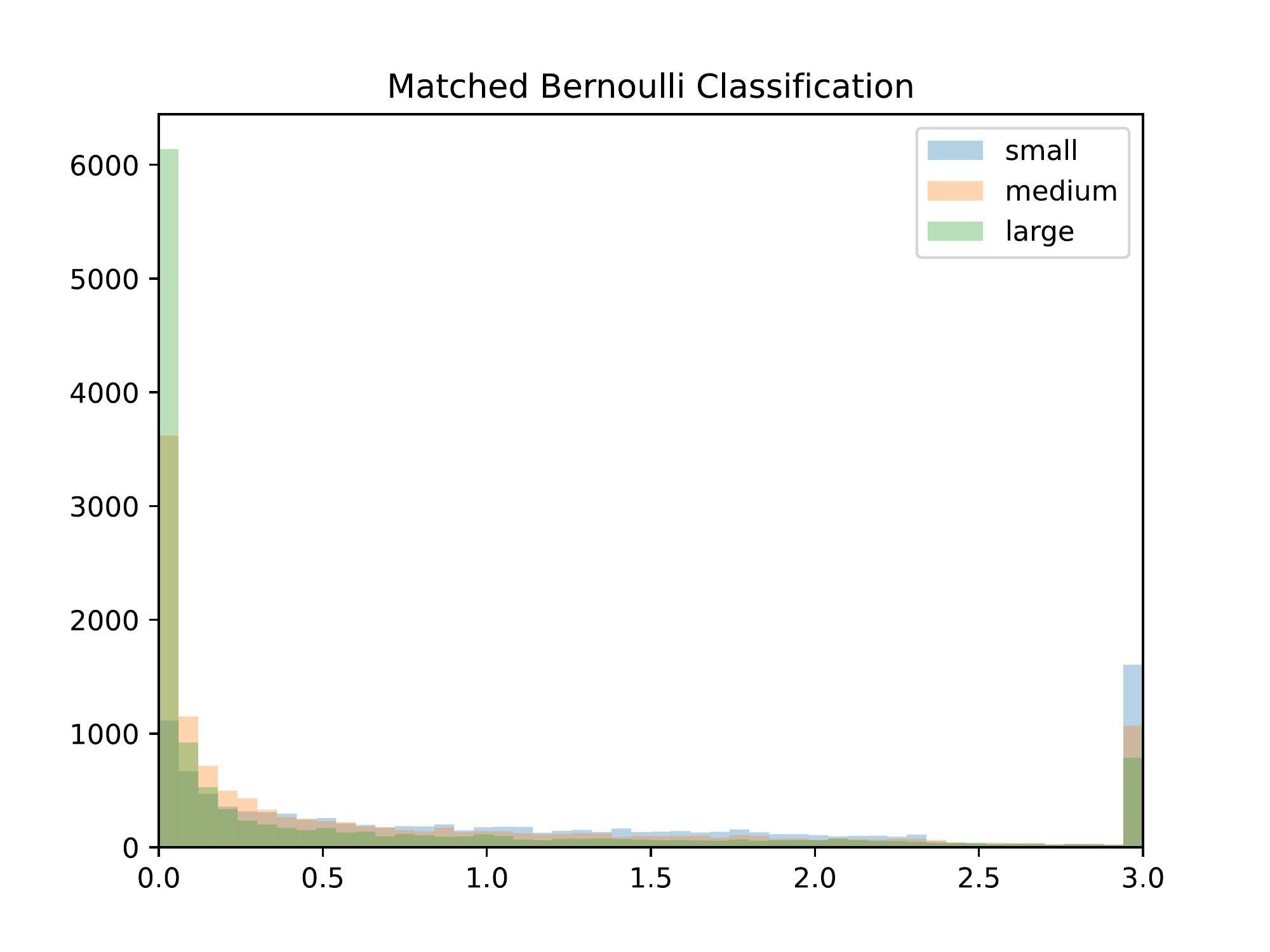}}
  \adjustbox{trim=0.2cm 0.2cm 0.2cm 0.2cm}{\includegraphics[width=4cm]{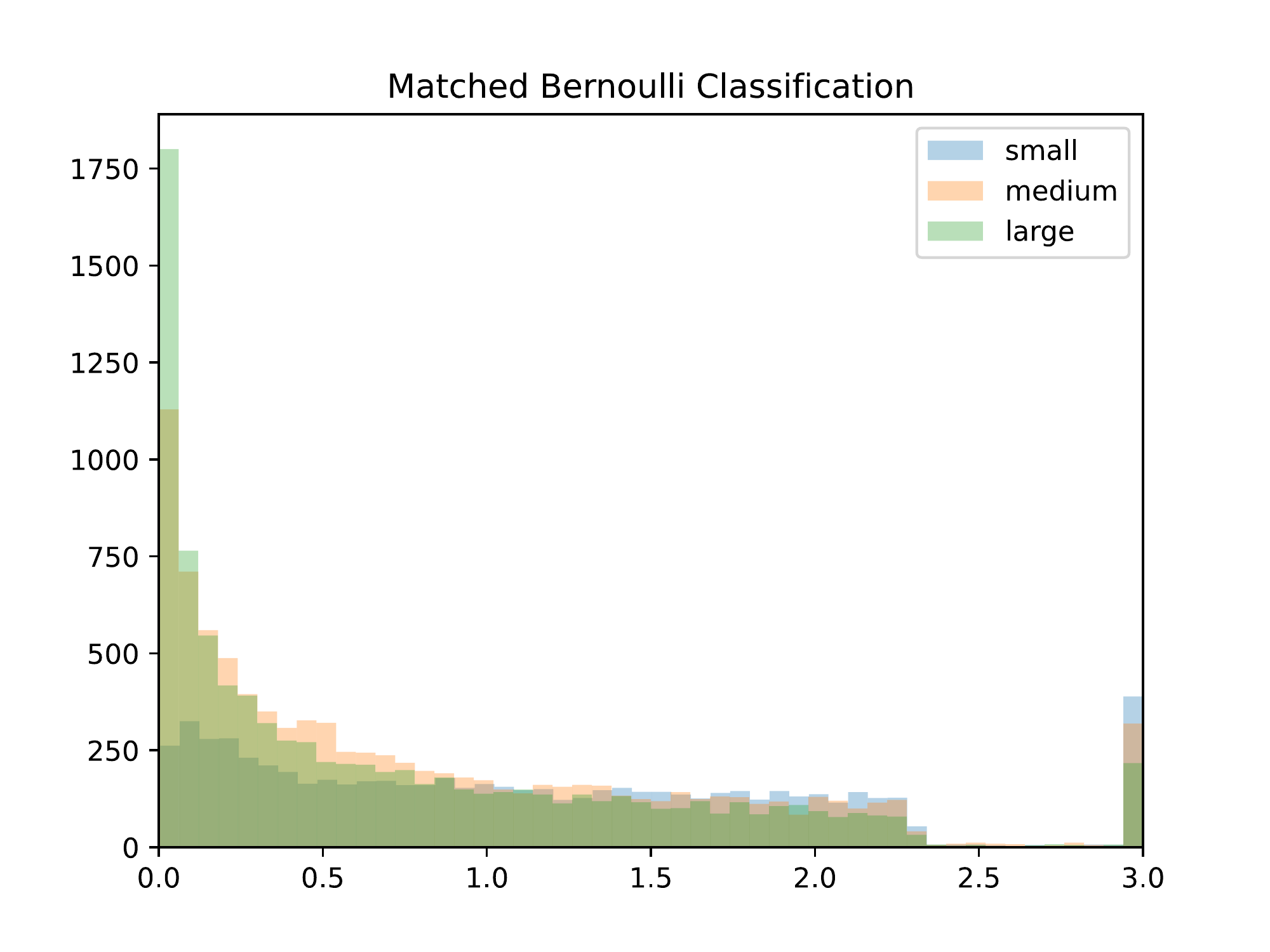}}
  \caption{Matched Bernoullis classification.}
  \label{fig:faster-rcnn-cls}
\end{subfigure}
\newline
\begin{subfigure}{\textwidth}
  \centering
  \adjustbox{trim=0.2cm 0.2cm 0.2cm 0.2cm}{\includegraphics[width=4cm]{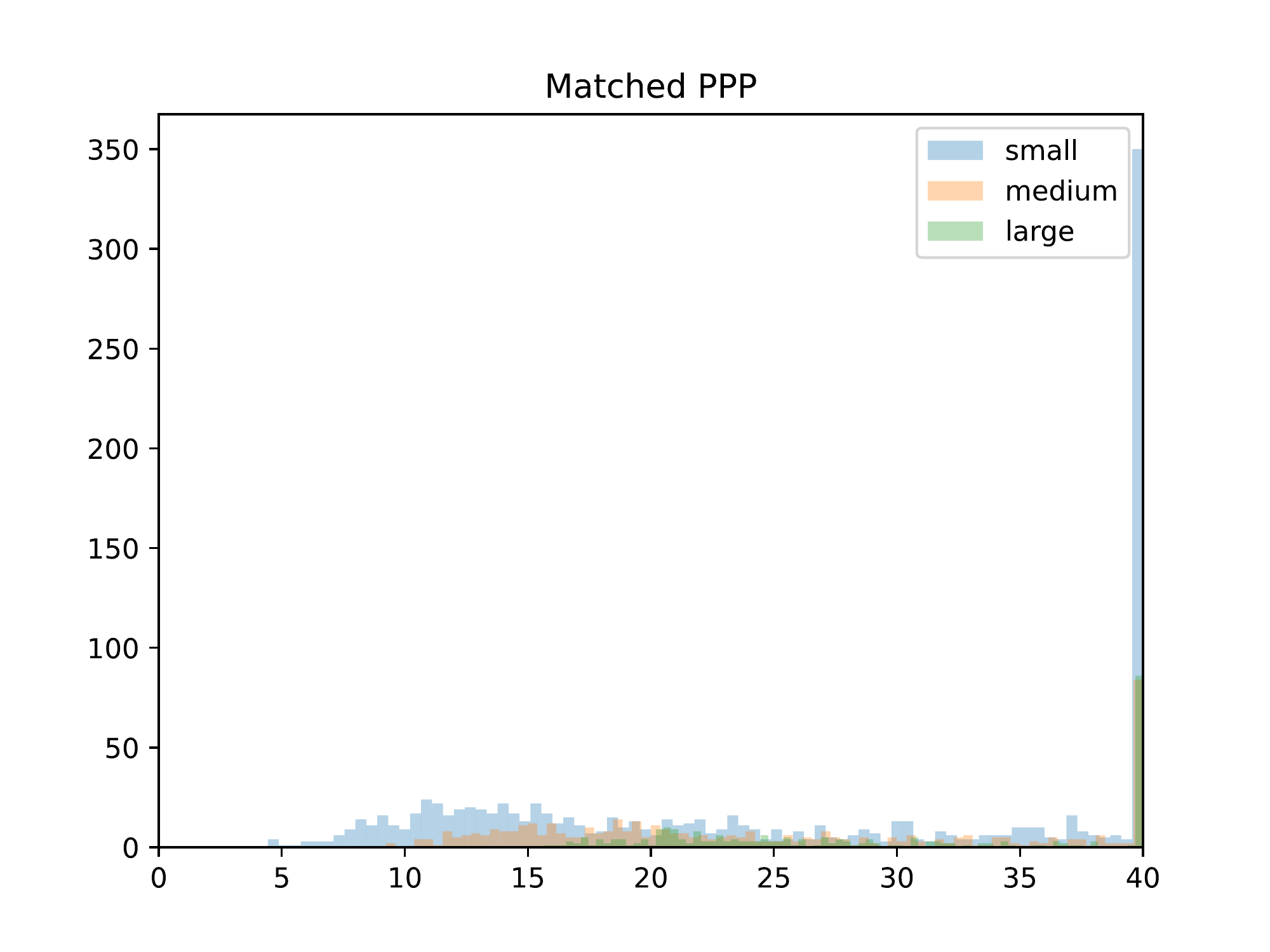}}
  \adjustbox{trim=0.2cm 0.2cm 0.2cm 0.2cm}{\includegraphics[width=4cm]{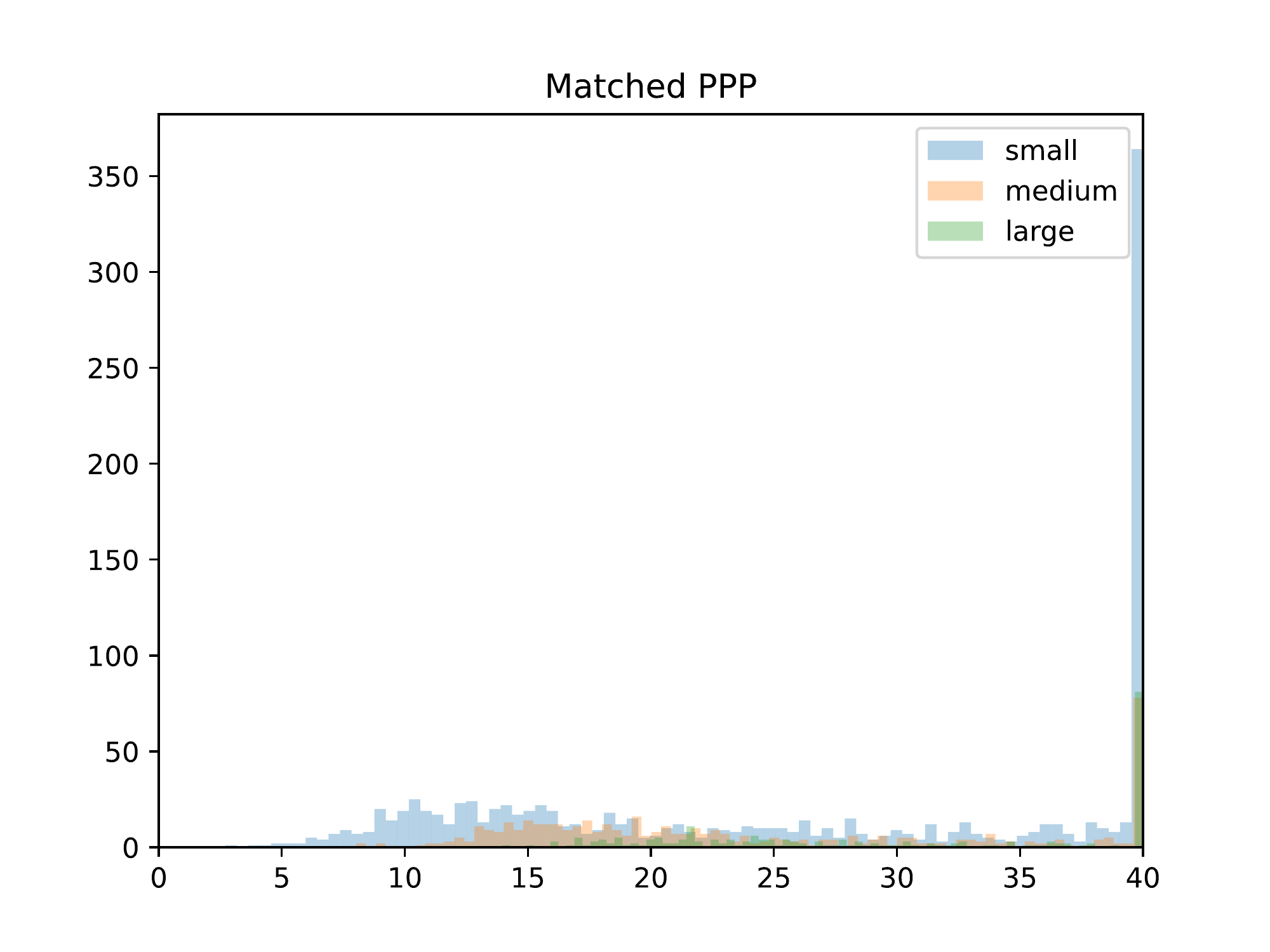}}
  \adjustbox{trim=0.2cm 0.2cm 0.2cm 0.2cm}{\includegraphics[width=4cm]{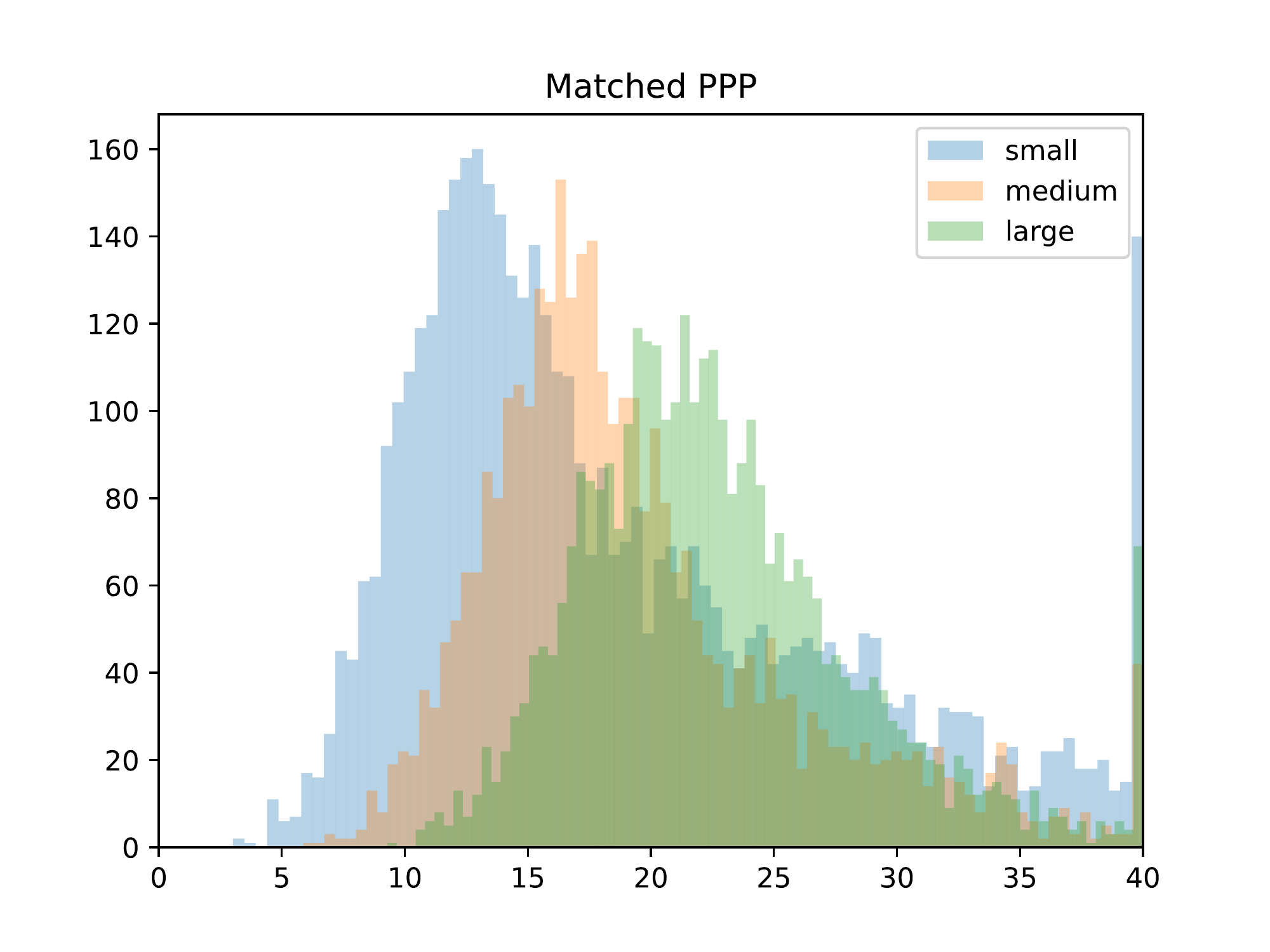}}
  \caption{Matched PPP.}
  \label{fig:faster-rcnn-ppp}
\end{subfigure}
\newline
\begin{subfigure}{\textwidth}
  \centering
  \adjustbox{trim=0.2cm 0.2cm 0.2cm 0.2cm}{\includegraphics[width=4cm]{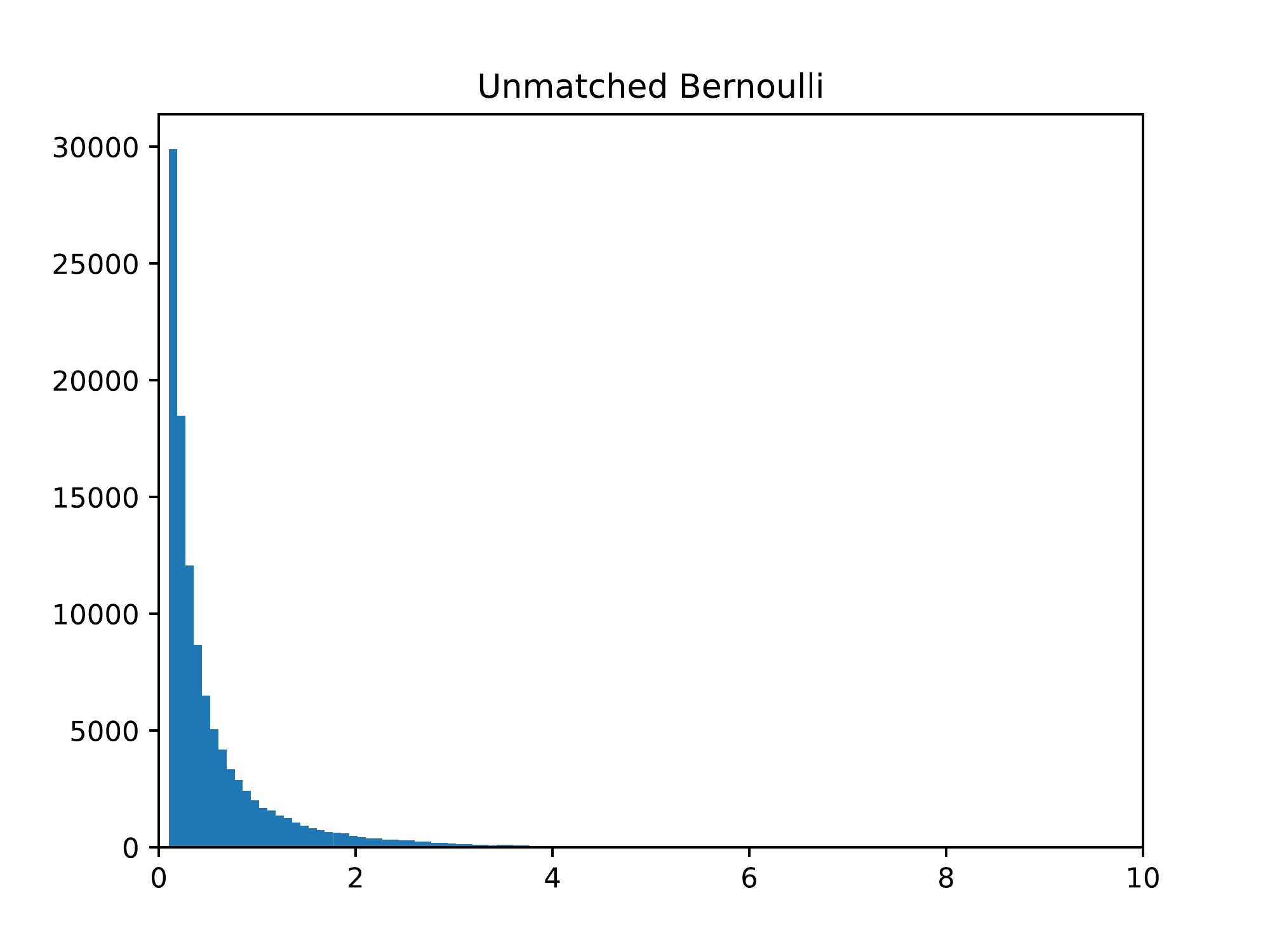}}
  \adjustbox{trim=0.2cm 0.2cm 0.2cm 0.2cm}{\includegraphics[width=4cm]{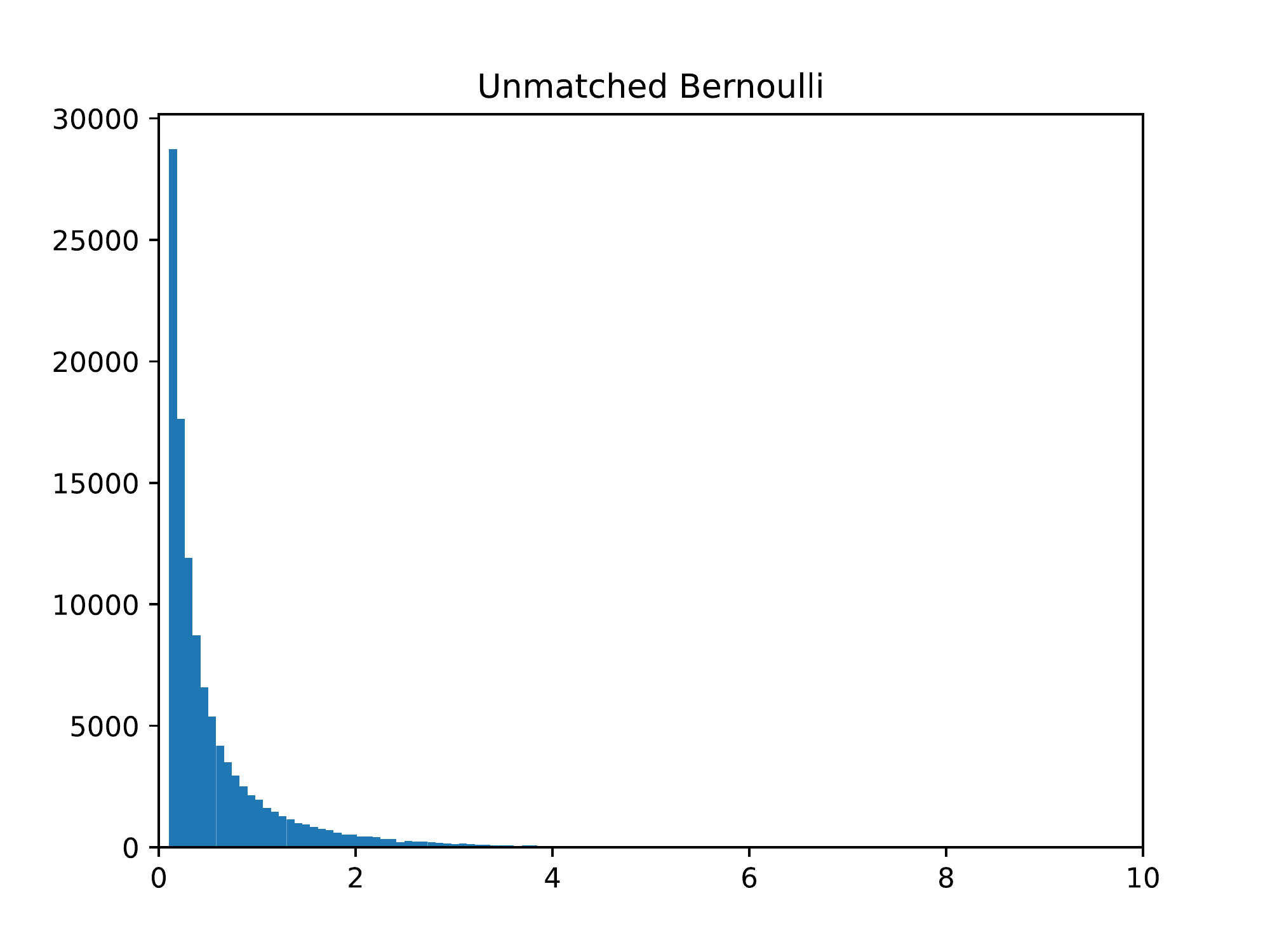}}
  \adjustbox{trim=0.2cm 0.2cm 0.2cm 0.2cm}{\includegraphics[width=4cm]{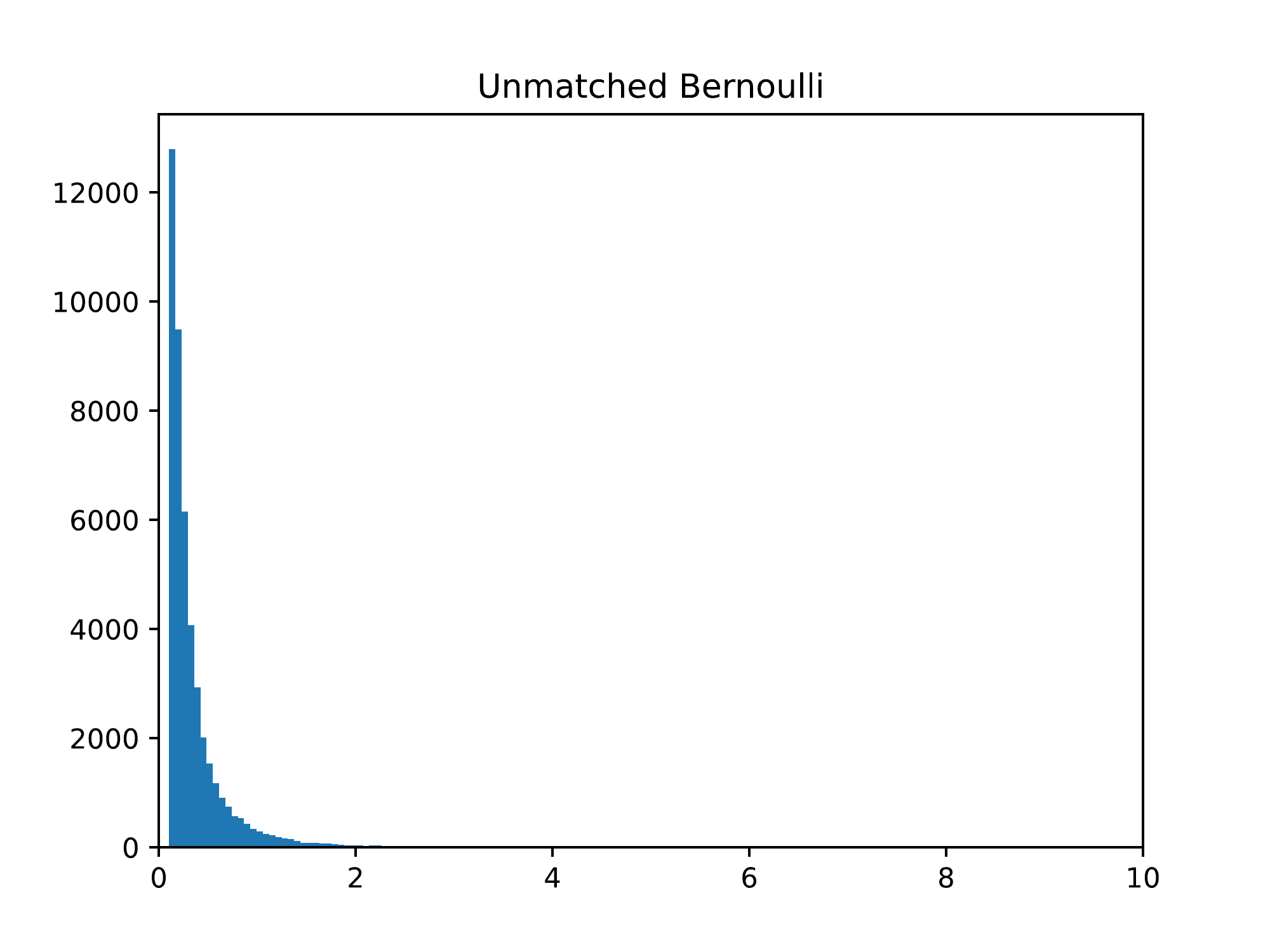}}
  \caption{Unmatched Bernoullis.}
  \label{fig:faster-rcnn-unmatched}
\end{subfigure}
    \caption{Histograms over PMB-NLL decomposition for Faster-RCNN trained with different loss functions: ES (left), NLL (middle), and MB-NLL (right). Note varying y-axes across models. }
    \label{fig:faster-rcnn-hist}
\end{figure}

To complement the PMB-NLL decomposition in Table 2, we also provide corresponding histograms for DETR, RetinaNet and Faster-RCNN in figures \ref{fig:detr-hist}, \ref{fig:retinanet-hist} and \ref{fig:faster-rcnn-hist}. The histograms show how predictions contribute to the overall PMB-NLL in the assignment with the highest likelihood. Further, for matched predictions, histograms are also decomposed based on the size of the true object. Here, we follow the COCO standard for defining small, medium and large objects. For the regression of matched Bernoullis and PPP, the values have been limited to 40 for enhanced visualizations. For the classification of matched Bernoullis, the upper limit is set to 3.

Generally, the models are worse at detecting small objects, which is shown by them being assigned to the PPP more often than medium or large objects. Further, the detectors are more confident when predicting larger objects, as can be seen from the classification histograms over matched Bernoullis. This is of course expected as large objects are inherently easier to detect. Lastly, we can observe that the histograms for models trained with ES or NLL tend to have more outliers, i.e., predictions whose values have been clipped to the visualization limits.

\section{Comparison to DETR loss}
The DETR object detector \cite{detr} popularized the concept of treating object detection as a direct set prediction task. Similar to our method, they rely on a one-to-one matching between predictions and ground truth objects. We aim to compare their formulation to ours, highlighting similarities and differences. 

\subsection{DETR Loss Revisited}
We start by reviewing the methods used in DETR. To find the matching between predictions and true objects, they rely on the Hungarian algorithm to minimize
\begin{equation}
    \label{eq:detr-minimize-objective}
    \hat{\sigma} = \argmin_{\sigma \in \mathfrak{S}_N} \sum_i^N\mathcal{L}_{\text{match}}(y_i, \hat{y}_{\sigma(i)}), 
\end{equation}
where $\hat{y}=\{\hat{y}_i\}_{i=1}^N$ is the set of predictions and $y$ is the ground truth set of objects, where we assume $y$ to be padded to size $N$ with $\varnothing$ (no object). An object $y_i=(c_i,b_i)$ consists of a class label (which can be $\varnothing$) and a bounding box $b_i\in \mathds{R}^4$, while a prediction $\hat{y_i}=(\hat{p}_{i,\text{cls}}(c_i),\hat{b}_i)$ consists of a class distribution $\hat{p}_{i,\text{cls}}(c_i)$ which assigns probabilities to all classes including $\varnothing$ and a predicted bounding box $\hat{b}_i$. Further, $\mathfrak{S}_N$ is the set of all permutations of $N$ elements and $\mathcal{L}_{\text{match}}(y_i, \hat{y}_{\sigma(i)})$ is a pair-wise matching cost defined as
\begin{equation}
    \label{eq:detr-matching-cost}
    \mathcal{L}_{\text{match}}(y_i, \hat{y}_{\sigma(i)}) = -\mathds{1}_{c_i\neq \varnothing}\hat{p}_{\sigma(i),\text{cls}}(c_i)+\mathds{1}_{c_i\neq \varnothing}\mathcal{L}_\text{box}(b_i,\hat{b}_{\sigma(i)}),
\end{equation}
with 
\begin{equation}
    \mathcal{L}_{\text{box}}(b_i,\hat{b}_{\sigma(i)}) =  \lambda_\text{iou}\mathcal{L}_{\text{iou}}(b_i,\hat{b}_{\sigma(i)})+\lambda_\text{L1}||b_i-\hat{b}_{\sigma(i)}||_1,
\end{equation}
where hyperparameters are typically set to $\lambda_\text{iou}=2$ and $\lambda_\text{L1}=5$. We can note that the matching cost for $c_i=\varnothing$ is zero.

Given the optimal matching $\hat{\sigma}$, the final loss is calculated as
\begin{equation}
    \label{eq:detr-loss}
    \mathcal{L}_\text{Hungarian}(y,\hat{y}) =  \sum_{i=1}^N \left[-\log(\hat{p}_{\hat{\sigma}(i),\text{cls}}(c_i)) 
    +
    \mathds{1}_{c_i\neq \varnothing}\mathcal{L}_\text{box}(b_i,\hat{b}_{\hat{\sigma}(i)})
    \right].
\end{equation}
However, rather than using the negative log-likelihood for class predictions, the log-probability term is down-weighted by a factor 10 when $c_i=\varnothing$. This is motivated in \cite{detr} to handle class imbalance.

\subsection{MB-NLL Relation to DETR}
This section aims at describing the MB-NLL using the same notation as DETR; to simplify the comparison we focus on MB-NLL rather than PMB-NLL. We start by comparing the matching costs before moving on to the loss formulation.

\subsubsection{Matching Cost.}
We can compare \eqref{eq:detr-matching-cost} with the cost of matching objects to Bernoulli predictions used in this work. As described in Section 3.1, we use $r=1-\hat{p}_{\text{cls}}(\varnothing)$. Further, for the Bernoulli predictions, the class distribution is assumed to be conditioned on existence, i.e., it has non-zero probability only for foreground classes. Hence, to clarify the relation to DETR we define
\begin{equation}
    {p}_{\text{cls}}(c)=
    \begin{cases}
    0 & \text{if } c=\varnothing, \\
    \hat{p}_{\text{cls}}(c)/r & \text{otherwise,}
    \end{cases}
\end{equation}
as the class distribution over foreground classes. We scale the predicted distribution $\hat{p}_{\text{cls}}(c)$ by $\frac{1}{r}$ such that ${p}_{\text{cls}}(c)$ becomes a proper distribution and fulfills $\sum_c {p}_{\text{cls}}(c) = 1$. 

In Appendix \ref{appendix:cost_matrix}, we showed how to find the assignment that maximizes the likelihood \eqref{eq:pmb_likelihood_propto} and consequently minimizes the negative log-likelihood. To enable easier comparison, we want to use the same notation as DETR and formulate a minimization over permutations
\begin{equation}
\label{eq:detr-mb-optimization-objective}
\begin{split}
    \hat{\sigma} 
    &= \argmin_{\sigma \in \mathfrak{S}_N}
    \sum_i^N\mathcal{L}_{\text{match,MB}}(y_i, p_{\sigma(i)}),
\end{split}
\end{equation}
where $\mathcal{L}_{\text{match,MB}}$ is a pair-wise matching cost. Using \eqref{eq:pmb_likelihood_propto}, we can express $\mathcal{L}_{\text{match,MB}}$ as
\begin{equation}
\label{eq:our-matching-cost-temp}
\begin{split}
    \mathcal{L}_{\text{match,MB}}(y_i, p_{\sigma(i)}) = &
    -\mathds{1}_{c_i\neq \varnothing}
    \left(\log( r_{\sigma(i)} p_{\sigma(i)}(y_i) ) \right)
    -\mathds{1}_{c_i=\varnothing} \log \left( 1-r_{\sigma(i)} \right) \\
    =& -\log(\hat{p}_{\sigma(i),\text{cls}}(c_i))
    -\mathds{1}_{c_i\neq \varnothing} \log({p}_{\sigma(i),\text{reg}}(b_i)),
\end{split}
\end{equation}
since the cost of assigning a prediction to a true object is $-\log( r_{\sigma(i)} p_{\sigma(i)}(y_i) )$, while the cost of assigning it to background is $ -\log(1-r_{\sigma(i)})$. In \eqref{eq:our-matching-cost-temp}, we have also used the fact that $p_{i}(y_i)= {p}_{i,\text{cls}}(c_i){p}_{i,\text{reg}}(b_i)$, and the relation 
\begin{equation}
    \hat{p}_{i,\text{cls}}(c_i) = \begin{cases}
    r {p}_{i,\text{cls}}(c_i) & \text{if } c_i \neq \varnothing \\
    1-r & \text{if } c_i = \varnothing,
    \end{cases}
\end{equation}
to obtain an expression that resembles \eqref{eq:detr-matching-cost} and \eqref{eq:detr-loss}. 

Further, if we assume $p_{\sigma(i),\text{reg}}(b_i)$ to be a Laplace distribution with independent box parameters $b = [b^1, b^2, b^3, b^4]$, with means $\hat{b}=[\hat{b}^1, \hat{b}^2, \hat{b}^3, \hat{b}^4]$ and scales $\hat{s}=[\hat{s}^1, \hat{s}^2, \hat{s}^3, \hat{s}^4]$, we can rewrite
\begin{equation}
\begin{split}
    - \log(p_{\sigma(i),\text{reg}}(b_i)) &= -\log\left(
    \prod_{k=1}^4 \frac{1}{2\hat{s}_{\sigma(i)}^k} \exp\left(
    -\frac{|b_i^k-\hat{b}_{\sigma(i)}^k|}{\hat{s}_{\sigma(i)}^k}
    \right)
    \right),
    \\
    &= 
    -\sum_{k=1}^4
    \log\left(
    \frac{1}{2\hat{s}_{\sigma(i)}^k} \exp\left(
    -\frac{|b_i^k-\hat{b}_{\sigma(i)}^k|}{\hat{s}_{\sigma(i)}^k}
    \right)
    \right),
    \\
    &= 
    \sum_{k=1}^4
    \frac{|b_i^k-\hat{b}_{\sigma(i)}^k|}{\hat{s}_{\sigma(i)}^k}
    +\log\left(
    \hat{s}_{\sigma(i)}^k
    \right)
    +\log\left(
    2
    \right).
\end{split}
\end{equation}
As $\log(2)$ is present in all matching costs between pairs of predictions and true objects, it does not affect the optimal assignment and can be disregarded. Further, if we let $\hat{s}_i^k=s,\forall i, k$, we can use the same argument to disregard $\log\left(\hat{s}_{\sigma(i)}^k\right)$ and replace $- \log(p_{j,\text{reg}}(b_i))$ in \eqref{eq:our-matching-cost-temp} with $||b_i-\hat{b}_{\sigma(i)}||_1 / s$, obtaining
\begin{equation}
\begin{split}
    \label{eq:mb-matching-detr}
    \mathcal{L}_{\text{match,MB}}(y_i, p_{\sigma(i)}) = & 
    -\log(\hat{p}_{\sigma(i),\text{cls}}(c_i)) 
    + \mathds{1}_{c_i\neq \varnothing} ||b_i-\hat{b}_{\sigma(i)}||_1 /s
    .
\end{split}
\end{equation}

We can now compare the expressions for the matching losses in \eqref{eq:mb-matching-detr} and \eqref{eq:detr-matching-cost}, used by MB-NLL and DETR, respectively, and analyze their similarities and differences. Rather than using the log-probabilities, the original DETR matching loss uses the class probabilities directly. In \cite{detr} this is motivated as making the classification part of the cost comparable to the regression part. Interestingly, we find that the classification log-probability is comparable to the L1 regression under the constant scale assumption. Further, the classification cost in \eqref{eq:detr-matching-cost} only evaluates the probability of the true class when the object is not $\varnothing$. For background, the cost is set to zero. In contrast, the MB matching cost also considers the log-probability of the prediction being background and favours predictions for which the probability of background $\hat{p}_\text{cls}(\varnothing)$ is small. The reason for also considering the cost of assigning predictions to background in MB-NLL is that predictions with large existence probabilities infer large penalties in the final loss function if they are not assigned to a true object.

We further highlight the difference in how $\hat{p}_\text{cls}(\varnothing)$ is handled by the two matching costs with an example. Imagine a scenario that contains two predictions and one true object, a car. Both predictions have the same regression error, but differ in their classification. Suppose that both predictions have the same $\hat{p}_\text{cls}(\text{car})=rp_\text{cls}(\text{car})$ but that the first prediction has a small $r$ and a large $p_\text{cls}(\text{car})$ whereas the second prediction has a large $r$ and a small $p_\text{cls}(\text{car})$. In \eqref{eq:detr-matching-cost}, both these predictions would be treated as equally good. For MB-NLL however, it is better to assign the second prediction to the car, simply because that implies that the first prediction, which has a small $r$, is assigned to the background.

For the regression part, \eqref{eq:detr-matching-cost} contains both an additional IoU-loss, and the hyperparameters $\lambda_\text{iou},\lambda_\text{L1}$ when compared to \eqref{eq:mb-matching-detr}. While the IoU-part has no related term in \eqref{eq:mb-matching-detr}, we find an inverse relationship for $\lambda_\text{L1}$ and the assumed constant Laplace scale $s$, i.e., $\lambda_\text{L1} = \frac{1}{s}$. Thus, the choice $\lambda_\text{L1}=5$ is equivalent of assuming a constant Laplace scale $s=0.2$, and increasing $\lambda_\text{L1}$ translates to assuming smaller spatial uncertainties.

\subsubsection{Loss Function.}
The MB-NLL training loss
\begin{equation}
    \label{eq:detr-mb-loss-LENNART}
    \begin{split}
    \mathcal{L}_\text{MB}(y,\hat{y}) 
    &= \sum_{i}^N \left[-\log(\hat{p}_{\hat{\sigma}(i)}(c_i)) 
    +
    \mathds{1}_{c_i\neq \varnothing}\sum_{k=1}^4
    \frac{|b_i^k-\hat{b}_{\hat{\sigma(i)}}^k|}{\hat{s}_{\hat{\sigma}(i)}^k}
    + \log\left(\hat{s}_{\hat{\sigma}(i)}^k\right)
    \right],
\end{split}
\end{equation}
is very similar to the matching loss, but makes use of the scaling parameters 
$\hat{s}_{\sigma(i)}^k$ predicted by our networks, whereas the matching loss uses a fixed scaling parameter $s$. 

We see that both \eqref{eq:detr-loss} and \eqref{eq:detr-mb-loss-LENNART} contain two terms, one for classification and one for regression. However, in contrast to the original DETR loss \eqref{eq:detr-loss}, we do not down-weigh the log-probability when predictions are assigned to $\varnothing$. We believe this is one of the contributing factors to the reduced number of false detections when training with the MB-NLL loss. By down-weighing the penalty for predictions assigned to $\varnothing$, the detector is encouraged to produce artificially high classification confidence. In mAP, the measure that DETR most likely has been optimized toward, the high classification confidence is generally not penalized as it only relies on the ranked confidences among predictions and not the absolute confidence values. 

Similar to the matching cost, \eqref{eq:detr-mb-loss-LENNART} lacks the IoU cost found in \eqref{eq:detr-loss}, but has an identical L1-loss when $\lambda_\text{L1}=\frac{1}{\hat s_{\sigma(i)}^1}=\dots = \frac{1}{\hat s_{\sigma(i)}^4}$. Naturally, L1-losses increase with larger $\lambda_\text{L1}$, but using \eqref{eq:detr-mb-loss-LENNART} this can also be explained as assuming smaller regression uncertainties. Finally, when training with MB-NLL, the relation between the Laplace scale and $\lambda_\text{L1}$ also shows how the network can self-regulate the regression penalties. For the network to chose suitable scales, it has to balance the two terms $\log\left(\hat{s}_{\hat{\sigma}(i)}^k\right)$ and $\frac{|b_i^k-\hat{b}_{\hat{\sigma(i)}}^k|}{\hat{s}_{\hat{\sigma}(i)}^k}$. While the first term encourages smaller scales, choosing too small scales may yield a large cost if the regression performance $|b_i^k-\hat{b}_{\hat{\sigma(i)}}^k|$ is poor.  


\end{document}